\documentclass[10pt,twocolumn,letterpaper]{article}

 \usepackage[pagenumbers]{cvpr} 

\usepackage{comment}
\usepackage[table]{xcolor}
\usepackage{makecell}
\usepackage{stmaryrd}

\usepackage[pagebackref,breaklinks,colorlinks]{hyperref}
\def\paperID{259} 
\def\confName{3DV\xspace}
\def\confYear{2024\xspace}

\usepackage{fancyhdr}
\fancypagestyle{arxivhdr}
{
   \fancyhf{}
   \setlength{\headheight}{15pt} 
\fancyfoot[C]{This paper has been accepted for publication at the International Conference on 3D Vision (3DV)}
\fancyhead[C]{\footnotesize Please cite this paper as:\\
M. Bonotto, L. Sarrocco, D. Evangelista, M. Imperoli, and A. Pretto "CombiNeRF: A Combination of Regularization Techniques for Few-Shot Neural Radiance Field View Synthesis," in International Conference on 3D Vision (3DV), 2024.}
}

\title{CombiNeRF: A Combination of Regularization Techniques for Few-Shot Neural Radiance Field View Synthesis}

\author{Matteo Bonotto$^{1,2}$\textsuperscript{\textasteriskcentered} \quad Luigi Sarrocco$^{1,2}$\thanks{Authors contributed equally to this work. This research is part of a FlexSight-supported project that has received funding from the European Union’s
Horizon 2020 research and innovation programme – INNOSUP under grant agreement No
101005711. Corresponding author: {\tt\small matteo.bonotto.2@phd.unipd.it}} \quad Daniele Evangelista$^{1,2}$ \quad Marco Imperoli$^{2}$ \quad Alberto Pretto$^{1}$\\
$^{1}$Department of Information Engineering, University of Padova, Italy \\ $^{2}$FlexSight, Via Prima Strada 35, Padova, Italy \\
}

\begin{document}
\maketitle
\thispagestyle{arxivhdr}

\begin{abstract} 
Neural Radiance Fields (NeRFs) have shown impressive results for novel view synthesis when a sufficiently large amount of views are available. When dealing with few-shot settings, i.e. with a small set of input views, the training could overfit those views, leading to artifacts and geometric and chromatic inconsistencies in the resulting rendering. Regularization is a valid solution that helps NeRF generalization. On the other hand, each of the most recent NeRF regularization techniques aim to mitigate a specific rendering problem. Starting from this observation, in this paper we propose CombiNeRF, a framework that synergically combines several regularization techniques, some of them novel, in order to unify the benefits of each. In particular, we regularize single and neighboring rays distributions and we add a smoothness term to regularize near geometries. After these geometric approaches, we propose to exploit 
Lipschitz regularization to both NeRF density and color networks and to use encoding masks for input features regularization. We show that CombiNeRF outperforms the state-of-the-art methods with few-shot settings in several publicly available datasets. We also present an ablation study on the LLFF and NeRF-Synthetic datasets that support the choices made. We release with this paper the open-source implementation of our framework.
\end{abstract}  
\section{Introduction}
\label{sec:intro}
\begin{figure}[t] \centering
    \makebox[0.21\textwidth]{Vanilla NeRF}
    \makebox[0.21\textwidth]{CombiNeRF}
    \\
    \includegraphics[width=0.23\textwidth]{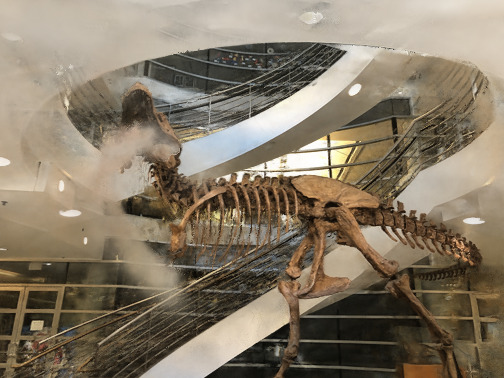}
    \includegraphics[width=0.23\textwidth]{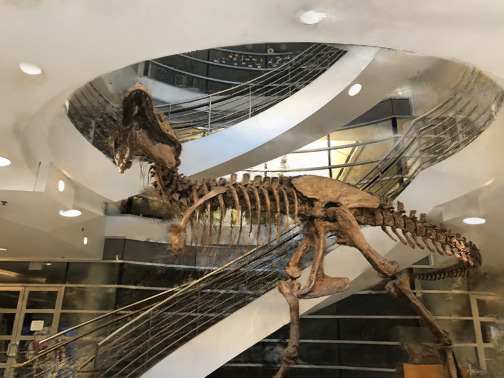}
    \\
    \includegraphics[width=0.23\textwidth]{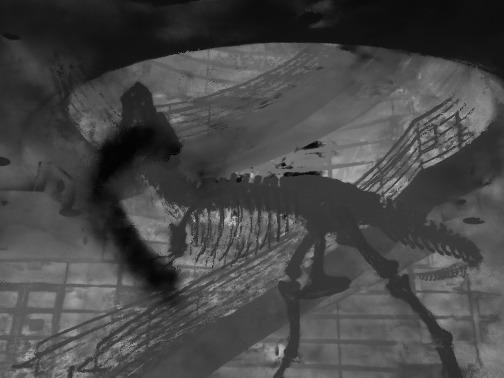}
    \includegraphics[width=0.23\textwidth]{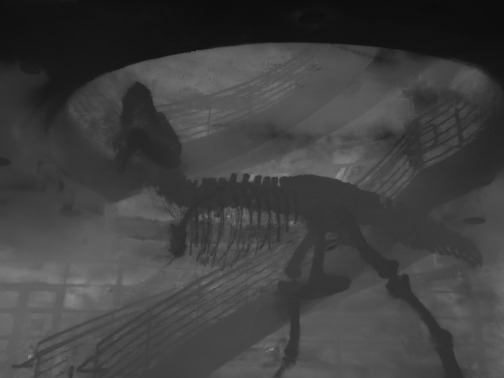}
    \caption{The figure shows how the proposed CombiNeRF achieves better results in terms of rendering and reconstruction quality in few-shot settings compared with the Vanilla NeRF \cite{Instant-NGP,torch-NGP}.} 
    \label{fig:figure0}
\vspace{-4mm}
\end{figure}
Neural Radiance Field \cite{NeRF} has emerged as a powerful approach for scene reconstruction and photorealistic rendering from a sparse set of 2D images. By leveraging multilayer perceptron (MLP) to model the volumetric scene representation, NeRF can generate high-quality novel views of scenes.
As the MLP networks are queried multiple times for each pixel to be rendered, a lightweight architecture allows to dramatically speed up the whole rendering pipeline.
The Multi-resolution Hash Encoding proposed in \cite{Instant-NGP} allows, for example, to improve scene reconstruction efficiency using a shallow MLP without any loss of visual accuracy.
However, to achieve high accuracy in its predictions and avoid artifacts (see \cref{fig:figure0}), NeRF often relies on a large number of images, which can be impractical in real-world scenarios. 

Regularization is a crucial tool for improving the visual fidelity of rendered images, ensuring more coherent predictions. Several works \cite{Reg-NeRF, Info-NeRF, Mip-NeRF360, DiffusioNeRF} have proposed to constrain the MLP network during training by adding regularization terms in the final loss.
Instead, Lipschitz regularization \cite{Lip_reg, LipMLP} is applied directly to network weights to enforce smoothness by controlling the rate of change of the network's output concerning its inputs.

In this work, we empirically select a combination of loss components to constrain the MLP network during the training phase. Firstly, (a) we modify the information-theoretic approach of InfoNeRF \cite{Info-NeRF} that includes regularization of neighboring rays distributions, (b) we adopt a smoothness term to regularize near geometries, and (c) we regularize single ray distributions in order to have high density values in regions corresponding to object surface. Additionally, as in \cite{Neuralangelo}, we add an encoding mask that encourages the learning of finer details only in the latest stages of the training phase. Finally, we impose Lipschitz regularization on both color and density networks, observing substantial improvements in the overall reconstruction and rendering quality and showing how NeRF benefits from Lipschitz regularization.
CombiNeRF avoids the need for pre-training required by similar approaches while showing promising improvements over the state of the art.

In summary, we present the following contributions: $i)$ we propose a modification of the KL-Divergence loss proposed in InfoNeRF; $ii)$ we select a combination of regularization losses, available in a unified framework, and we validate its effectiveness through ablation studies; $iii)$ to our knowledge we are the first to impose Lipschitz regularization on all the network layers on NeRF, recording a performance increase in all the tested scenarios; $iv)$ an extensive performance evaluation on public datasets, showing that CombiNeRF achieves state-of-the-art (SOTA) performance in few-shot setting; $v)$ we open-sourced the CombiNeRF implementation\footnote{\url{https://github.com/SarroccoLuigi/CombiNeRF}}.

\section{Related Work}
\label{sec:relwork}
Novel view synthesis aims to generate images from different viewpoints by utilizing a collection of pre-existing views \cite{deepvoxels, NeuralVolumes, NVSGCD, rgbd, genvs}.
Among the novel view synthesis solutions, NeRF emerged as a result of several factors as the compactness offered by the underlying structure, its domain-agnostic nature and the impressive visual quality offered. Subsequent works managed to improve over the fidelity of rendered images \cite{refnerf, lightfields, DIVeR, rawnerf, blocknerf}, reduce required time \cite{TermiNeRF, donerf, plenoxels, f2nerf} and extend NeRF capabilities and application domain \cite{dnerf, dualNeRF, xnerf, CAMPARI, fignerf, NeRFReN}.\\\\
\textbf{NeRF Regularization.}
NeRF models struggle to render novel high-quality images when supervised with only a few input views during training. 
To reduce the impact of the artifacts introduced by overfitting on training samples, previous works \cite{Reg-NeRF, PixelNerf, sparsenerf, selfnerf, harnessing, consistentnerf} focused on adding loss terms to provide additional constraints to the model, integrating prior knowledge and proposing a custom pipeline.
In InfoNeRF \cite{Info-NeRF} the entropy of the rays and the KL-divergence between the distribution of neighboring rays are both minimized in order to restrict the range of the prediction of high-density values on the object surface. Similarly, in Mip-NeRF 360 \cite{Mip-NeRF360} distortion loss was introduced to consolidate weights into as small a region as possible when rays intersect a scene object. In RegNeRF and DiffusioNeRF \cite{Reg-NeRF, DiffusioNeRF} rendered patches that are less likely to appear, according to a determined image distribution, are penalized. 
In PixelNerf \cite{PixelNerf}  prior knowledge acquired from different scenes and global context information are both leveraged by concatenating image features extracted by a Convolutional Neural Network Encoder with the input 3D positions.
In Aug-NeRF \cite{Aug-NeRF} noise is injected into input, feature and output during training to improve the generalization capabilities.\\\\
\textbf{Neural Surface Reconstruction.}
Implicit functions like signed distance functions (SDFs) \cite{IDR, physg, iron, GO-Surf} and occupancy maps \cite{DVR, NeuralBlox} are best suited for representing objects on the scene with a defined surface geometry.
Both NeuS and HF-NeuS \cite{NeuS, HF-NeuS} reparametrize the rendering equation used in NeRF to exploit SDFs properties.
In order to increase the training and inference speed and facilitate the learning of high-frequency details, Instant-NSR and Neuralangelo \cite{Instant-NSR, Neuralangelo} exploit the hash grid encoding proposed in Instant-NGP \cite{Instant-NGP}, while PermutoSDF \cite{PermutoSDF} employs a permutohedral lattice to decrease the memory accesses. To recover smoother surfaces in reflective or untextured areas, Neuralangelo and PermutoSDF add a curvature term loss. Additionally, Neuralangelo discards values from the finer levels of the hash grid encoding to encourage the learning of coarse details on the first training iterations. To prevent the color network overfitting due to the injection of the curvature loss term, PermutoSDF employs the Lipschitz constant regularization method proposed by  Liu \etal \cite{LipMLP}.\\
In our work, we exploit both Lipschitz regularization, leveraged in neural surface reconstruction to avoid geometry over-smoothness, and an encoding mask, finding out they provide additional benefits in the few-shot setting.
\begin{figure*}[t]
\centering
\includegraphics[width=0.86\textwidth]{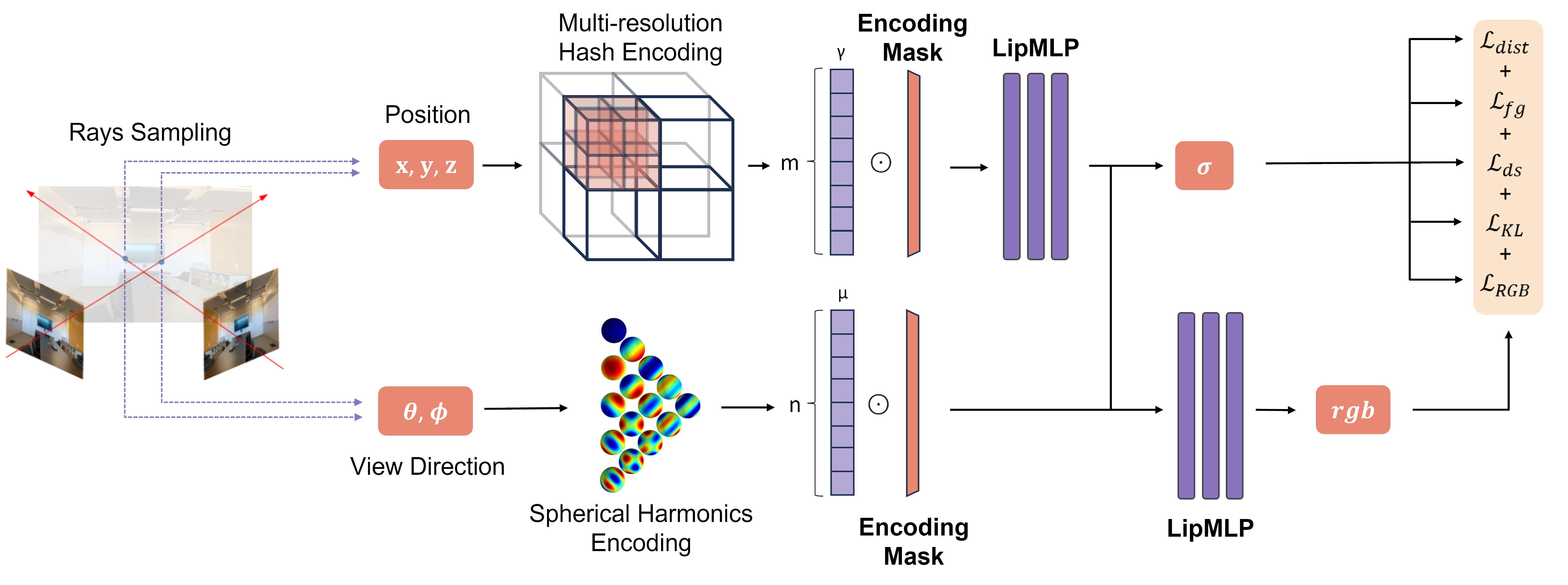}
    \caption{Overview of the CombiNeRF framework. We sample 3D points over a batch of rays passing through the scene. Position and view direction are respectively encoded through Multi-resolution Hash Encoding and Spherical Harmonics and fed to the Lipschitz network (LipMLP) after being masked. Networks' outputs are used by volumetric rendering for estimating the expected color $C$ and depth $d$ of each ray, while different loss terms are computed to regularize the training process. CombiNeRF combines $i)$ all these regularization losses, $ii)$ the Lipschitz network instead of the original MLP, $iii)$ the Encoding Mask approach used for masking the networks' input.} 
    \label{fig:combinerf_method}
\vspace{-3mm}
\end{figure*}

\section{Method}
\label{sec:method}
A graphical overview of the CombiNeRF method proposed in this work is given in \cref{fig:combinerf_method}. The proposed approach combines multiple regularization techniques in a unified and flexible framework that helps NeRF generalization in few-shot scenarios. In this section, after providing a preliminary theoretic background, all the combined losses will be described together with the Encoding Mask and Lipschitz regularization techniques. For each technique, our improvements and modifications will be described in detail to better highlight the main contributions of this work.
\subsection{Preliminaries}
\textbf{Neural Radiance Fields.} 
Neural Radiance Field \cite{NeRF} exploits an MLP to encode the scene. For each input 3D point and view direction, it outputs the density \(\sigma\) of the point and its color value \(c\). For each pixel, points are sampled on its corresponding camera ray \(r(\mathbf{t})= o + \mathbf{t}d\) and fed to the MLP network. The rendered color is approximated by: 
\begin{equation}
\begin{split}
\begin{gathered}
\label{eq:rendering}
    C(r) = \sum_{i=1}^{N} T_i \left(1 - \exp(-\sigma_i \delta_i)\right) c_i\ ,
\end{gathered}
\end{split}
\end{equation}
where \(\delta_{i} = \mathbf{t_{i+1}}-\mathbf{t_{i}}\) is the distance between the \(i^{th} \) point and its adjacent sample, \(N\) is the number of samples and the transmittance \(T_{i}\) is computed as:
\begin{equation}
\begin{split}
\begin{gathered}
\label{eq:transmittance}
 T_{i} = \exp\left(-\sum_{j=1}^{i-1} \sigma_j \delta_j\right).
 \end{gathered}
\end{split}
\end{equation}
The rendered pixel color \(C(r)\) can be expressed in terms of a weighted sum of the color values \(C(r) = \sum_{i=1}^{N} w_{i}c_{i}\) where \(w_{i}=T_{i}\alpha_{i}\) with  \(\alpha_{i} = \left(1 - \exp(-\sigma_i \delta_i)\right)\).
The final loss to train the network is the total squared error between
the rendered and true pixel colors:
\begin{equation}
    \mathcal{L}_{RGB}= \sum_{r\in\mathcal{R}} \| C(r) - C^{\ast}(r) \|_2^2,
\end{equation}
where $\mathcal{R}$ is the set of rays in each batch, and $C^{\ast}(r)$ is the
ground truth RGB color for ray $r$.\\\\
\textbf{Multi-resolution Hash Encoding.}
In NeRF, a sine-based positional encoding is used to map the 3D point position to a high-dimensional feature vector in order to capture high-frequency details. On the other hand, the Multi-resolution Hash Encoding proposed by Müller \etal \cite{Instant-NGP}  adopts multi-resolution grids, achieving a similar result while significantly accelerating training and inference times. 
This is achieved by first identifying the grid cell containing the point and then assigning the result of the spatial interpolation of the features defined on the corners.
As a result, a lightweight MLP can be employed to process the encoded points. Given an input point \(x \in \mathbb{R}^{3}\), its feature vector \(\gamma(x)\in \mathbb{R}^{FL}\) can be expressed as:
\begin{equation}
\begin{split}
\begin{gathered}
\label{eq:hashenc}
 \gamma(x) = \left[\gamma_{1, 1}(x), ..., \gamma_{F, 1}(x), ..., \gamma_{F, L}(x)\right] ,
 \end{gathered}
\end{split}
\end{equation}
where \(F\) is the dimension of the feature space and \(L\) is the number of levels.
To minimize memory usage, each grid corner index is mapped into a hash table that holds its corresponding feature vector. Collisions are limited to finer levels where \((N_{l}+1)^3 > E\), \(N_{l}\) and \(E\) being the resolution at level \(l\) and the maximum number of entries in the hash maps respectively, and they are handled through gradient averaging. 
This ensures that the results maintain comparable quality to using a multi-resolution grid without hash encoding.
\subsection{Regularization Losses}
\label{sec:regularization_losses}
\textbf{KL-Divergence Loss.}
When dealing with a few training views, the standard models \cite{NeRF,Instant-NGP} struggle to generalize to unseen views.     
As 3D scenes exhibit piece-wise smooth surfaces, we enforce similar distributions of weight values among neighboring rays.
Differently from \cite{Info-NeRF}, we compute the ray density $p(\textbf{r})$ on weights \(w_i\), instead of alpha values \(\alpha_i\):

\begin{equation}
\label{eq:probweight}
    p(\textbf{r}_i) = \frac{w_i}{\sum_j w_j} = \frac{T_i \alpha_i}{\sum_j T_j \alpha_j}.
\end{equation}

Given an observed ray $\textbf{r}$, we sample another neighboring ray $\hat{\textbf{r}}$ minimizing the KL-Divergence ($D_{KL}$) between their weight distribution. The corresponding loss is defined as follows:
\begin{equation}
\label{eq:KL}
\begin{split}
\begin{gathered}
    \mathcal{L}_{KL} = D_{KL}\bigg(  P(\textbf{r}) || P(\hat{\textbf{r}}) \bigg) = \sum_{i=1}^N p(\textbf{r}_i) \log \frac{p(\textbf{r}_i)}{p(\hat{\textbf{r}}_i)} .
\end{gathered}
\end{split}
\end{equation}
We select the neighboring ray by choosing one of the possible adjacent pixel rays. After sampling uniformly at random one of the four adjacent rays for each training ray, we compute $\mathcal{L}_{KL}$.\\\\
\textbf{Distortion and Full Geometry Loss.}
A common problem with NeRF is the presence of "floaters", disconnected regions of dense space, usually located close to the camera planes, which cause the presence of blurry cloud artifacts on images rendered from novel views.
Previous works \cite{Mip-NeRF360, DiffusioNeRF} showed that those artifacts can be removed by adding a loss term $\mathcal{L}_{dist}$ that takes into account the distances $t_i$ and weights $w_i$ of the $N$ points sampled on the rays, and the depth
\begin{equation}
    d(r) = \frac{\sum_{i=1}^Nw_it_i}{\sum_{i=1}^Nw_i}, 
\end{equation}
of each ray. Thus, the $\mathcal{L}_{dist}$ loss can be written as:
\begin{equation}
    \label{eq:dist}
    \begin{split}
    \begin{gathered}
    \mathcal{L}_{dist} =  \frac{1}{d(r)} \Bigg( \sum_{i,j}w_iw_j\left| \frac{t_i+t_{i+1}}{2} - \frac{t_j+t_{j+1}}{2}\right| \\
    + \frac{1}{3} \sum_{i=1}^Nw_{i}^{2}(t_{i+1}-t_i)  \Bigg),
    \end{gathered}
    \end{split}
\end{equation}
where the depth factor penalizes dense regions near the camera. The first term minimizes the weighted distances between all pairs of interval midpoints and the second term minimizes the weighted size of each individual interval. In this way, weights on the ray are encouraged to be as compact as possible by pulling distant intervals towards each other, consolidating weights into a single interval or a small number of nearby intervals, and minimizing the width of each interval.

In \cref{eq:dist}, all possible intervals for each ray are considered. As the total combination of all $(i,j)$ would drastically increase the amount of memory required, to reduce the total load we consider only a subset over the total number of sampled rays on which we compute the loss.

Furthermore, a normalization loss term $\mathcal{L}_{fg}$ is considered, following \cite{DiffusioNeRF}, that encourages the weights to sum to unit, as the ray is expected to be fully absorbed by the scene geometry in real scenarios:
\begin{equation}
    \mathcal{L}_{fg} =  \left(  1-\sum_{i=1}^Nw_i  \right)^2,
\end{equation}

In addition \(\mathcal{L}_{fg}\) enforces \(\mathcal{L}_{KL}\) by driving the model to treat the weights \(w_{i}\) as probabilities, \(p(\textbf{r}_i) \approx w_{i}\) (See \cref{eq:probweight}).\\\\
\textbf{Depth Smoothness Loss.}
Similarly to KL-Divergence loss, the depth smoothness constraint encourages smooth surfaces. However, while the former term requires similarity between the compared distributions, the latter term only computes the difference between the expected depth values.
As in \cite{Reg-NeRF}, given the estimated depth $d(r)$ of a pixel ray and a patch of size $S_{patch}$, the depth smoothness constraint is defined as follows:
\begin{equation}
    \begin{split}
    \begin{gathered}
    \mathcal{L}_{ds} = \sum_{p\in P}\sum_{i,j=1}^{S_{patch}-1} \Bigg( \big( d(r_{ij})-d(r_{i+1j})\big)^2 \\
    +\big( d(r_{ij})-d(r_{ij+1}) \big)^2 \Bigg),
    \end{gathered}
    \end{split}
\end{equation}
where $P$ is the set of all the rendered patches and $r_{ij}$ is the ray passing through the pixel $(i,j)$ of patch $p$. 

\subsection{Encoding Mask}
Input encoding allows preserving of high-frequency details \cite{NeRF}. However, when only few images are available, the network is more sensitive to noise. In this scenario, high-frequency components exacerbate the problem, preventing the network from exploring more in depth low-frequency information and consequently learning a coherent geometry.
As in \cite{Free-NeRF, Neuralangelo}, we use a mask to filter out high-frequency components of the input in early iterations in order to let the network focus on robust low-frequency information. Given the length $l=L\cdot F$ of the resulting Multi-resolution Hash Encoding given by \cref{eq:hashenc}, the mask $m$ is defined as:
\begin{equation}
    m= [1_{1},...1_{\llbracket l\cdot x \rrbracket},0,...0_{l}],
\end{equation}
where $x$ is the ratio of features to keep active. The resulting feature vector given in input to the MLP network will be the element-wise product $\gamma \odot m$ between the Multi-resolution Hash Encoding of the input and the mask.
The ratio $x$ is set to keep only the coarsest features during the initial phase of the training, progressively including the remaining features.
In CombiNeRF, we apply Instant-NGP Multi-resolution Hash Encoding on input positions and Spherical Harmonics Encoding on the view directions (see \cref{fig:combinerf_method}). The progressive mask can be used on both inputs in order to filter high-frequency terms on early training iterations.


\subsection{Lipschitz Network}
We define $f_\Theta(x,t)$ as the function representing the implicit shape, expressed by means of a neural network, where $\Theta = \{W_i,b_i\}$ is the set of parameters containing weights $W_i$ and biases $b_i$ of each layer $l_{i}$ of the network.

A function is called Lipschitz continuous if there exists a constant $k\geq0$ such that: 
\begin{equation}
\label{eq_lipschitz}
    \|f_\Theta(x_0) - f_\Theta(x_1)\| \leq k \|x_0 - x_1\| ,
\end{equation}
for all possible inputs $x_0$, $x_1$, where $k$ is called the Lipschitz constant which bounds how fast the function $f_\Theta$ can change. 

We employ the regularization method proposed in \cite{LipMLP}, which was already used by PermutoSDF \cite{PermutoSDF} on the color network to balance out the effect introduced by the curvature regularization applied to the SDF network. However, differently from PermutoSDF we apply the regularization to both NeRF density and color networks, recording an increase in performance with respect to the original MLP network.
Given the trainable Lipschitz bound $k_i$ associated to layer $l_{i}$,
the normalization scheme is described as:
\begin{equation}
        \begin{split}
        \begin{gathered}
        y = \sigma(\hat{W_i} x + b_i)   \\
        \hat{W_i} = f_n(W_i, \ln(1+e^{k_i})),
        \end{gathered}
        \end{split}
\end{equation}
where $f_n$ is a normalization function and the term $\ln(1+e^{k_i})$ allows to avoid negative values for $k_i$. The normalization scales each row of $W_i$ to have an absolute value row-sum less or equal to $\ln(1+e^{k_i})$ in order to satisfy \cref{eq_lipschitz}.


\subsection{Overall Method}
CombiNeRF combines the previously described regularization techniques regarding losses and network structure, hence the name \textit{CombiNeRF}.
Thus, we can write the final loss as:
\begin{equation}
    \begin{split}
        \begin{gathered}
        \mathcal{L}_{CombiNeRF} = \mathcal{L}_{RGB}+\lambda_{dist}\cdot\mathcal{L}_{dist}+\\\lambda_{fg}\cdot\mathcal{L}_{fg}+\lambda_{ds}\cdot \mathcal{L}_{ds}+\lambda_{KL}\cdot \mathcal{L}_{KL} ,
        \end{gathered}
    \end{split}
\end{equation}
where $\lambda$ are the hyperparameters controlling the contribution of each loss.
In addition, CombiNeRF includes Lipschitz regularization and the Encoding Mask technique.
The proposed CombiNeRF offers a unified implementation of all the regularization techniques described above, outperforming current SOTA methods on few-shot scenarios as demonstrated in the following experimental section.
\definecolor{White}{rgb}{1,1,1}
\definecolor{LightCyan}{rgb}{0.74,0.83,0.9}

\section{Experiments} 
\label{sec:experiment}
In this section we present the dataset and metrics we use for the experiments, we provide relevant details about CombiNeRF and we show quantitative and qualitative comparisons made against the SOTA methods. 
Finally, an ablation study is made to evaluate the contribution of each component of which CombiNeRF is built.

\subsection{Setup}

\textbf{Dataset.}
Our experiments include the LLFF dataset \cite{LLFF} and the NeRF-Synthetic dataset \cite{NeRF} under few-shot settings. LLFF is composed by 8 complex scenes, representing real-world scenarios, with 20-62 images for each scene captured with a consumer camera. Instead, NeRF-Synthetic is composed of 8 synthetic scenes with view-dependent light transport effects, in which each scene is composed of 100 training images and 200 test images.

We evaluate LLFF on 3/6/9 input views, following the protocol proposed in RegNeRF \cite{Reg-NeRF}, while in NeRF-Synthetic we train 8 views and test 25 views following DietNeRF \cite{Diet-NeRF}.\\\\
\textbf{Metrics.}
To evaluate the performance of CombiNeRF against the SOTA methods, we use 3 different metrics: peak signal-to-noise ratio (PSNR), structural similarity index measure (SSIM) \cite{SSIM}, and learned perceptual image patch similarity (LPIPS) \cite{LPIPS}. We also take into consideration the geometric mean of $\text{MSE}=10^{-\text{PSNR}/10}$, $\sqrt{1-\text{SSIM}}$ and LPIPS following \cite{Reg-NeRF}, in order to have an easier and unified comparison.\\\\
\textbf{Implementation Details.}
For our experiments, we used the torch-ngp \cite{torch-NGP} implementation of Instant-NGP \cite{Instant-NGP} as base code, since from our experience it generally outperforms the original NeRF implementation \cite{NeRF} from both a runtime and accuracy points of view. We refer to torch-ngp as "Vanilla NeRF" in the experiments. We developed CombiNeRF on top of torch-ngp, thus obtaining a unified and unique implementation that embeds all the contributions shown in the previous sections. As already mentioned, the final implementation of CombiNeRF is made publicly available with this paper.

For the LLFF dataset, we set $\lambda_{dist} = 0$ for the first 1000 iterations and then $\lambda_{dist} = 2\cdot10^{-5}$ until the end and $\lambda_{fg} = 10^{-4}$, $\lambda_{KL} = 10^{-5}$, $\lambda_{ds} = 0.1$ and $S_{patch} = 4$, while we use the encoding mask only on the density network in which $x$ saturates after 90\% of the total iterations. In the NeRF-Synthetic dataset we set $\lambda_{dist} = 0$ for the first 1000 iterations and then $\lambda_{dist} = 2\cdot10^{-3}$ until the end and $\lambda_{fg} = 10^{-3}$, $\lambda_{KL} = 10^{-5}$, $\lambda_{ds} = 0.01$ and $S_{patch} = 4$, $x$ saturates after 20\% of the total iterations. We also set the number of sampled rays for each iteration to 4096 and 7008 and  the number of levels for the Instant-NGP Multi-resolution Hash Encoding to 16 and 32 for LLFF and NeRF-Synthetic, respectively.

\begin{table*}[t] \centering
    \resizebox{\textwidth}{!}{
    \Huge
    \rowcolors{2}{LightCyan}{White}
        \begin{tabular}{p{10cm}|*{3}{c}|*{3}{c}|*{3}{c}|*{3}{c}}
        \toprule
        & \multicolumn{3}{c}{PSNR $\uparrow$} 
                               & \multicolumn{3}{c}{SSIM $\uparrow$} 
                               & \multicolumn{3}{c}{LPIPS $\downarrow$} 
                               & \multicolumn{3}{c}{Average $\downarrow$}  \\
                               & 3-view & 6-view & 9-view
                               & 3-view & 6-view & 9-view 
                               & 3-view & 6-view & 9-view 
                               & 3-view & 6-view & 9-view \\
        \midrule
        Vanilla NeRF   &  17.71 & 22.03 & 24.21 & 0.544 & 0.738 & 0.811 & 0.303 & 0.149 & 0.101 & 0.155 & 0.081 & 0.057\\
        RegNeRF    &  19.08 & 23.10 & 24.86 & 0.587 & 0.760 & 0.820 & 0.336 & 0.206 & 0.161 & 0.146 & 0.086 & 0.067\\    
        FreeNeRF    &  19.63 & 23.73 & 25.13 & 0.612 & 0.779 & 0.827 & 0.308 & 0.195 & 0.160 & 0.134 & 0.075 & 0.064 \\        
        DiffusioNeRF     &  19.88 & \textbf{24.28} & 25.10 & 0.590 & 0.765 & 0.802 & 0.192 & \textbf{0.101} & \textbf{0.084} & 0.118 & 0.071 & 0.060\\
        \midrule
            CombiNeRF       &  \textbf{20.37} & 23.99 & \textbf{25.15} & \textbf{0.686} & \textbf{0.805} & \textbf{0.841} & \textbf{0.191} & 0.106 & \textbf{0.084} & \textbf{0.101} & \textbf{0.060} & \textbf{0.049} \\
        \bottomrule
        \end{tabular}
    }
    \caption{Comparison of CombiNeRF with SOTA methods on the LLFF dataset with 3/6/9 input view few-shot settings. DiffusioNeRF row relates to the "Geometric Baseline" of the corresponding paper, as it has the highest metric scores.}
    \label{tab:table8}
\vspace{-2mm}
\end{table*}
\begin{figure*}[ht!] \centering
    \makebox[0.19\textwidth]{Vanilla NeRF}
    \makebox[0.19\textwidth]{RegNeRF}
    \makebox[0.19\textwidth]{FreeNeRF}
    \makebox[0.19\textwidth]{CombiNeRF}
    \makebox[0.19\textwidth]{Ground Truth}
    \\
    \includegraphics[width=0.19\textwidth]{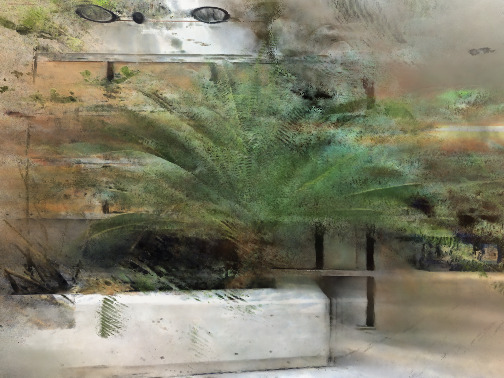}
    \includegraphics[width=0.19\textwidth]{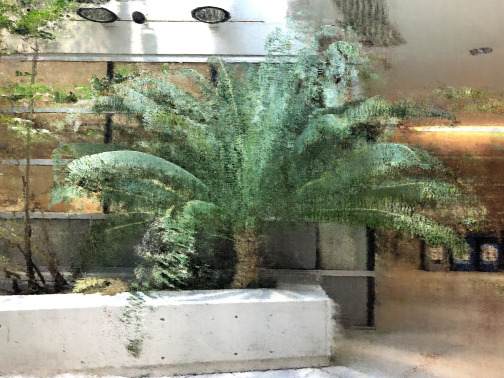}
    \includegraphics[width=0.19\textwidth]{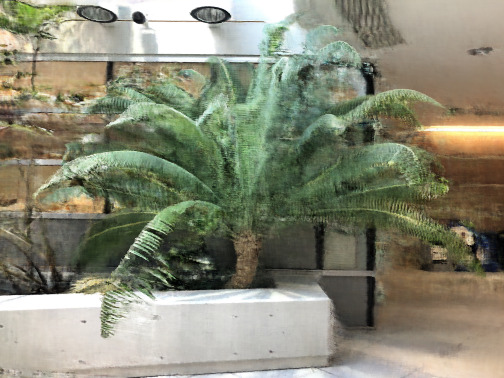}
    \includegraphics[width=0.19\textwidth]{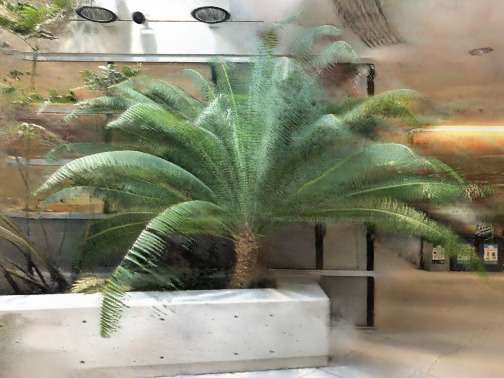}
    \includegraphics[width=0.19\textwidth]{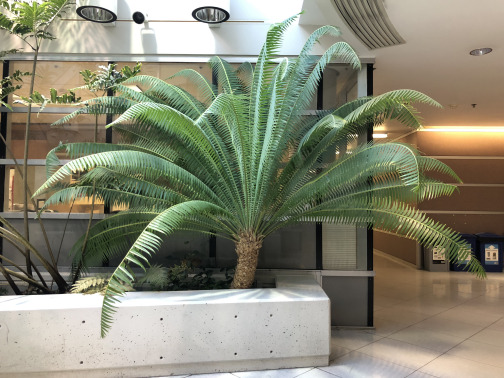}
    \\
    \includegraphics[width=0.19\textwidth]{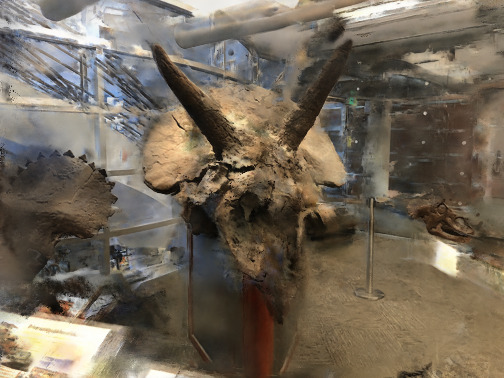}
    \includegraphics[width=0.19\textwidth]{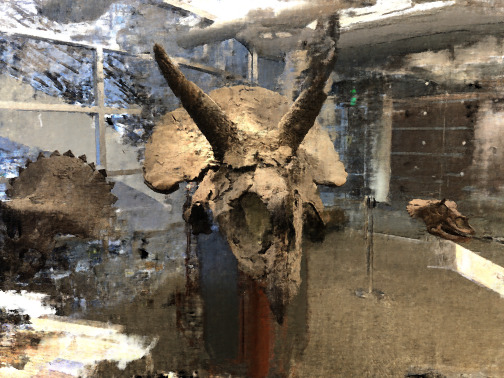}
    \includegraphics[width=0.19\textwidth]{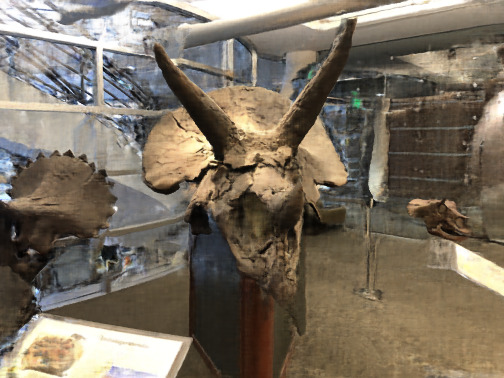}
    \includegraphics[width=0.19\textwidth]{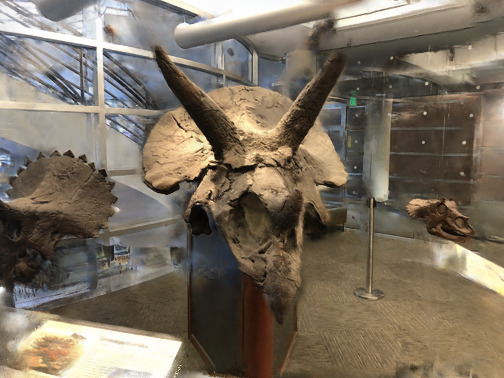}
    \includegraphics[width=0.19\textwidth]{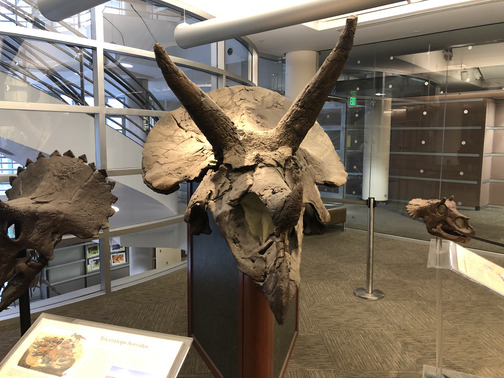}
    \\
    \includegraphics[width=0.19\textwidth]{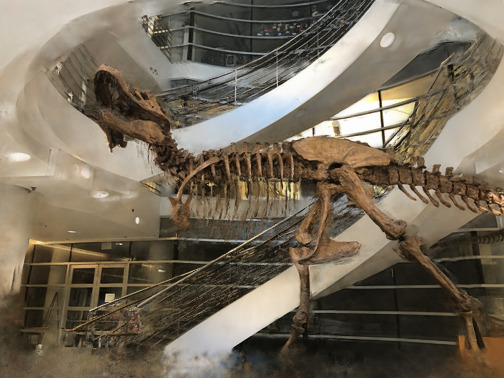}
    \includegraphics[width=0.19\textwidth]{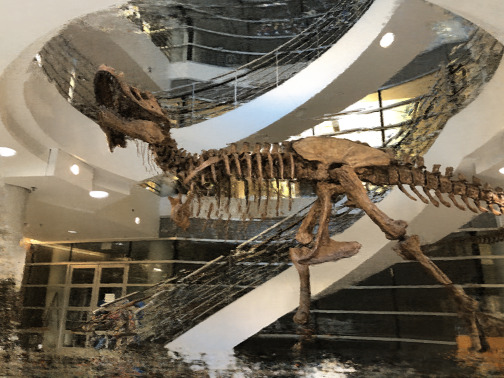}
    \includegraphics[width=0.19\textwidth]{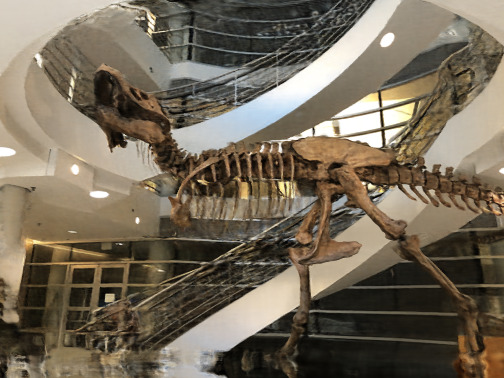}
    \includegraphics[width=0.19\textwidth]{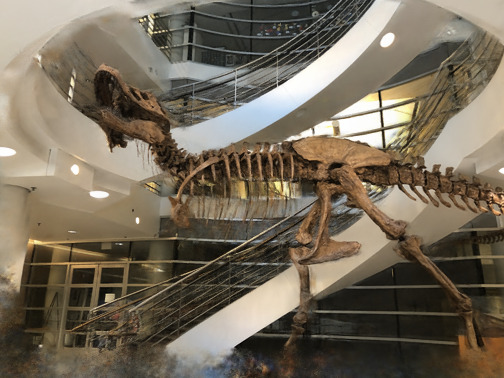}
    \includegraphics[width=0.19\textwidth]{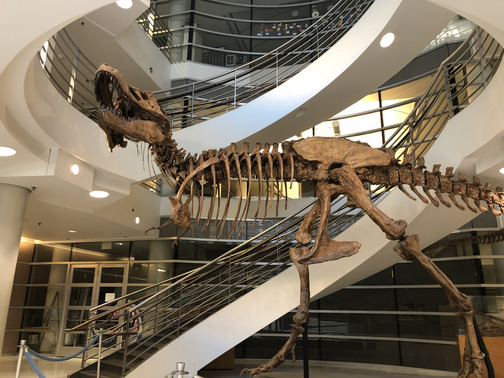}
    \caption{Comparison of our CombiNeRF against RegNeRF, FreeNeRF and Vanilla NeRF on Fern, Horns and Trex scenarios with 3-view setting.} 
    \label{fig:figure1}
\vspace{-3mm}
\end{figure*}
\subsection{Comparison}
We compare the performance achieved by CombiNeRF against the SOTA methods in the few-shot scenario. In LLFF we consider RegNeRF \cite{Reg-NeRF}, FreeNeRF \cite{Free-NeRF} and DiffusioNeRF \cite{DiffusioNeRF}, while in NeRF-Synthetic we take into account DietNeRF \cite{Diet-NeRF} and FreeNeRF \cite{Free-NeRF}. All quantitative evaluation results for the other methods, along with images showing the qualitative results, are taken from the respective papers. If some results are not present, it means that the related paper has not reported quantitative or qualitative results on the related dataset.\\\\
\textbf{LLFF Dataset.}
\begin{figure}[t] \centering
    \makebox[0.15\textwidth]{FreeNeRF}
    \makebox[0.15\textwidth]{CombiNeRF}
    \makebox[0.15\textwidth]{Ground Truth}
    \\
    \includegraphics[width=0.15\textwidth]{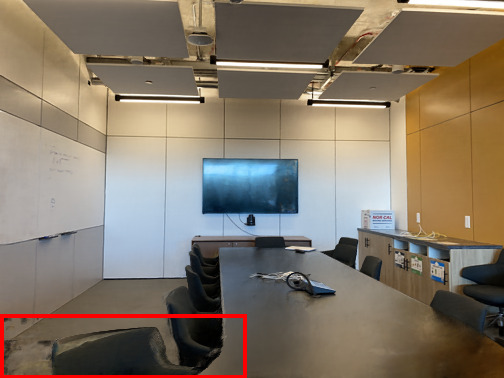}
    \includegraphics[width=0.15\textwidth]{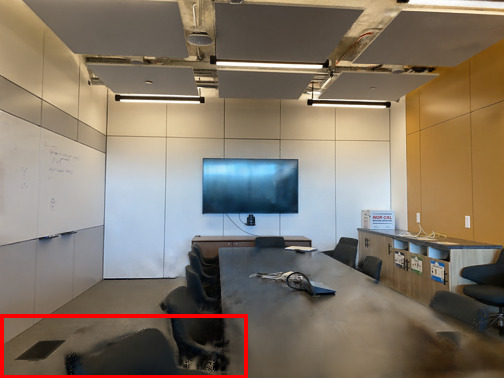}
    \includegraphics[width=0.15\textwidth]{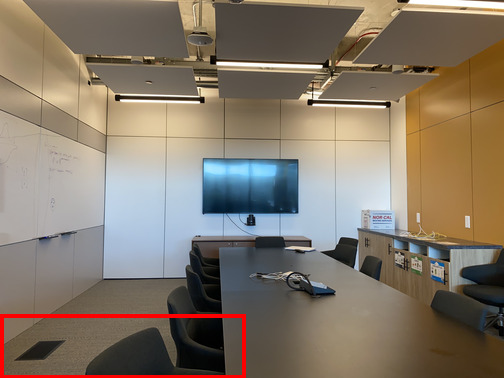}
    \\
    \makebox[0.15\textwidth]{%
        \includegraphics[width=0.1\textwidth]{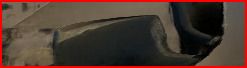}}
    \makebox[0.15\textwidth]{%
        \includegraphics[width=0.1\textwidth]{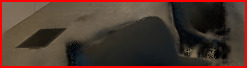}}
    \makebox[0.15\textwidth]{%
        \includegraphics[width=0.1\textwidth]{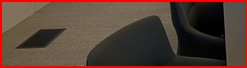}}
    \\[-0.1em]
    \makebox[\columnwidth]{\small (a) 9-view Room scene} 
    \\[0.5em]
    \includegraphics[width=0.15\textwidth]{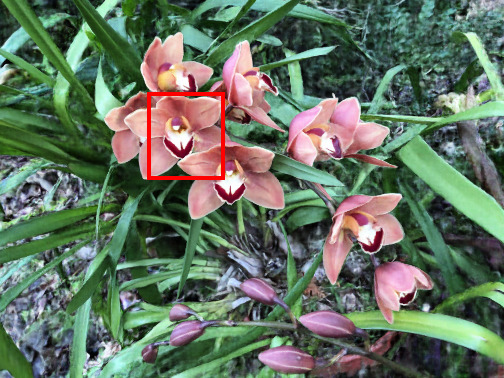}
    \includegraphics[width=0.15\textwidth]{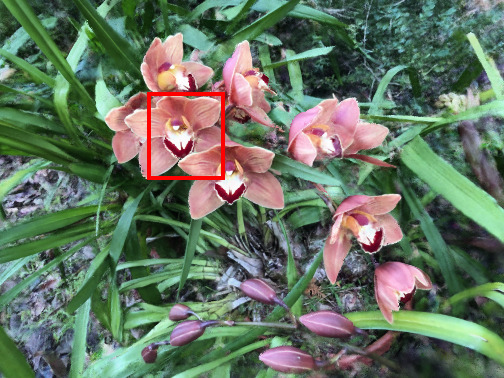}
    \includegraphics[width=0.15\textwidth]{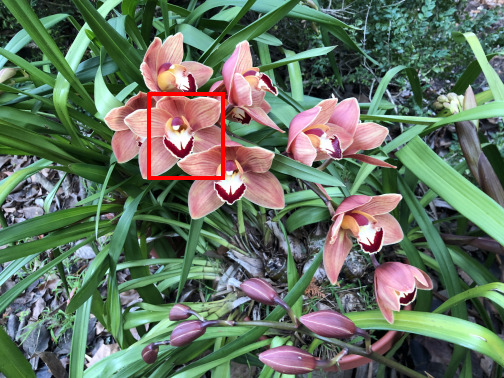}
    \\
    \makebox[0.15\textwidth]{%
        \includegraphics[width=0.08\textwidth]{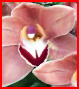}}
    \makebox[0.15\textwidth]{%
        \includegraphics[width=0.08\textwidth]{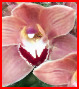}}
    \makebox[0.15\textwidth]{%
        \includegraphics[width=0.08\textwidth]{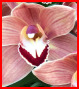}}
    \\[-0.1em]
    \makebox[\columnwidth]{\small (b) 6-view Orchids scene.} 
    \\[0.5em]
    \includegraphics[width=0.15\textwidth]{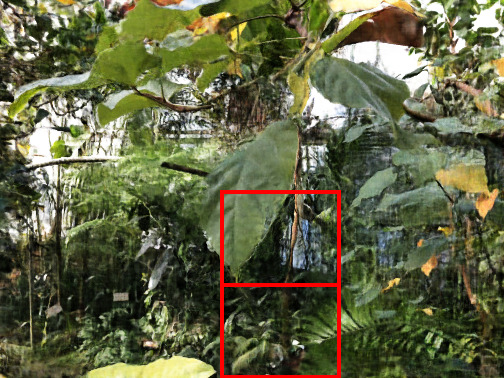}
    \includegraphics[width=0.15\textwidth]{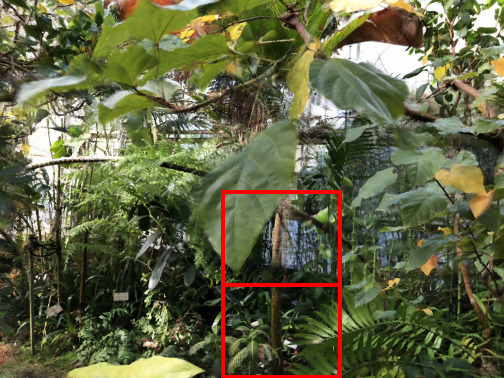}
    \includegraphics[width=0.15\textwidth]{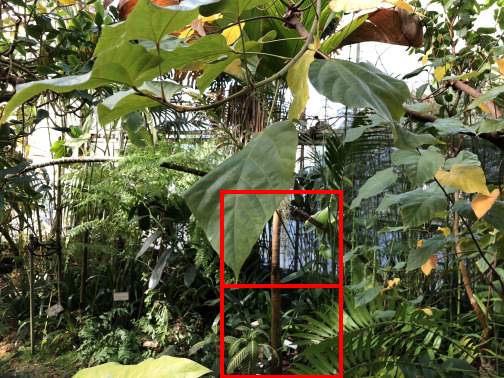}
    \\
    \makebox[0.15\textwidth]{%
        \includegraphics[width=0.07\textwidth]{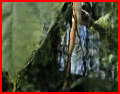}
        \hspace{0.2pt}
        \includegraphics[width=0.07\textwidth]{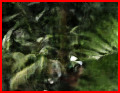}}
    \makebox[0.15\textwidth]{%
        \includegraphics[width=0.07\textwidth]{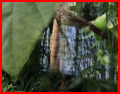}
        \hspace{0.2pt}
        \includegraphics[width=0.07\textwidth]{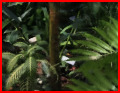}}
    \makebox[0.15\textwidth]{%
        \includegraphics[width=0.07\textwidth]{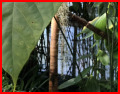}
        \hspace{0.2pt}
        \includegraphics[width=0.07\textwidth]{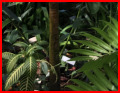}}
    \\[-0.1em]
    \makebox[\columnwidth]{\small (c) 3-view Leaves scene.} 
    \caption{In-depth comparison of CombiNeRF against FreeNeRF on some LLFF scenes with 3/6/9 input views.} 
    \label{fig:figure4}
\vspace{-2mm}
\end{figure}
This dataset is a collection of complex real-world scenes where, from our experiments, we observed that Vanilla NeRF heavily overfits the training images when only a few of them are used as input. 
\cref{tab:table8} shows quantitative results under 3/6/9 training images. For 3 and 9 views, CombiNeRF outperforms the other methods in all the metrics. With 6 views, DiffusioNeRF achieves a better LPIPS score, however, CombiNeRF scores higher on average. Interestingly, with more views (6 or 9), Vanilla NeRF approaches and sometimes outperforms all other methods except CombiNeRF, while its performance heavily degrades with a reduced number of views. CombiNeRF, on the other hand, shows cutting-edge and consistent performance regardless of the number of views.

Qualitative results on LLFF are shown in \cref{fig:figure1}. CombiNeRF is able to reconstruct better high-frequency details maintaining the geometry of the scene. Other methods generate lots of artifacts, like in the "Fern" scene, and they struggle to reconstruct the background. In the "Horns" scene, the environment is noisy in RegNeRF and FreeNeRF while it becomes sharper in CombiNeRF. We also notice that some fine details like the stair handrail in the "Trex" scene are preserved.
In \cref{fig:figure4} we additionally compare CombiNeRF against FreeNeRF. In "Room" with the 9-view setting, the near part of the table gets deformed while in CombiNeRF some details on the floor are still visible. In "Orchids" with the 6-view setting, CombiNeRF is able to render more high-frequency details, for example the stripes in the orchid petals. In "Leaves" with the 3-view setting, the central leaf rendered in FreeNeRF contains several artifacts and, in general, even in this case, CombiNeRF is able to reconstruct better high-frequency details.\\\\
\textbf{NeRF-Synthetic Dataset.}
\begin{table}[t]
\resizebox{\columnwidth}{!}{
\rowcolors{2}{LightCyan}{White}
\begin{tabular}{c | c c c}
\toprule
NeRF-Synthetic 8-views & PSNR $\uparrow$ & SSIM $\uparrow$ & LPIPS $\downarrow$ \\
\midrule
Vanilla NeRF & 22.335 & 0.845 & 0.144    \\
DietNeRF & 23.147 & 0.866 & 0.109 \\
DietNeRF, $\mathcal{L}_{MSE}$ ft & 23.591 & 0.874 & 0.097    \\
FreeNeRF & 24.259 & \textbf{0.883} & 0.098    \\
\midrule
CombiNeRF & \textbf{24.394} & \textbf{0.883} & \textbf{0.088}    \\
\bottomrule
\end{tabular}
}
\caption{NeRF SOTA comparison on NeRF-Synthetic dataset with 8 input views. "$\mathcal{L}_{MSE}$ ft" is the fine-tuned version of DietNeRF.}
\label{tab:table7}
\vspace{-3mm}
\end{table}
\begin{figure}[t] \centering
    \makebox[0.105\textwidth]{Vanilla NeRF}
    \makebox[0.105\textwidth]{}
    \makebox[0.105\textwidth]{CombiNeRF}
    \makebox[0.105\textwidth]{}
    \\
    \includegraphics[width=0.105\textwidth]{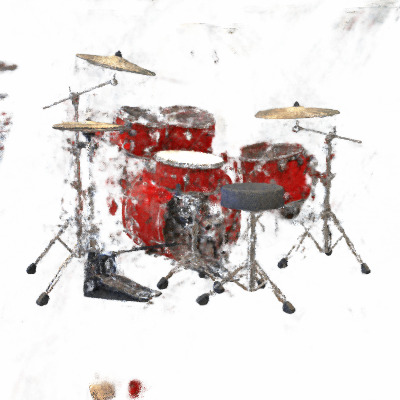}
    \includegraphics[width=0.105\textwidth]{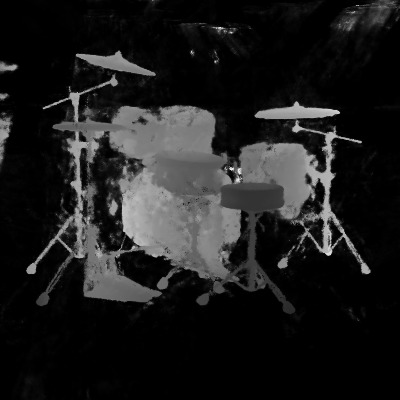}
    \includegraphics[width=0.105\textwidth]{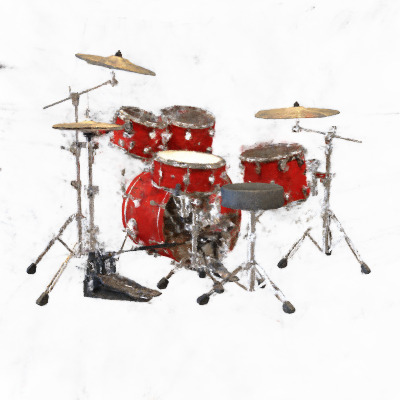}
    \includegraphics[width=0.105\textwidth]{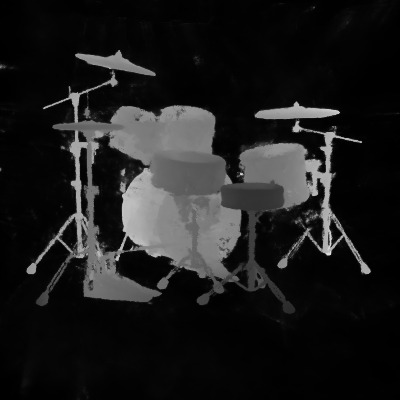}
    \\
    \includegraphics[width=0.105\textwidth]{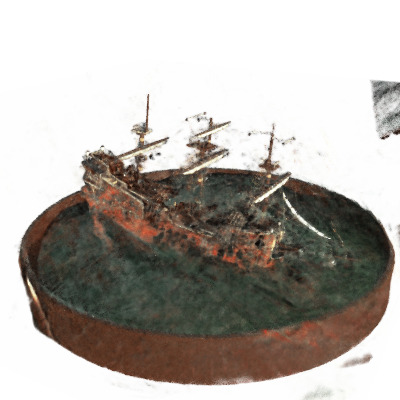}
    \includegraphics[width=0.105\textwidth]{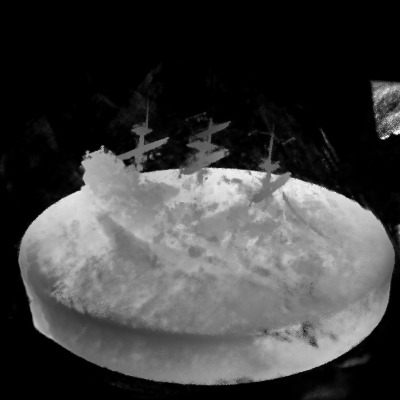}
    \includegraphics[width=0.105\textwidth]{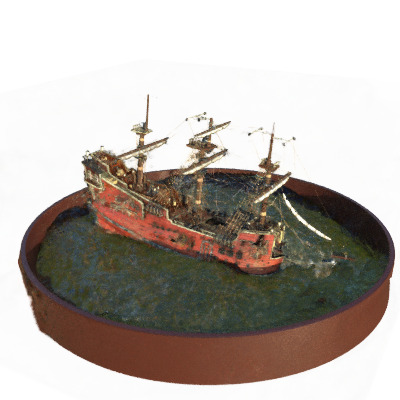}
    \includegraphics[width=0.105\textwidth]{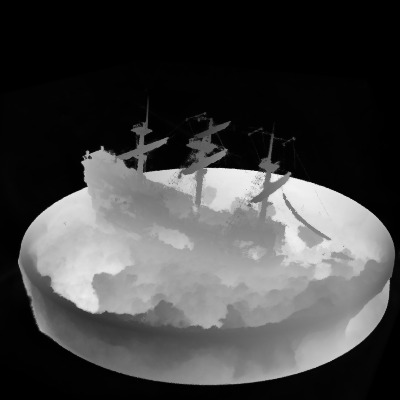}
    \\
    \includegraphics[width=0.105\textwidth]{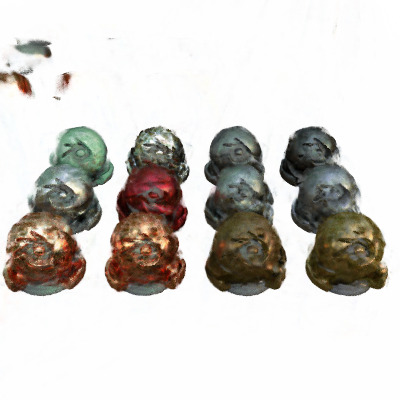}
    \includegraphics[width=0.105\textwidth]{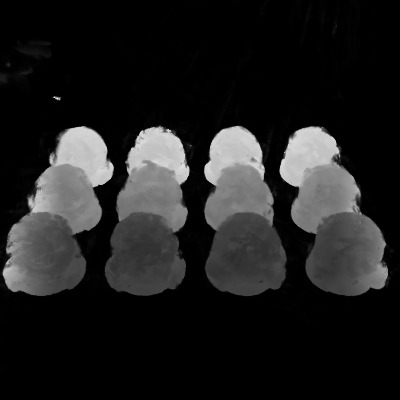}
    \includegraphics[width=0.105\textwidth]{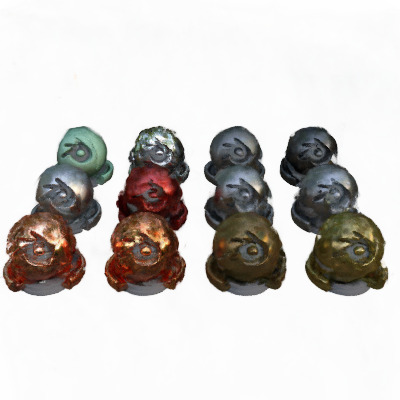}
    \includegraphics[width=0.105\textwidth]{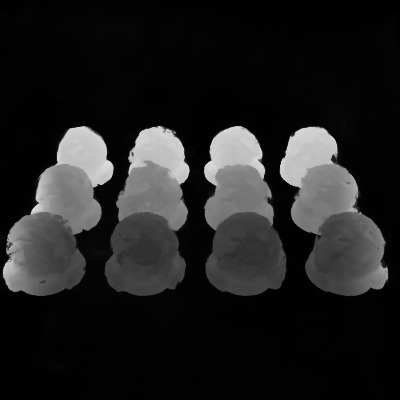}
    \\
    \includegraphics[width=0.105\textwidth]{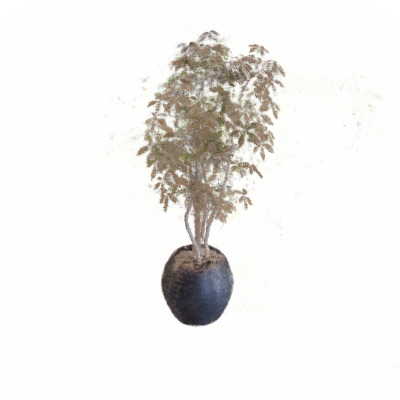}
    \includegraphics[width=0.105\textwidth]{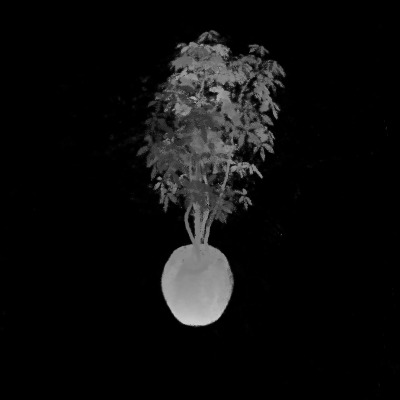}
    \includegraphics[width=0.105\textwidth]{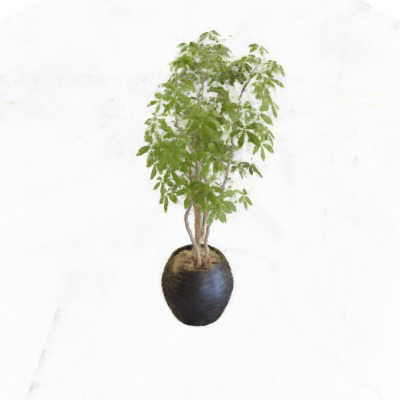}
    \includegraphics[width=0.105\textwidth]{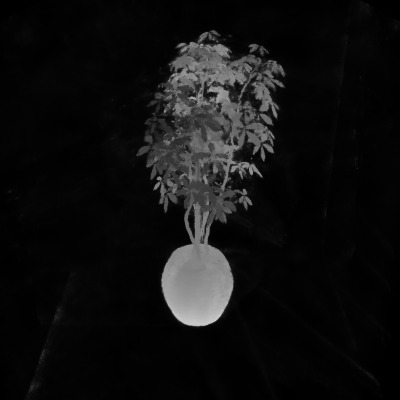}
    \\
    \includegraphics[width=0.105\textwidth]{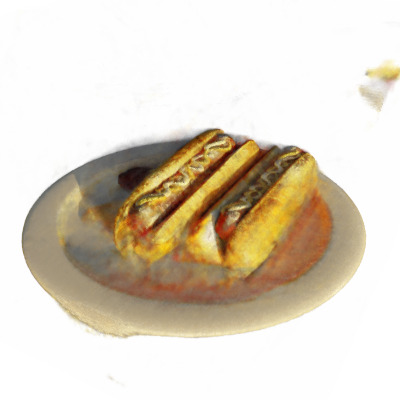}
    \includegraphics[width=0.105\textwidth]{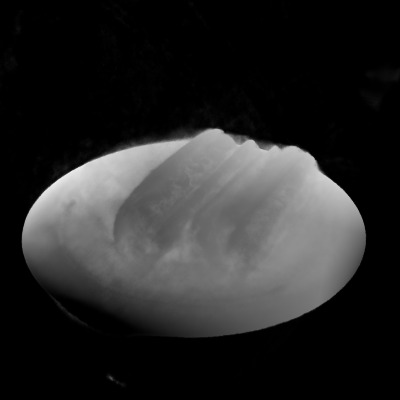}
    \includegraphics[width=0.105\textwidth]{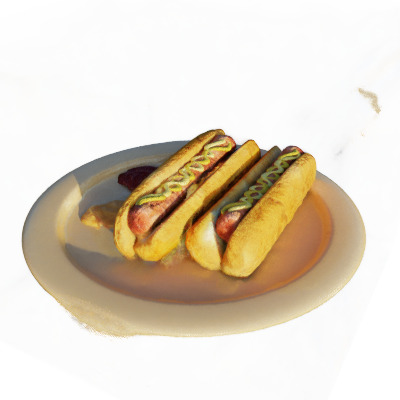}
    \includegraphics[width=0.105\textwidth]{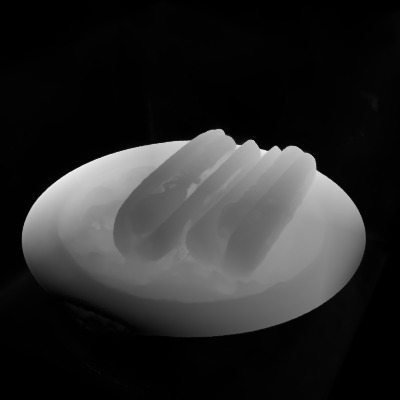}
    \caption{Comparison of CombiNeRF against the Vanilla NeRF method in Drums, Ship, Materials, Ficus and Hotdog scenarios.} 
    \label{fig:figure5}
\vspace{-3mm}
\end{figure}
This dataset contains synthetic renderings of objects exhibiting high-frequency geometric details and reflective effects, thus making this dataset particularly challenging.
In \cref{tab:table7} we show the quantitative results of CombiNeRF and the compared methods. We can see that CombiNeRF outperforms the other SOTA methods on the PSNR and LPIPS metrics while achieving the same result on the SSIM. CombiNeRF is the overall best-performing method also on the NeRF-Synthetic dataset, performing better on scenes like "Lego", "Mic", and "Ficus" which exhibit complex geometries, while "Materials" represents the most challenging scenario due to the presence of strong view-dependent reflection.

In \cref{fig:figure5} we show qualitative results of CombiNeRF in different scenes. The reconstructed depth is more consistent, presenting fewer artifacts. Rendered images are far less noisy with respect to the ones rendered by the Vanilla NeRF implementation. Colors are better preserved, as can be seen in the "Hotdog" and "Ficus" scenes where the color of the leaves is more realistic and closer to the ground truth.

\subsection{Ablation Study}
In this section, we assess the contribution of each of the regularization techniques used in CombiNeRF, thus highlighting their importance.
Instead of showing all the possible combinations, we empirically select only the more relevant regularization techniques.
We conducted an ablation study of our CombiNeRF in the 3-view and 8-view few-shot settings for the LLFF and NeRF-Synthetic datasets respectively. For this ablation study, we used the same parameters of the experiments previously described.\\\\
\textbf{LLFF Dataset.}
\begin{table}[t]
\resizebox{\columnwidth}{!}{
\rowcolors{2}{LightCyan}{White}
\begin{tabular}{c | c c c | c}
\toprule
LLFF 3-views & PSNR $\uparrow$ & SSIM $\uparrow$ & LPIPS $\downarrow$ & Average $\downarrow$  \\
\midrule
Vanilla NeRF   &  17.71 & 0.544 & 0.303 & 0.155 \\
L & 18.74 & 0.608 & 0.249 & 0.130     \\
$\mathcal{L}_{dist}$+$\mathcal{L}_{fg} $ & 17.93 & 0.554 & 0.288 & 0.151     \\
$\mathcal{L}_{dist}$+$\mathcal{L}_{fg} (S_{patch=4})$ & 19.10 & 0.608 & 0.249 & 0.128     \\
$\mathcal{L}_{dist}$+$\mathcal{L}_{fg}$+$\mathcal{L}_{ds}$ & 19.50 & 0.626 & 0.237 & 0.122     \\
$\mathcal{L}_{dist}$+$\mathcal{L}_{fg}$+$\mathcal{L}_{ds}$+L & 20.09 & 0.674 & 0.198 & 0.105     \\
$\mathcal{L}_{dist}$+$\mathcal{L}_{fg}$+$\mathcal{L}_{ds}$+L+EM & 20.27 & 0.683 & 0.194 & 0.103     \\
CombiNeRF$^{\sharp}$ & 19.52 & 0.637 & 0.236 & 0.119     \\
\midrule
CombiNeRF  &  \textbf{20.37} & \textbf{0.686} & \textbf{0.191} & \textbf{0.101}  \\
\bottomrule
\end{tabular}
}
\caption{Ablation on LLFF dataset with 3-view setting. Lipschitz and Encoding Mask are named L and EM respectively. 
When using $\mathcal{L}_{ds}$, we set the patch size $S_{patch}=4$ by default.
We call CombiNeRF$^{\sharp}$ our method using Lipschitz only in the color network.
}
\label{tab:table6}
\vspace{-2mm}
\end{table}
\begin{figure}[t] \centering
    \makebox[0.11\textwidth]{CombiNeRF$^{\sharp}$}
    \makebox[0.11\textwidth]{}
    \makebox[0.11\textwidth]{CombiNeRF}
    \makebox[0.11\textwidth]{}
    \\
    \includegraphics[width=0.11\textwidth]{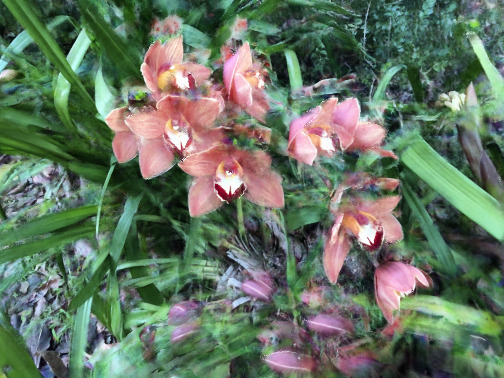}
    \includegraphics[width=0.11\textwidth]{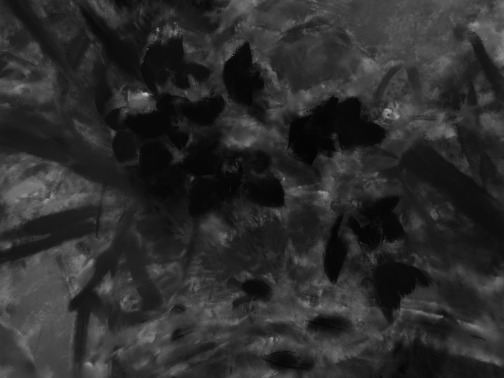}
    \includegraphics[width=0.11\textwidth]{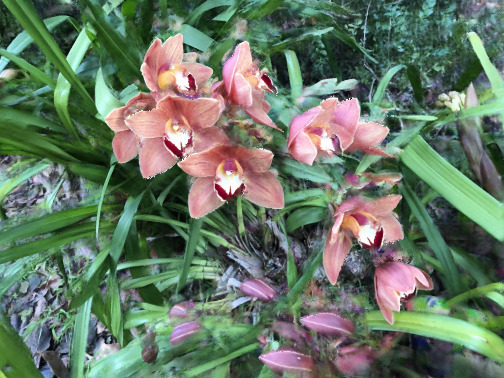}
    \includegraphics[width=0.11\textwidth]{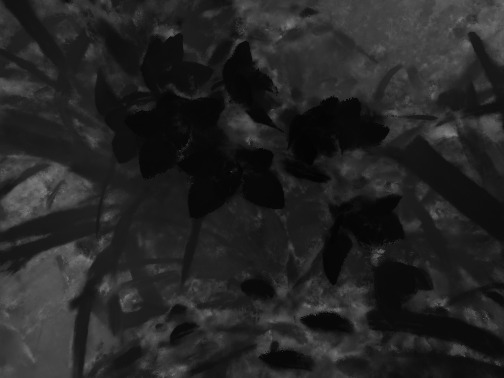}
    \\
    \includegraphics[width=0.11\textwidth]{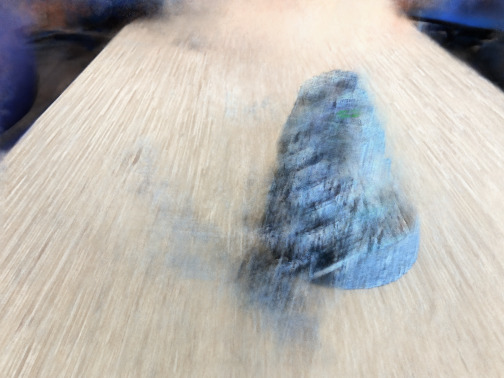}
    \includegraphics[width=0.11\textwidth]{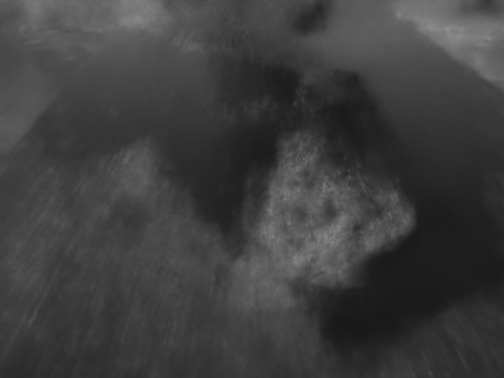}
    \includegraphics[width=0.11\textwidth]{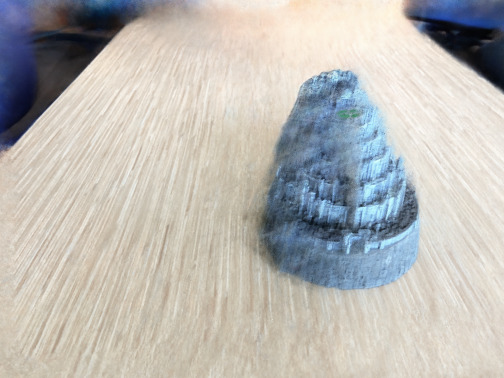}
    \includegraphics[width=0.11\textwidth]{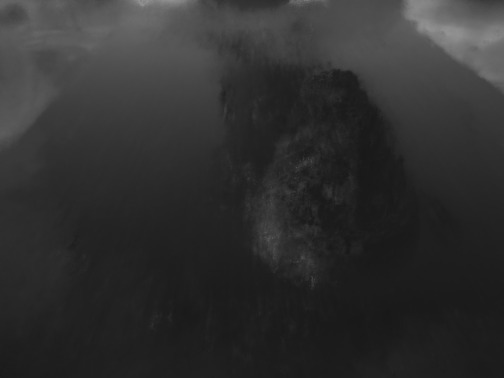}
    \\
    \includegraphics[width=0.11\textwidth]{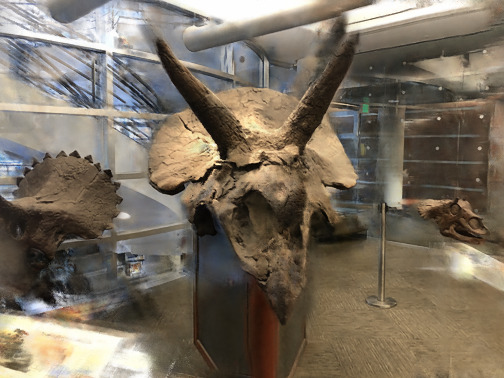}
    \includegraphics[width=0.11\textwidth]{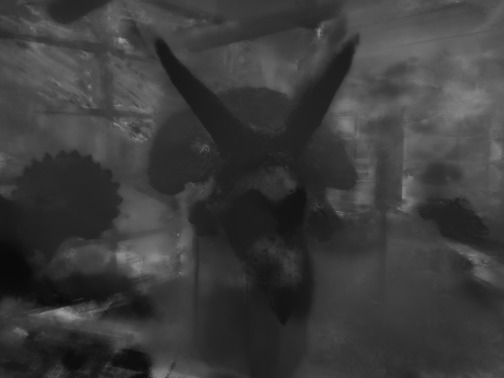}
    \includegraphics[width=0.11\textwidth]{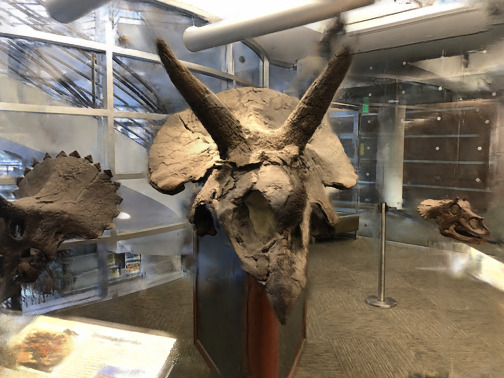}
    \includegraphics[width=0.11\textwidth]{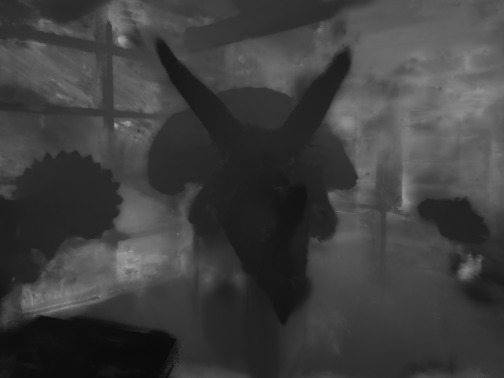}
    \caption{Qualitative result on the ablation study of LLFF with 3-view setting. We compare our CombiNeRF with CombiNeRF$^{\sharp}$ (our method using Lipschitz only in the color network).} 
    \label{fig:figure2}
\vspace{-2mm}
\end{figure}
In \cref{tab:table6} we show the ablation on the LLFF dataset with the 3-view setting. Starting from the Vanilla NeRF solution, we notice that each contribution in CombiNeRF brings an increase in performance, with Lipschitz regularization and $\mathcal{L}_{dist}+\mathcal{L}_{fg}$ losses playing a crucial role in improving the overall performance. Applying Lipschitz regularization on both density and color networks also allows to further improve the quality of the obtained results. 
From the same table, we can observe that when using the term losses $\mathcal{L}_{dist}+\mathcal{L}_{fg}$ greatly benefit from sampling patches of rays instead of single rays.

In \cref{fig:figure2} we show qualitative results of CombiNeRF using Lipschitz regularization in both sigma and color network against the only color network. The latter tentative approach is named CombiNeRF$^{\sharp}$ in the figures and tables.
CombiNeRF is able to model in a more coherent way the geometry of the scenarios, removing also the blurry effects which characterize the renderings of the implementation using only Lipschitz regularization in the color network.\\\\
\textbf{NeRF-Synthetic Dataset.}
\begin{table}[t]
\resizebox{\columnwidth}{!}{
\rowcolors{2}{LightCyan}{White}
\begin{tabular}{c | c c c | c}
\toprule
NeRF-Synthetic 8-views & PSNR $\uparrow$ & SSIM $\uparrow$ & LPIPS $\downarrow$ & Average $\downarrow$  \\
\midrule
Vanilla NeRF & 22.34 & 0.845 & 0.144 & 0.073     \\
$\mathcal{L}_{dist}$+$\mathcal{L}_{fg}$+$\mathcal{L}_{ds}$+L & 23.61 & 0.865 & 0.115 & 0.059     \\
$\mathcal{L}_{dist}$+$\mathcal{L}_{fg}$+$\mathcal{L}_{ds}$+$\mathcal{L}_{KL}^{\dagger}$+L & 22.93 & 0.863 & 0.126 & 0.064   \\
$\mathcal{L}_{dist}$+$\mathcal{L}_{fg}$+$\mathcal{L}_{ds}$+L+EM & 24.04 & 0.871 & 0.105 & 0.056   \\
CombiNeRF$^{\dagger}$ & 24.36 & \textbf{0.883} & \textbf{0.088} & \textbf{0.050}   \\
\midrule
CombiNeRF & \textbf{24.39} & \textbf{0.883} & \textbf{0.088} & 0.051   \\
\bottomrule
\end{tabular}
}
\caption{Ablation on NeRF-Synthetic dataset with 8-view setting. Lipschitz is defined as L and Encoding Mask is defined as EM. We call $\mathcal{L}_{KL}^{\dagger}$ the KL-Divergence loss presented by InfoNeRF and we call CombiNeRF$^{\dagger}$ our method using their $\mathcal{L}_{KL}^{\dagger}$.} 
\label{tab:table5}
\vspace{-3mm}
\end{table}
\begin{figure}[t] \centering
    \makebox[0.105\textwidth]{CombiNeRF$^{\dagger}$}
    \makebox[0.105\textwidth]{}
    \makebox[0.105\textwidth]{CombiNeRF}
    \makebox[0.105\textwidth]{}
    \\
    \includegraphics[width=0.105\textwidth]{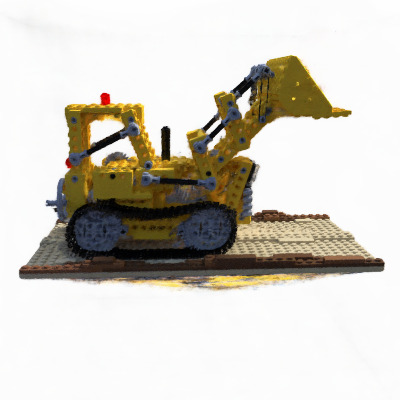}
    \includegraphics[width=0.105\textwidth]{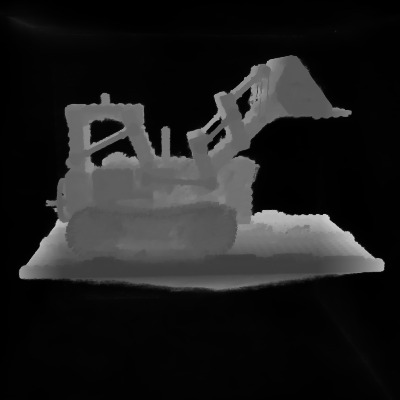}
    \includegraphics[width=0.105\textwidth]{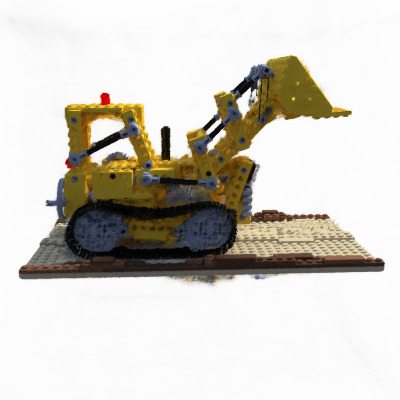}
    \includegraphics[width=0.105\textwidth]{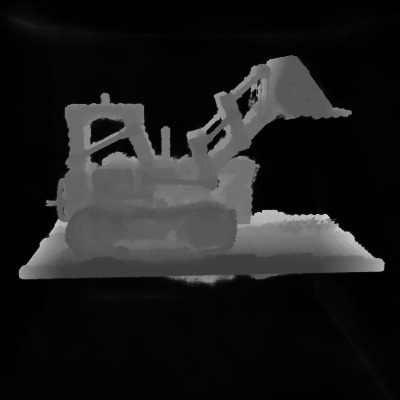}
    \\
    \includegraphics[width=0.105\textwidth]{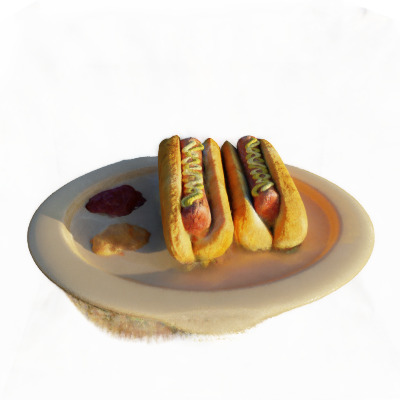}
    \includegraphics[width=0.105\textwidth]{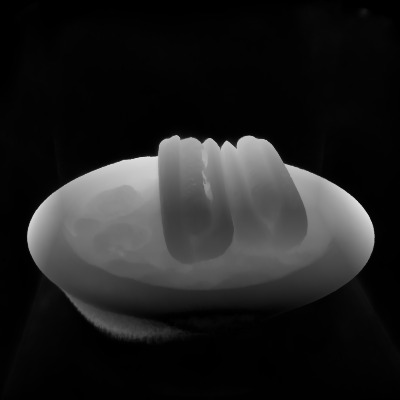}
    \includegraphics[width=0.105\textwidth]{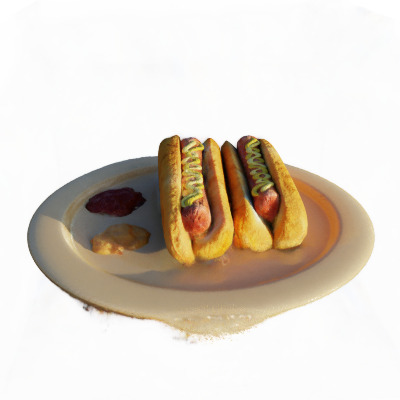}
    \includegraphics[width=0.105\textwidth]{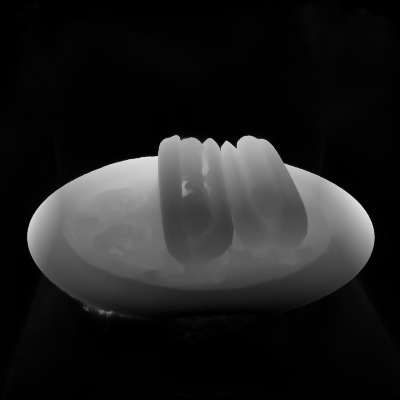}
    \\
    \includegraphics[width=0.105\textwidth]{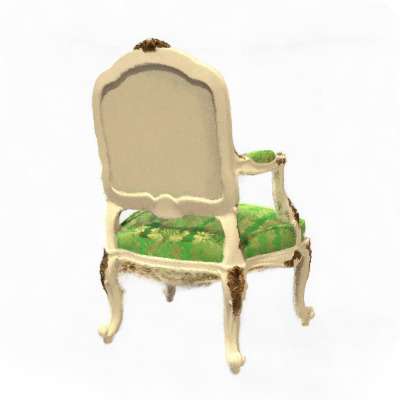}
    \includegraphics[width=0.105\textwidth]{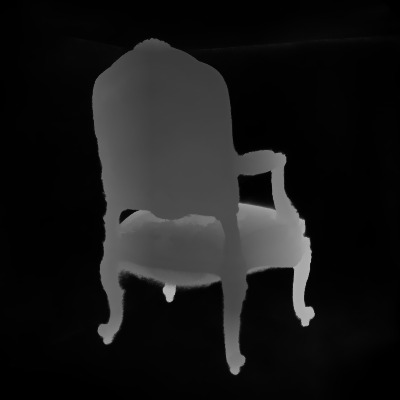}
    \includegraphics[width=0.105\textwidth]{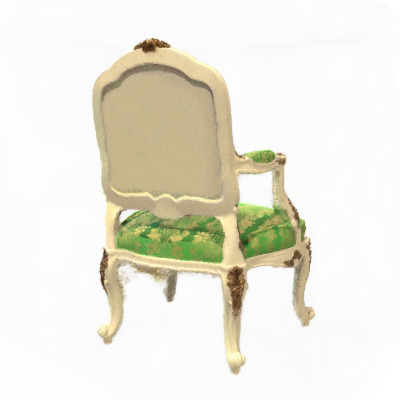}
    \includegraphics[width=0.1105\textwidth]{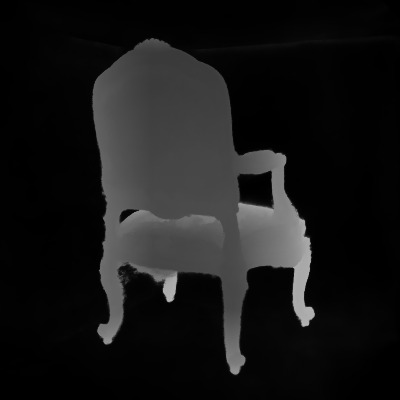}
    \caption{Qualitative result on the ablation study of NeRF-Synthetic with 8-view setting. We compare our CombiNeRF with CombiNeRF$^{\dagger}$ (our method with KL-Divergence as in InfoNeRF).} 
    \label{fig:figure3}
\vspace{-4mm}
\end{figure}
As already done for the LLFF dataset, we also provide an ablation study on the NeRF-Synthetic dataset. In \cref{tab:table5} we show the result of this study with the 8-view setting. In this study, we additionally focused on the $\mathcal{L}_{KL}$ loss. In particular, we both tested the same implementation of the KL-Divergence loss used in \cite{Info-NeRF} (implemented in an intermediate version of CombiNeRF called CombiNeRF$^{\dagger}$) and our modified version (as described in \cref{sec:regularization_losses}) used in the final CombiNeRF.

In \cref{fig:figure3} we show the qualitative results of both the two implementations. While quantitative results remain almost the same (see \cref{tab:table5}), CombiNeRF is able to remove most of the artifacts generated below the objects, thus indicating an improved reconstruction of the final geometries.

\section{Conclusion} 
\label{sec:conclusion}
In this work, we presented CombiNeRF, a combination of regularization techniques in a unified framework. We regularize neighboring rays distributions, we also take into account the single ray distribution and a smoothness term is adopted to regularize near geometries. Besides these additional loss functions, we also consider Lipschitz regularization and an encoding mask to regularize high-frequency components. 

CombiNeRF shows cutting-edge and consistent results in the quality of the reconstructions and it outperforms the SOTA methods in LLFF and NeRF-Synthetic datasets with few-shot settings. The increasing performance of the combination of multiple regularization techniques is validated through an ablation study that highlights the contribution of the CombiNeRF components and the differences with the previous implementations.

Additional techniques like Ray Entropy Loss \cite{Info-NeRF}, Adversarial Perturbation \cite{Aug-NeRF} and Space Annealing \cite{Reg-NeRF} have been taken into account but not considered in CombiNeRF for their apparently poor contribution. Additional details are given in the supplementary material but more experiments are required as future works on these and other techniques to achieve a 
more general purpose NeRF framework.

{
    \small
    \bibliographystyle{ieeenat_fullname}
    \bibliography{main}
}
\clearpage
\setcounter{page}{1}
\maketitlesupplementary

\section{Additional Implementations}
\label{sec:add_loss}
In the following sections, we describe three additional implementations considered during the empirical selection of the best methods used in CombiNeRF. The following implementations were not included in CombiNeRF for their apparently poor contribution with respect to Vanilla NeRF \cite{Instant-NGP}. 
Due to a lack of time for further testing, we were only able to partially evaluate these methods, which will eventually be considered in future CombiNeRF extensions.
As done for the other considered techniques, we initially tested their performance on the LLFF dataset with the 3-view setting, comparing their results against Vanilla NeRF. For completeness, we provide below a brief summary of these techniques. 

\subsection{Regularization Losses}

\textbf{Ray Entropy Loss}.
The aim is to introduce a sparsity constraint to focus only on the scene of interest. 
This constraint can be achieved by minimizing the entropy of each sampled ray density function, as done in \cite{Info-NeRF}, in order to force the rays to have a few high peaks, i.e. the peaks should only represent the hit surfaces.

To minimize the entropy of a ray $\textbf{r}$, we start defining the normalized ray density $q(\textbf{r})$ as:
\begin{equation}
    q(\textbf{r}_i) = \frac{\alpha_i}{\sum_j\alpha_j} = \frac{1-\exp(-\sigma_i\delta_i)}{\sum_j 1-\exp(-\sigma_j\delta_j)},  
\end{equation}
where $\textbf{r}_i$ ($i=1,...,N$) is a sampled point in the ray, $\sigma_i$ is the observed density at $\textbf{r}_i$, $\delta_i$ is a sampling interval around $\textbf{r}_i$ and $\alpha_i$ is the opacity at $\textbf{r}_i$.

Starting from the Shannon Entropy, we can define the entropy of a discrete ray density function:
\begin{equation}
    H(\textbf{r})= -\sum_{i=1}^N q(\textbf{r}_i) \log q(\textbf{r}_i),
    \label{eq:entropy_ray_density}
\end{equation}
where the calculation of $q(\textbf{r}_i)$, which includes $\sigma_i$ and $\delta_i$, is already given thanks to volume rendering computation.


A mask $M$ is applied to filter the rays in the entropy defined in \cref{eq:entropy_ray_density} with the aim of discarding too low-density rays not intersecting or hitting any object in the scene:

\begin{equation}
    M(\textbf{r}) = 
    \begin{cases}
        1 \;\;\; \textnormal{if} \;\; Q(\textbf{r}) > \epsilon   \\
        0\;\;\;  \textnormal{otherwise}
    \end{cases},
\end{equation}
where
\begin{equation}
    Q(\textbf{r}) = \sum_{i=1}^N 1-\exp(-\sigma_i\delta_i) ,
\end{equation}
is the cumulative ray density.

Because unobserved viewpoints can help generalization, for the entropy loss computation both rays from training views and from unseen images can be considered, whose sets are denoted respectively by $R_t$ and $R_u$. The final entropy loss is defined as:
\begin{equation}
    \mathcal{L}_{entr} = \lambda_{entr} \cdot \frac{1}{|R_t|+|R_u|} \; \sum_{\textbf{r} \in R_t\cup R_u} M(\textbf{r}) \odot H(\textbf{r}),
\end{equation}
where $\odot$ denotes the element-wise multiplication and $\lambda_{entr}$ defines the contribution of this loss.\\\\
\textbf{Adversarial Perturbation}.
Previous works demonstrate how neural networks benefit from random or learned data augmentation. Generalization can be achieved by injecting noise during the training phase, as done in \cite{Aug-NeRF}. Here, noise is introduced in the form of worst-case perturbations applied to the input coordinates, features of the network and the pre-rending output of the network. The worst-case perturbation can be formulated as a min-max problem defined as follows:
\begin{equation}
    \min_{\Theta} \max_{\delta} \big\| C^{\dagger}(r | \Theta, \delta) - C^{\ast}(r) \big\|^2_2 ,
\end{equation}
where
\begin{equation}
    \delta = (\delta_p, \delta_f, \delta_r) \in \mathcal{S}_p \times \mathcal{S}_f \times \mathcal{S}_r ,
\end{equation}
and $\delta_p, \delta_f, \delta_r$ are the learned perturbation, $\mathcal{S}_p, \mathcal{S}_f, \mathcal{S}_r$ are the corresponding search range and $C^{\dagger}(r | \Theta, \delta)$ is the perturbed rendered color.

The input coordinates perturbation $\delta_p = (\delta_t, \delta_{xyz}, \delta_{\theta})$ consists of along-ray perturbation $\delta_t\in\mathbb{R}^3$, point position perturbation $\delta_{xyz}\in\mathbb{R}^ 3$ and view-direction perturbation $\delta_{\theta}\in\mathbb{R}^3$ (hence $S_p \subseteq \mathbb{R}^6$). The feature perturbation $\delta_f$ consists of perturbing a $D$-dimensional features vector, hence $ \mathcal{S}_f \subseteq \mathbb{R}^D$. Finally, a pre-rendering output perturbation is applied to RGB color and density values ($\mathcal{S}_r \subseteq \mathbb{R}^4$).

In order to estimate the worst-case perturbation of the injected noise, \cite{Aug-NeRF} proposes a multi-step Projected Gradient Descent (PGD) approach. Given $l_{\inf}$ as the norm ball defining all search spaces $\mathcal{S}_p, \mathcal{S}_f, \mathcal{S}_r$ and the radius $\epsilon$ as the maximum magnitude of the perturbation, a PGD step is defined as:
\begin{equation}
\begin{split}
    \begin{gathered}
    \delta^{(t+1)} = \prod_{\| \delta\|_{\infty} \leq \epsilon} \Big[ \delta^{(t)} + \\
    \alpha \cdot \mbox{sgn}(\big\| C^{\dagger}(r | \Theta, \delta) - C^{\ast}(r) \big\|^2_2) \Big] ,
    \end{gathered}
\end{split}
\end{equation}
where $\alpha$ is the step size, $\prod[\cdot]$ is a projection operator and $\mbox{sgn}(\cdot)$ is the sign function.

Following this procedure, the adversarial perturbation loss is obtained as:
\begin{equation}
    \mathcal{L}_{adv} = \lambda_{adv} \cdot \sum_{r\in\mathcal{R}} \big\| C^{\dagger}(r | \Theta, \delta) - C^{\ast}(r) \big\|^2_2,
\end{equation}
where $\lambda_{adv}$ defines the contribution of this loss.\\\\
\subsection{Space Annealing}
When few images are provided, NeRF could converge to undesired solutions. In this case, high-density values are concentrated close to ray origins. Annealing the sampled scene space over the early iterations of the training \cite{Reg-NeRF} can be a solution to avoid this specific problem: the idea is to restrict the sampling space to an initial smaller region where the scene is centered.

Given the camera's near and far planes $t_n$ and $t_f$, let $t_m$ be a point between them, likely their midpoint, and define the new near and far plane per-iteration as:
\begin{equation}
    \begin{split}
    \begin{gathered}
        t_n(i) = t_m + (t_n-t_m)\eta(i)\\
        t_f(i) = t_m + (t_f-t_m)\eta(i)\\
        \eta(i) = \min(\max(i/N_t,p_s), 1),
    \end{gathered}
    \end{split}
\end{equation}
where $i$ is the current training iteration, the parameter $N_t$ represents the maximum number of iterations in which to perform the space annealing and $p_s$ indicates the starting range. These parameters should be tuned according to the particular scene under consideration.

Annealing the sampled space, especially during the first iteration of training, may allow NeRF to focus on the right region of interest, avoiding degenerate solutions.

\subsection{Results}
\begin{table}[t]
\resizebox{\columnwidth}{!}{
\rowcolors{2}{LightCyan}{White}
\begin{tabular}{c | c c c | c}
\toprule
LLFF 3-views & PSNR $\uparrow$ & SSIM $\uparrow$ & LPIPS $\downarrow$ & Average $\downarrow$  \\
\midrule
Vanilla NeRF   &  17.71 & 0.544 & 0.303 & 0.155 \\
$\mathcal{L}_{adv}$ &   17.75 & 0.541 & 0.300 & 0.155 \\
$\mathcal{L}_{entr}$ &  17.34 & 0.526 & 0.312 & 0.161  \\
Anneal &  18.04 & 0.562 & 0.284 & 0.147  \\
\midrule
CombiNeRF  &  \textbf{20.37} & \textbf{0.686} & \textbf{0.191} & \textbf{0.101}  \\
\bottomrule
\end{tabular}
}
\caption{Quantitative results comparison among the other three regularization techniques on LLFF dataset with 3-view setting.}
\label{tab:table15}
\end{table}
\begin{figure*}[t] \centering
    \makebox[0.01\textwidth]{}
    \makebox[0.16\textwidth]{}
    \makebox[0.16\textwidth]{}
    \makebox[0.16\textwidth]{}
    \makebox[0.16\textwidth]{}
    \makebox[0.16\textwidth]{}
    \makebox[0.16\textwidth]{}
    \\
    \raisebox{0.1\height}{\makebox[0.01\textwidth]{\rotatebox{90}{\makecell{\scriptsize Vanilla NeRF}}}}
    \includegraphics[width=0.16\textwidth]{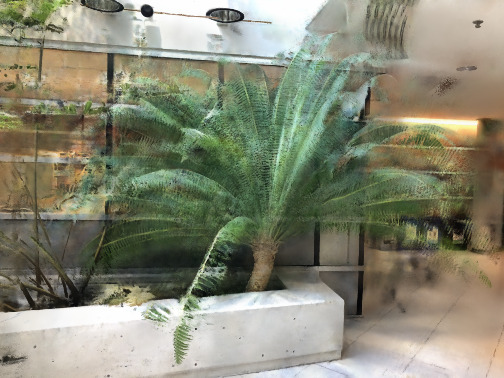}
    \includegraphics[width=0.16\textwidth]{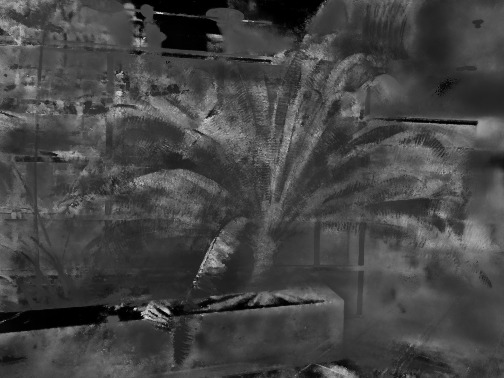}
    \includegraphics[width=0.16\textwidth]{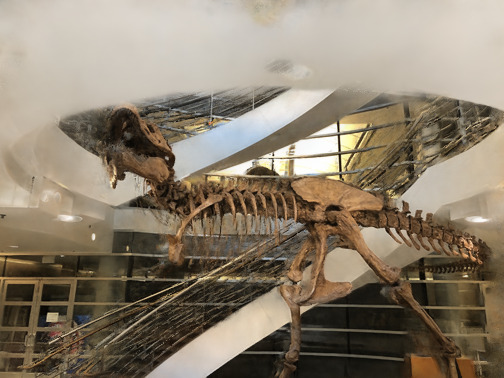}
    \includegraphics[width=0.16\textwidth]{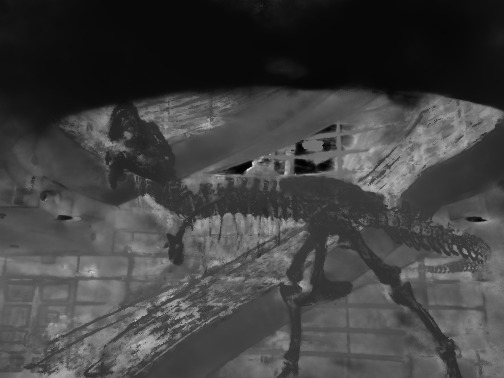}
    \includegraphics[width=0.16\textwidth]{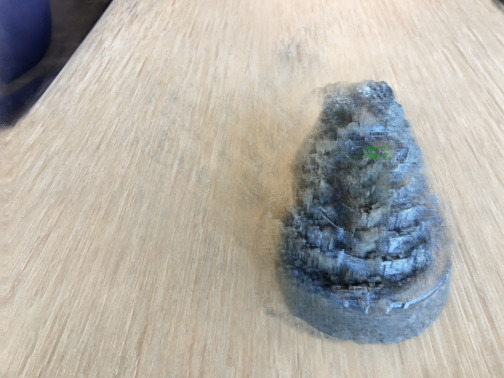}
    \includegraphics[width=0.16\textwidth]{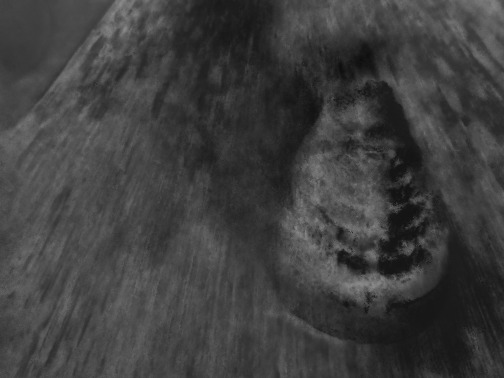}
    \\
    \raisebox{0.1\height}{\makebox[0.01\textwidth]{\rotatebox{90}{\makecell{\scriptsize Anneal}}}}
    \includegraphics[width=0.16\textwidth]{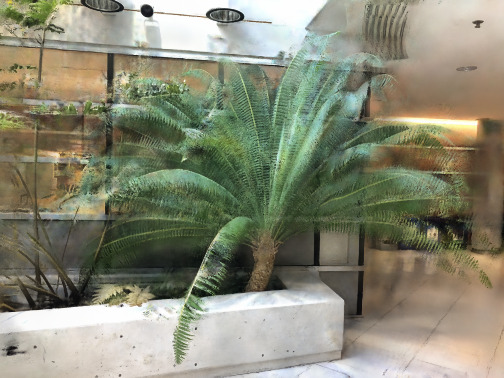}
    \includegraphics[width=0.16\textwidth]{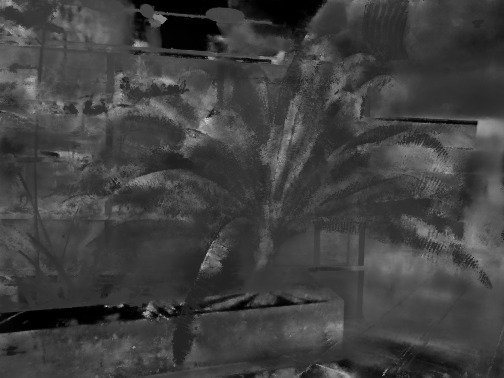}
    \includegraphics[width=0.16\textwidth]{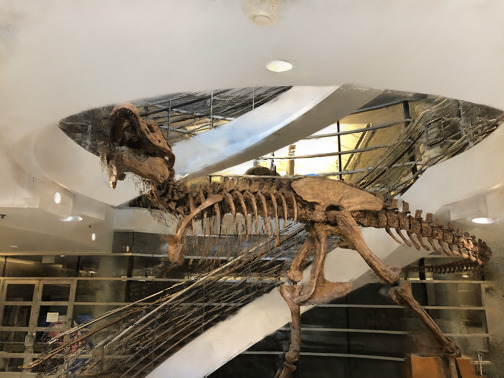}
    \includegraphics[width=0.16\textwidth]{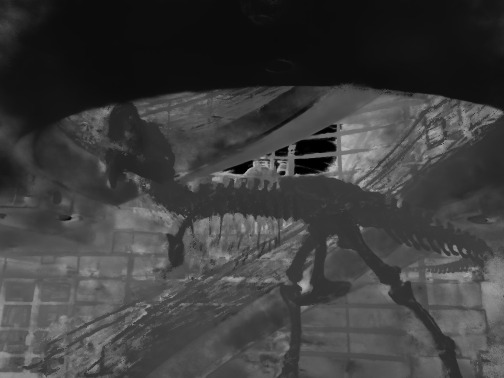}
    \includegraphics[width=0.16\textwidth]{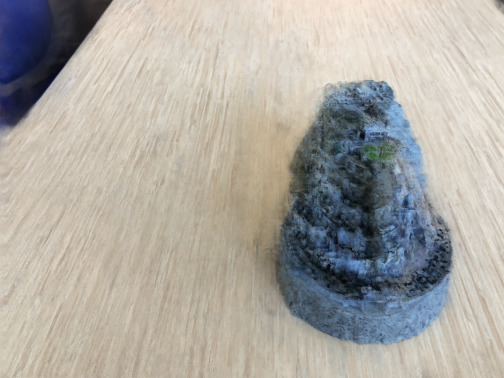}
    \includegraphics[width=0.16\textwidth]{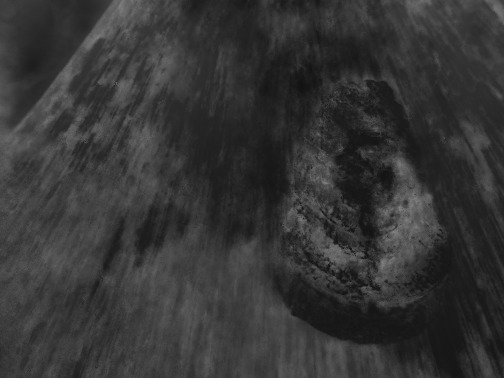}
    \\
    \raisebox{0.1\height}{\makebox[0.01\textwidth]{\rotatebox{90}{\makecell{\scriptsize Perturb}}}}
    \includegraphics[width=0.16\textwidth]{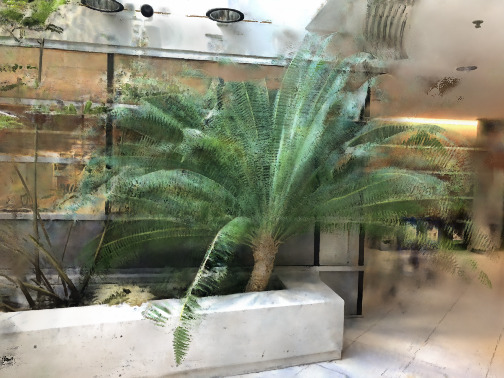}
    \includegraphics[width=0.16\textwidth]{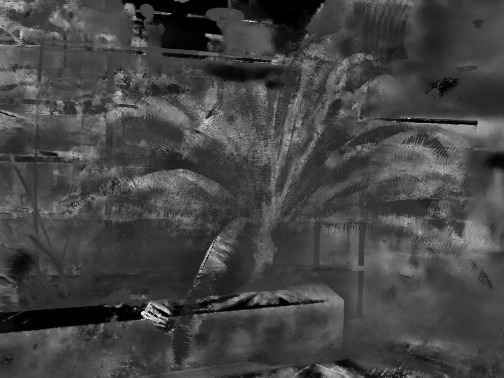}
    \includegraphics[width=0.16\textwidth]{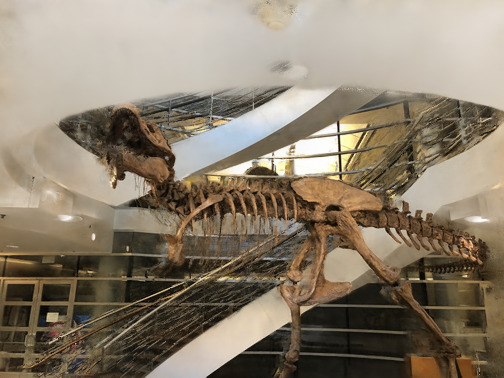}
    \includegraphics[width=0.16\textwidth]{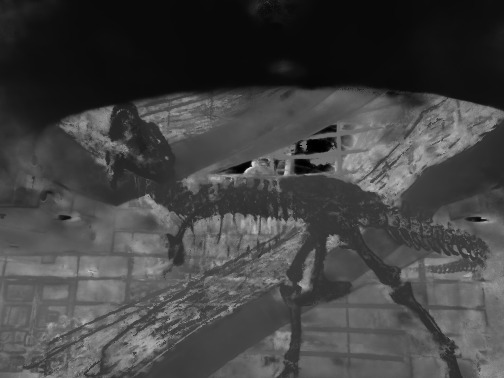}
    \includegraphics[width=0.16\textwidth]{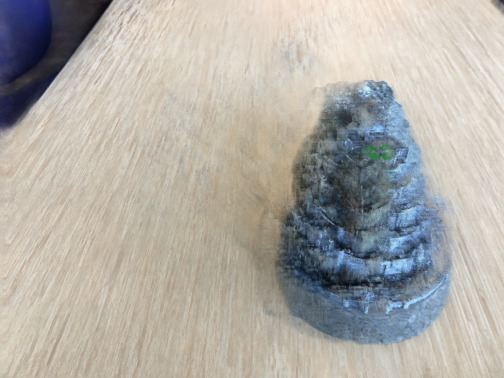}
    \includegraphics[width=0.16\textwidth]{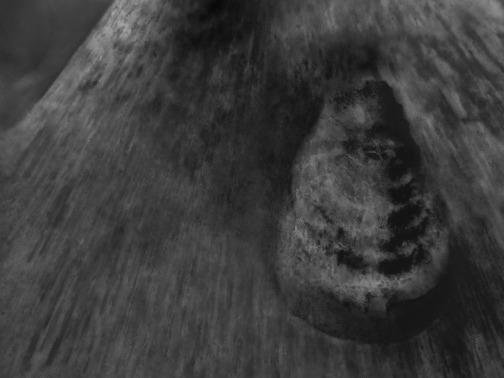}
    \\
    \raisebox{0.1\height}{\makebox[0.01\textwidth]{\rotatebox{90}{\makecell{\scriptsize Entropy}}}}
    \includegraphics[width=0.16\textwidth]{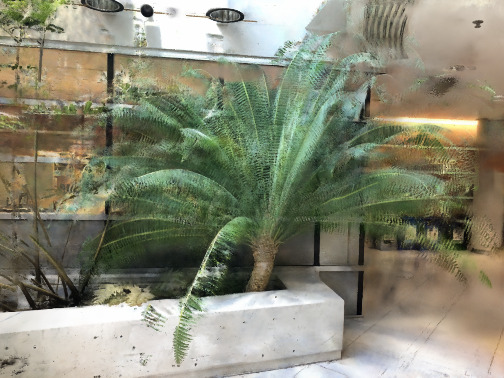}
    \includegraphics[width=0.16\textwidth]{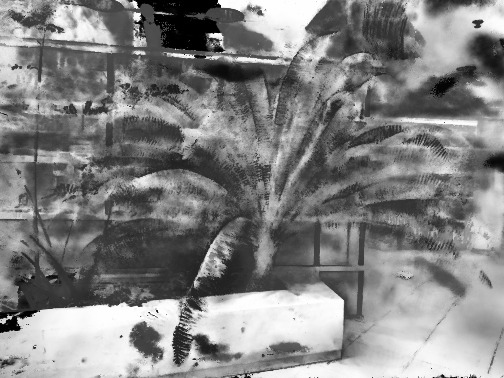}
    \includegraphics[width=0.16\textwidth]{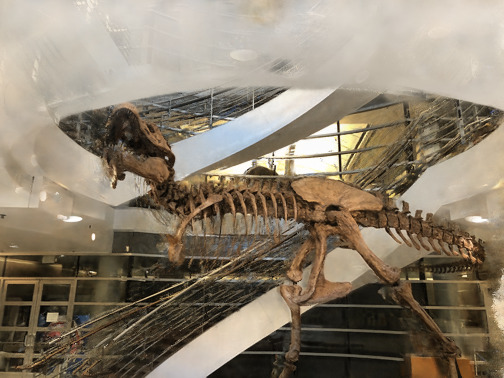}
    \includegraphics[width=0.16\textwidth]{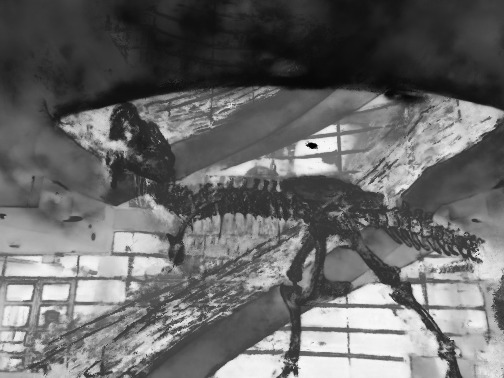}
    \includegraphics[width=0.16\textwidth]{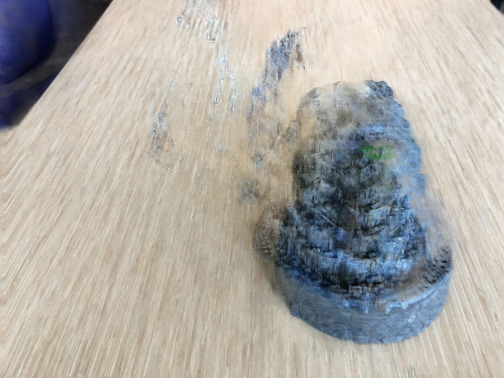}
    \includegraphics[width=0.16\textwidth]{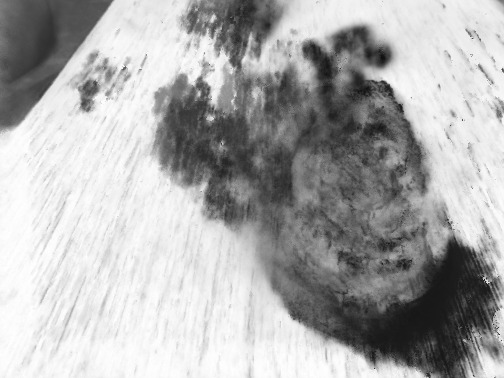}
    \caption{Comparison of Vanilla NeRF against Space Annealing, Adversarial Perturbation and Entropy loss on LLFF with 3-view setting.} 
    \label{fig:figure6}
\end{figure*}
We evaluated the three implementations presented above on the LLFF dataset with a 3-view setting. 
We compared the obtained results with the results obtained by Vanilla NeRF to see if a specific technique should be considered for further experiments or should be revised or better fine-tuned.

In \cref{tab:table15} we show the quantitative results of this study. CombiNeRF remains the best-performing method, outperforming all the other methods by a large margin.  $\mathcal{L}_{entr}$ is not able to improve NeRF generalization and, on the contrary, degrades its performance.
$\mathcal{L}_{adv}$ slightly improves some metrics but it doesn't affect the overall score. Moreover, the time complexity required to perform PDG for worst-case perturbation considerably increases the total computational time. 
Instead, the Space Annealing technique outperforms Vanilla NeRF in all metrics.
This result would lead us to focus on the latter as a promising method to be tested and eventually integrated into CombiNeRF.
However, we argue that the contribution provided by Space Annealing could not offer additional benefits when paired with the distortion loss of \cref{eq:dist}, as both methods try to remove high-density values on points close to the camera.

In \cref{fig:figure6} we show qualitative results of Vanilla NeRF against the other three implementations.
In Space Annealing results, the depths are smoother and the reconstruction quality increases as visible in the white wall and the light on top of "T-rex", in the leaves' details on "Fern" and in the overall object reconstruction of "Fortress". With $\mathcal{L}_{adv}$ is very difficult to see improvements in the resulting renderings, which reflects the quantitative results seen in \cref{tab:table15}. In $\mathcal{L}_{entr}$ results, both depth and RGB images are less accurate, containing more artifacts (e.g., in "Fortress") and noise, particularly visible in the white wall of "T-rex".


\section{Training Details}
For an easier reproduction of the results, we provide the training procedure adopted for the LLFF dataset, which follows \cite{Reg-NeRF}, and for the NeRF-Synthetic dataset, which follows \cite{Diet-NeRF}.

\subsection{LLFF}
In LLFF we use all 8 scenarios ("Fern", "Room", "T-rex", "Flower", "Leaves", "Horns", "Orchids" and "Fortress"). Each view of each scene is 378$\times$504, thus 8$\times$ downsampled with respect to the original resolution. Test views are taken every 8th view and input images are evenly sampled among the remaining ones.
\subsection{NeRF-Synthetic}
In Nerf-Synthetic we use all 8 scenarios ("Lego", "Hotdog", "Mic", "Drums", "Materials", "Ship", "Chair" and "Ficus"). Each view of each scene is 400$\times$400, thus 2$\times$ downsampled with respect to the original resolution. We take 25 test views evenly sampled from the original test set, while training views are chosen according to the following IDs: 2, 16, 26, 55, 73, 75, 86, 93.

\section{Additional Results}
In this section we show additional quantitative and qualitative results of CombiNeRF on LLFF with 3/6/9-views setting and NeRF-Synthetic dataset with 8-view setting. Evaluation results are taken from the respective paper, when reported.
\subsection{Qualitative Results}
\begin{figure*}[t] \centering
    \makebox[0.19\textwidth]{Ground Truth}
    \makebox[0.19\textwidth]{Vanilla NeRF}
    \makebox[0.19\textwidth]{}
    \makebox[0.19\textwidth]{CombiNeRF}
    \makebox[0.19\textwidth]{}
    \\
    \includegraphics[width=0.19\textwidth]{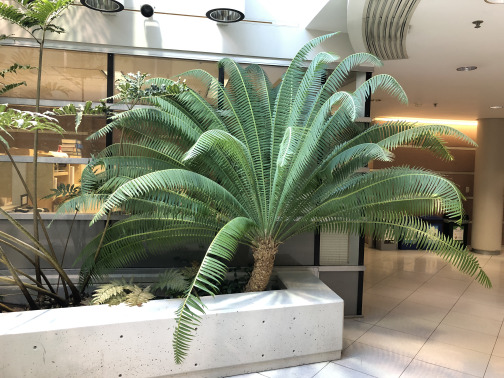}
    \includegraphics[width=0.19\textwidth]{images/Vanilla/fern/ngp_ep3000_0001_rgb.jpg}
    \includegraphics[width=0.19\textwidth]{images/Vanilla/fern/ngp_ep3000_0001_depth.jpg}
    \includegraphics[width=0.19\textwidth]{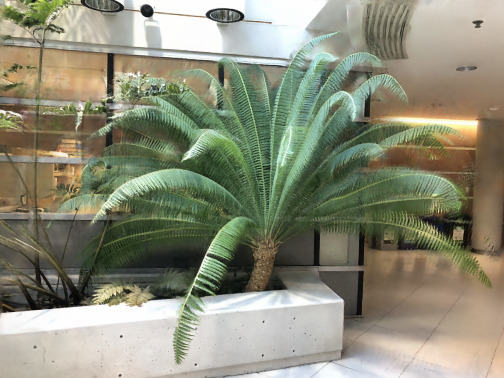}
    \includegraphics[width=0.19\textwidth]{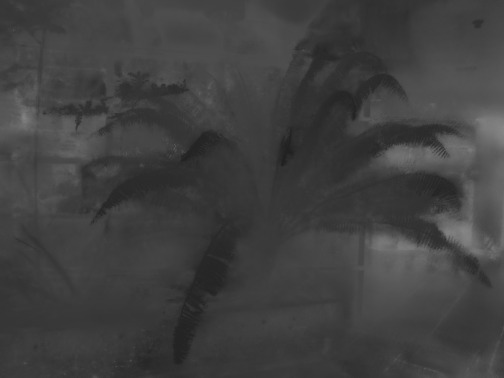}
    \\
    \includegraphics[width=0.19\textwidth]{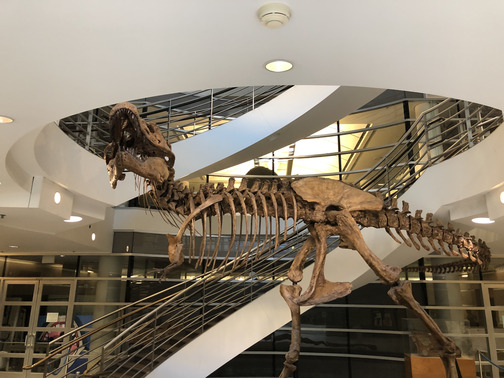}
    \includegraphics[width=0.19\textwidth]{images/Vanilla/trex/ngp_ep3000_0001_rgb.jpg}
    \includegraphics[width=0.19\textwidth]{images/Vanilla/trex/ngp_ep3000_0001_depth.jpg}
    \includegraphics[width=0.19\textwidth]{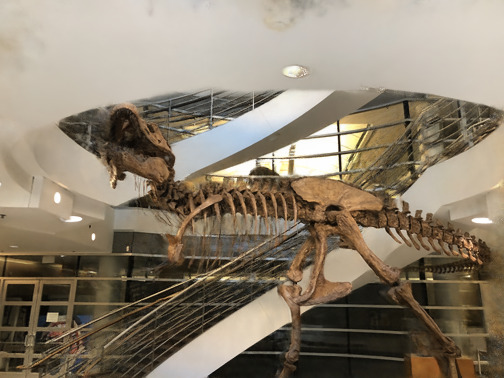}
    \includegraphics[width=0.19\textwidth]{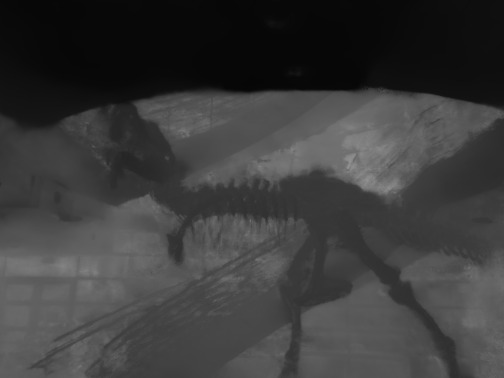}
    \\
    \includegraphics[width=0.19\textwidth]{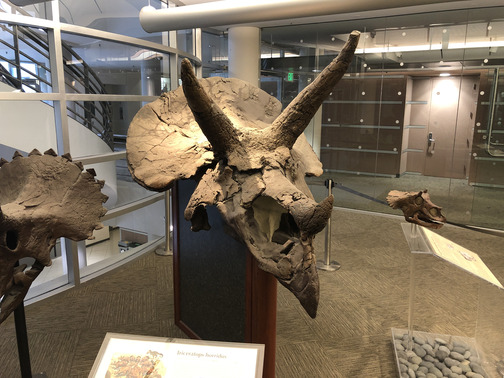}
    \includegraphics[width=0.19\textwidth]{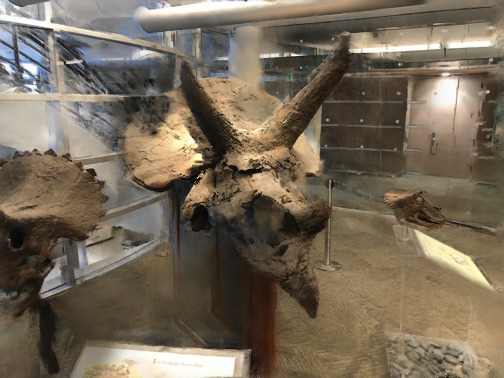}
    \includegraphics[width=0.19\textwidth]{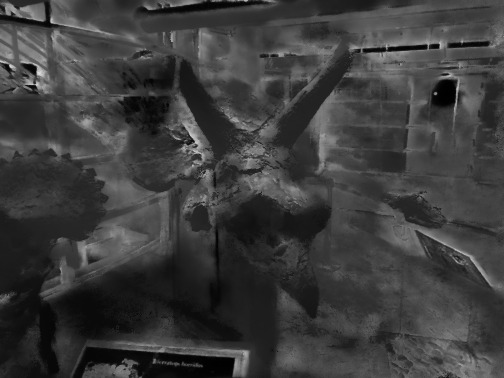}
    \includegraphics[width=0.19\textwidth]{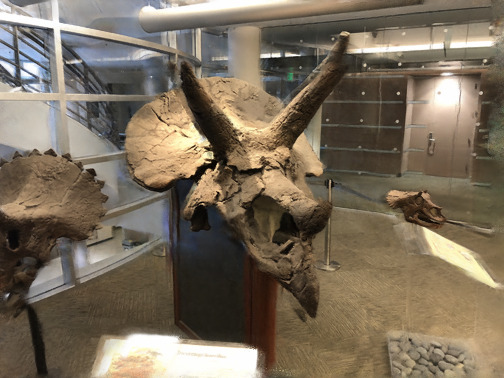}
    \includegraphics[width=0.19\textwidth]{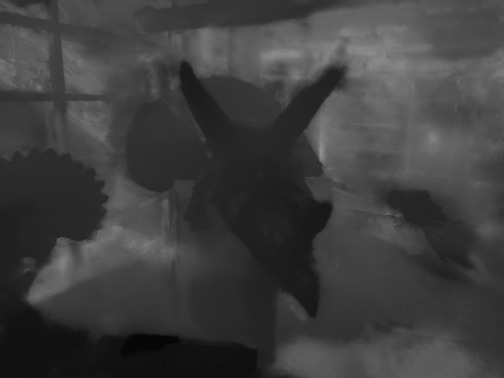}
    \\
    \includegraphics[width=0.19\textwidth]{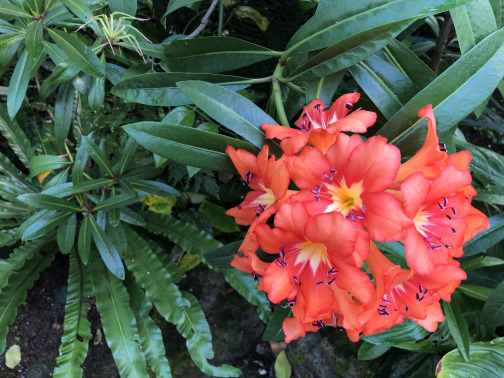}
    \includegraphics[width=0.19\textwidth]{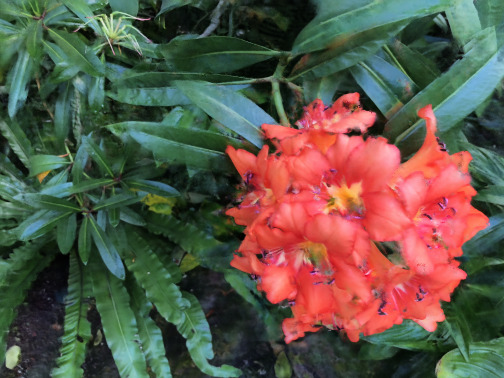}
    \includegraphics[width=0.19\textwidth]{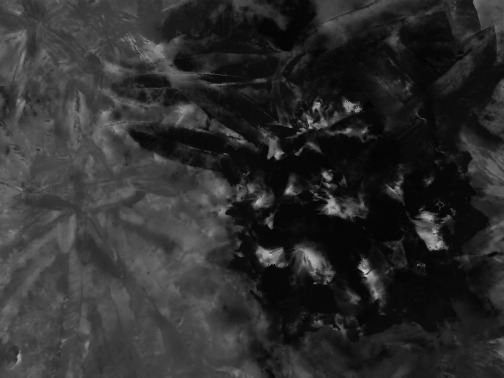}
    \includegraphics[width=0.19\textwidth]{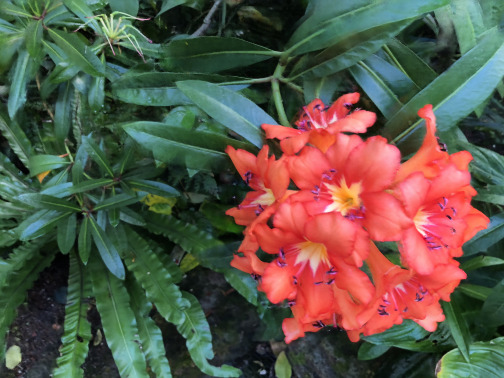}
    \includegraphics[width=0.19\textwidth]{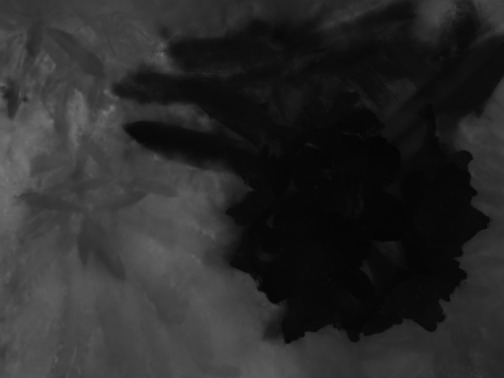}
    \\
    \includegraphics[width=0.19\textwidth]{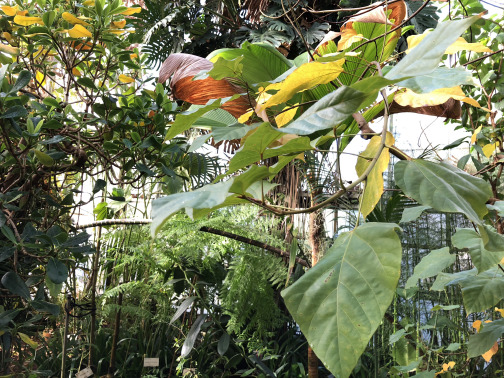}
    \includegraphics[width=0.19\textwidth]{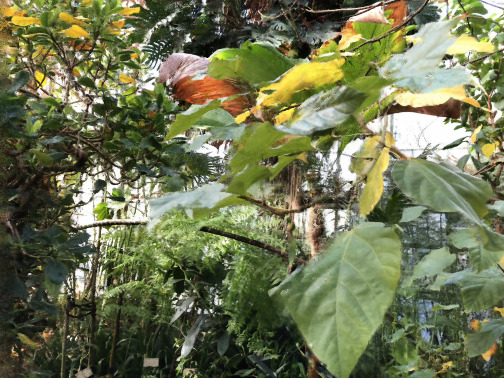}
    \includegraphics[width=0.19\textwidth]{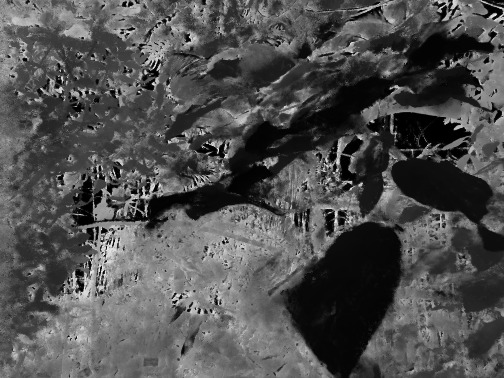}
    \includegraphics[width=0.19\textwidth]{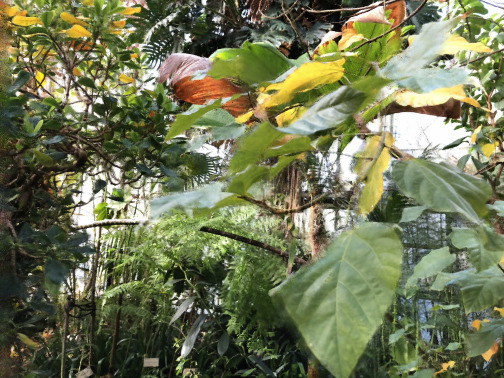}
    \includegraphics[width=0.19\textwidth]{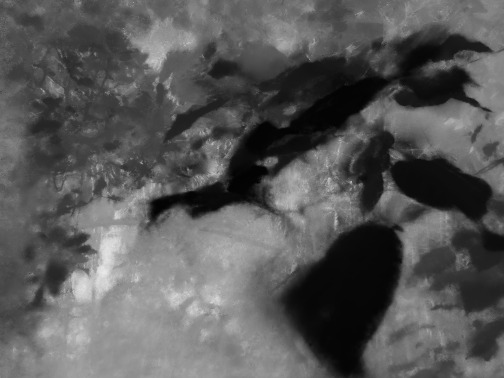}
    \\
    \includegraphics[width=0.19\textwidth]{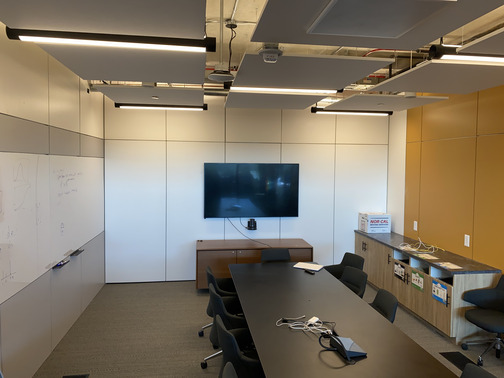}
    \includegraphics[width=0.19\textwidth]{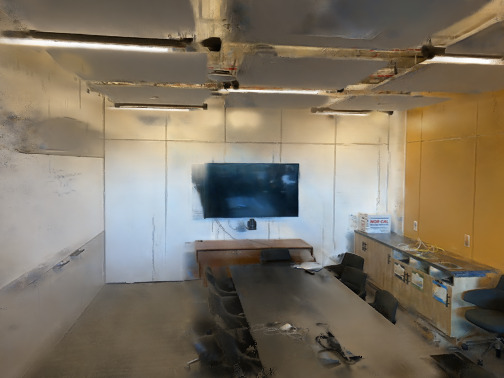}
    \includegraphics[width=0.19\textwidth]{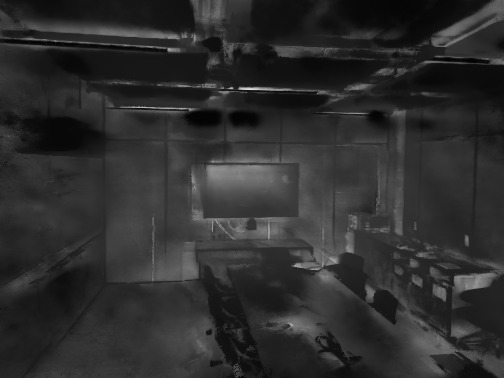}
    \includegraphics[width=0.19\textwidth]{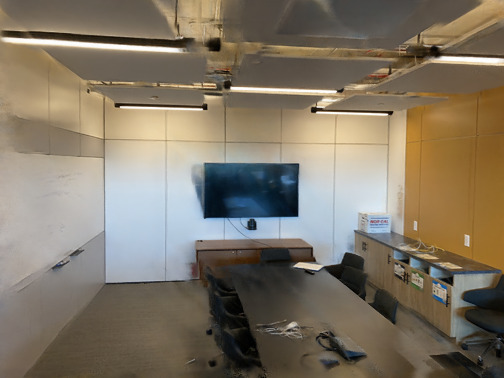}
    \includegraphics[width=0.19\textwidth]{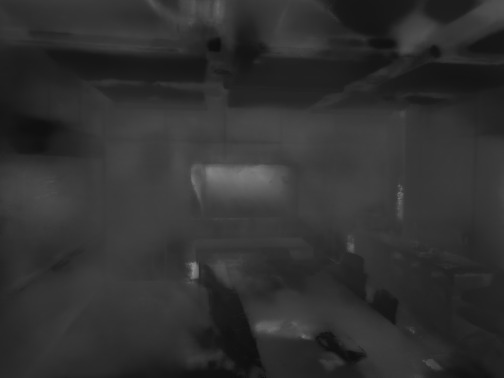}
    \\
    \includegraphics[width=0.19\textwidth]{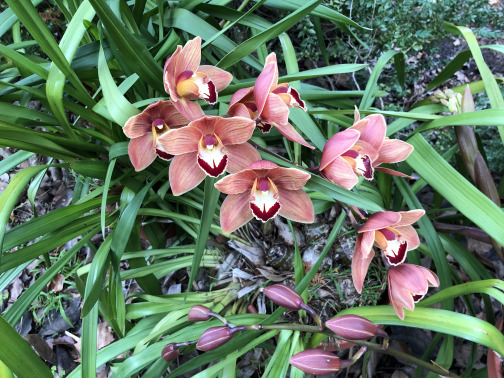}
    \includegraphics[width=0.19\textwidth]{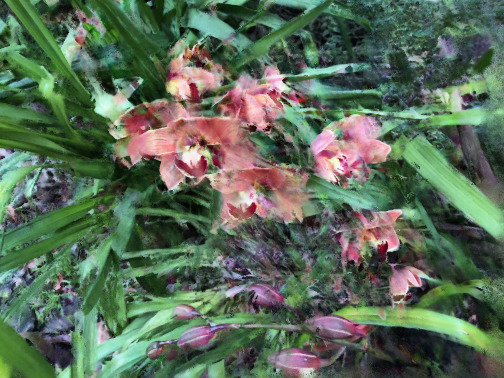}
    \includegraphics[width=0.19\textwidth]{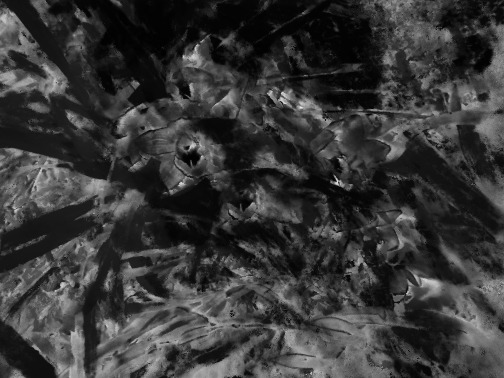}
    \includegraphics[width=0.19\textwidth]{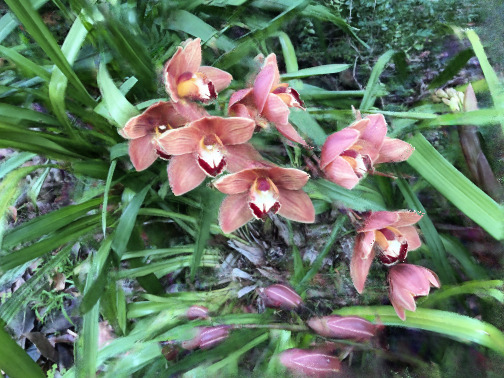}
    \includegraphics[width=0.19\textwidth]{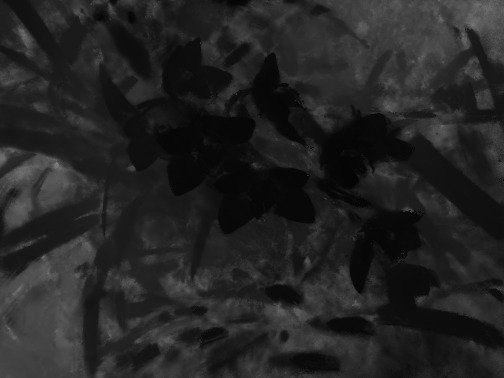}
    \\
    \includegraphics[width=0.19\textwidth]{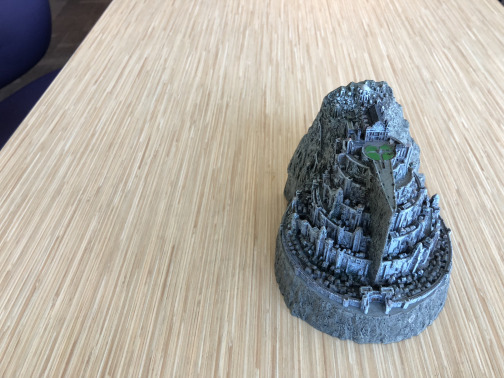}
    \includegraphics[width=0.19\textwidth]{images/Vanilla/fortress/ngp_ep3000_0001_rgb.jpg}
    \includegraphics[width=0.19\textwidth]{images/Vanilla/fortress/ngp_ep3000_0001_depth.jpg}
    \includegraphics[width=0.19\textwidth]{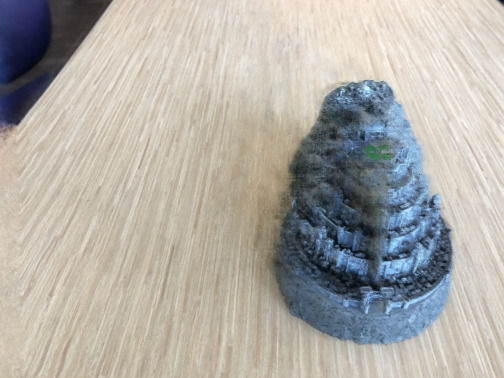}
    \includegraphics[width=0.19\textwidth]{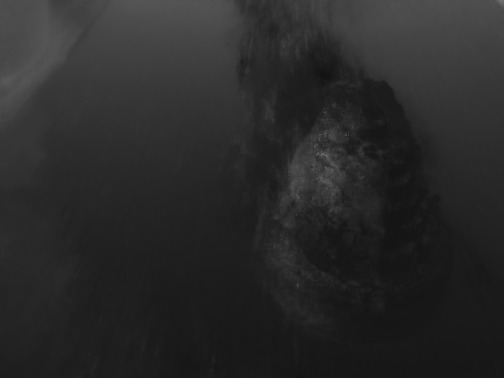}
    \caption{Additional qualitative results on LLFF dataset with 3-view setting.} 
    \label{fig:figure7}
\end{figure*}
\begin{figure*}[t] \centering
    \makebox[0.19\textwidth]{Ground Truth}
    \makebox[0.19\textwidth]{Vanilla NeRF}
    \makebox[0.19\textwidth]{}
    \makebox[0.19\textwidth]{CombiNeRF}
    \makebox[0.19\textwidth]{}
    \\
    \includegraphics[width=0.19\textwidth]{images/GT/fern/image000.jpg}
    \includegraphics[width=0.19\textwidth]{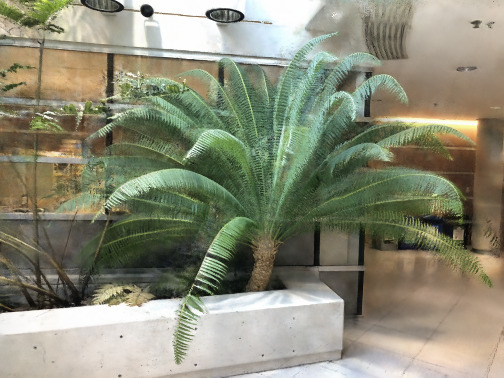}
    \includegraphics[width=0.19\textwidth]{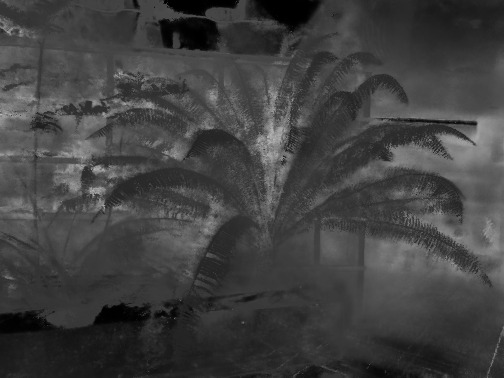}
    \includegraphics[width=0.19\textwidth]{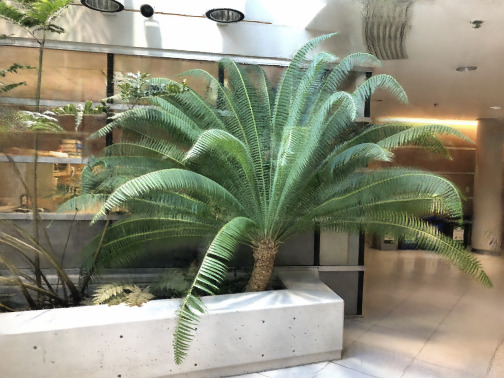}
    \includegraphics[width=0.19\textwidth]{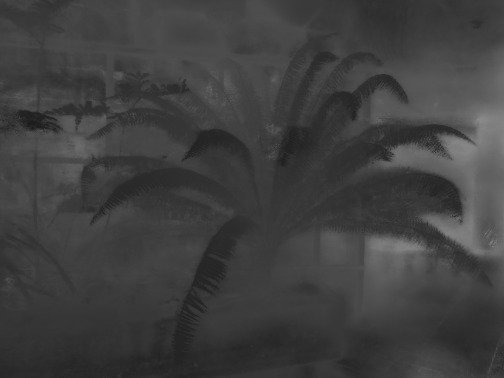}
    \\
    \includegraphics[width=0.19\textwidth]{images/GT/trex/DJI_20200223_163548_810.jpg}
    \includegraphics[width=0.19\textwidth]{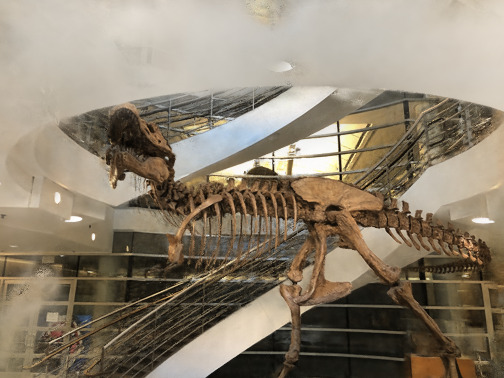}
    \includegraphics[width=0.19\textwidth]{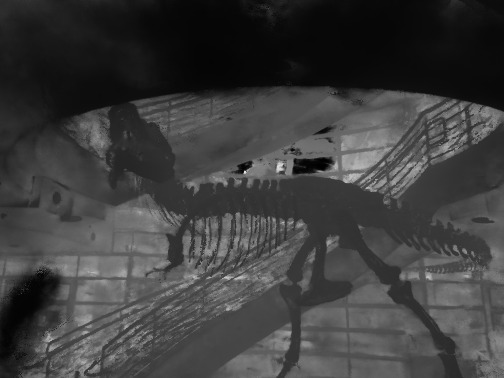}
    \includegraphics[width=0.19\textwidth]{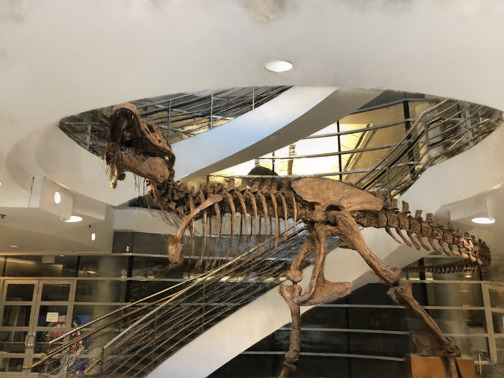}
    \includegraphics[width=0.19\textwidth]{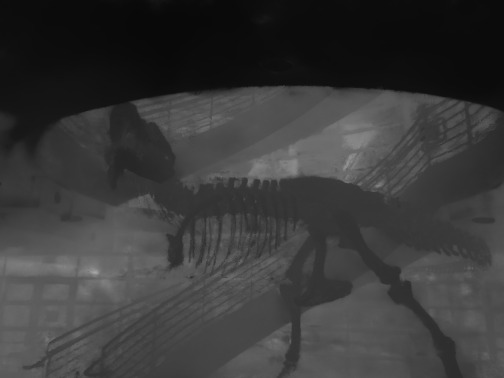}
    \\
    \includegraphics[width=0.19\textwidth]{images/GT/horns/DJI_20200223_163016_842.jpg}
    \includegraphics[width=0.19\textwidth]{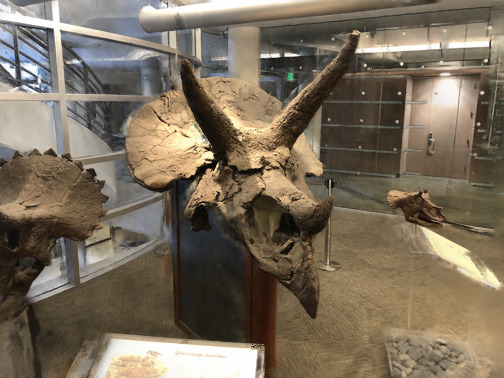}
    \includegraphics[width=0.19\textwidth]{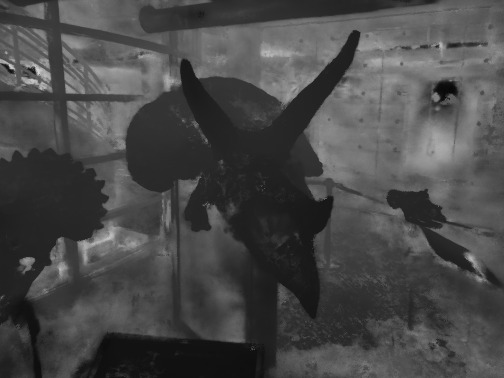}
    \includegraphics[width=0.19\textwidth]{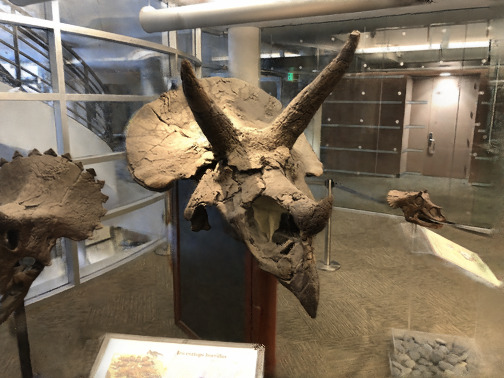}
    \includegraphics[width=0.19\textwidth]{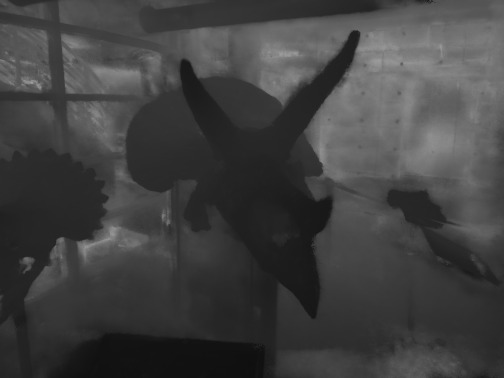}
    \\
    \includegraphics[width=0.19\textwidth]{images/GT/flower/image000.jpg}
    \includegraphics[width=0.19\textwidth]{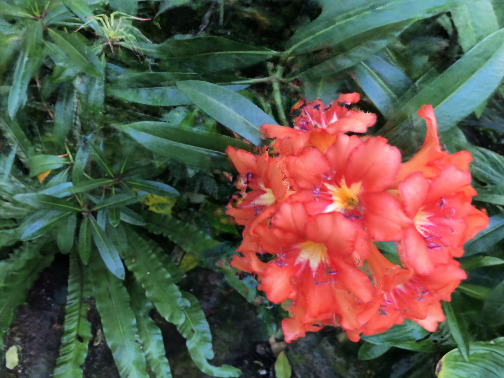}
    \includegraphics[width=0.19\textwidth]{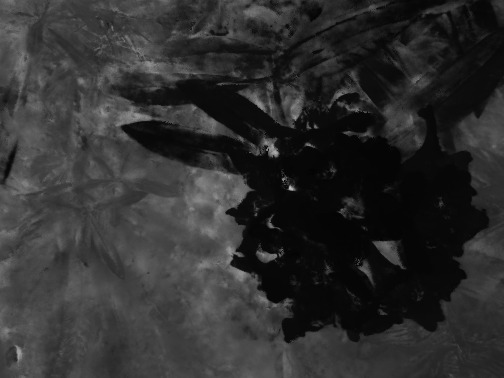}
    \includegraphics[width=0.19\textwidth]{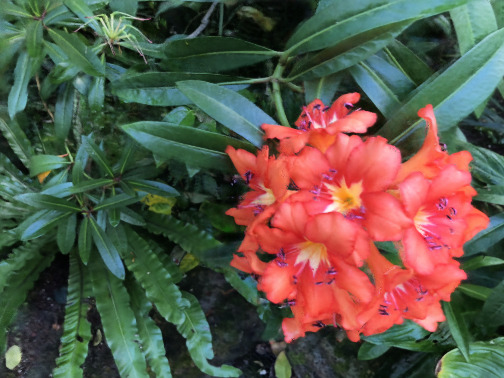}
    \includegraphics[width=0.19\textwidth]{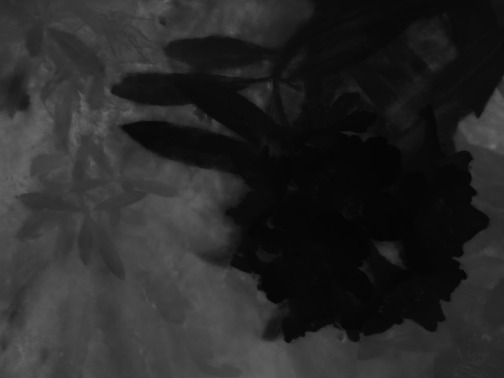}
    \\
    \includegraphics[width=0.19\textwidth]{images/GT/leaves/image000.jpg}
    \includegraphics[width=0.19\textwidth]{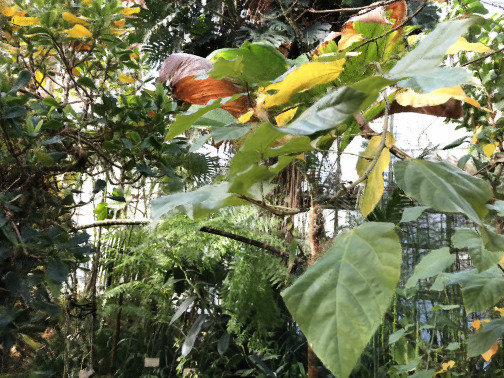}
    \includegraphics[width=0.19\textwidth]{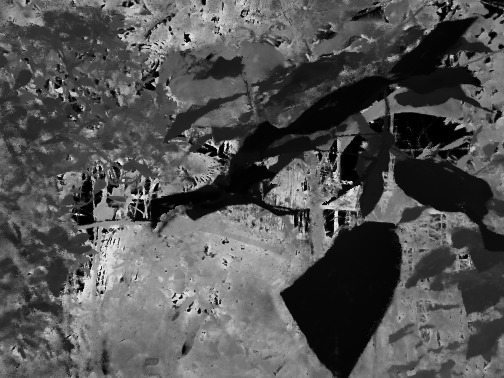}
    \includegraphics[width=0.19\textwidth]{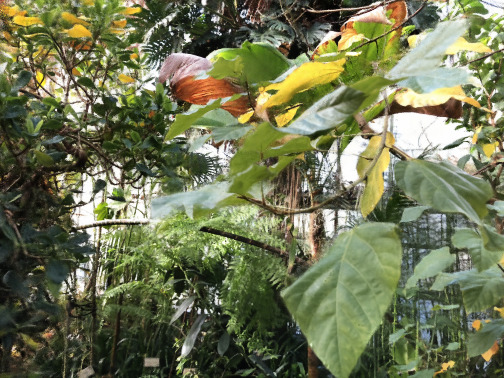}
    \includegraphics[width=0.19\textwidth]{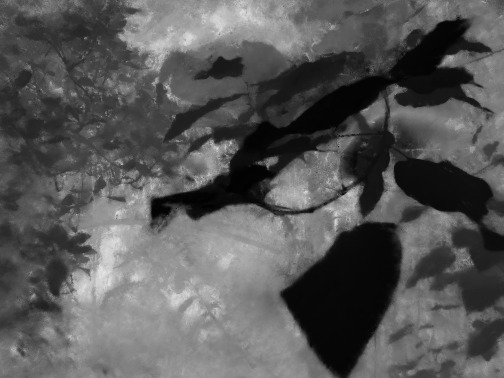}
    \\
    \includegraphics[width=0.19\textwidth]{images/GT/room/DJI_20200226_143850_006.jpg}
    \includegraphics[width=0.19\textwidth]{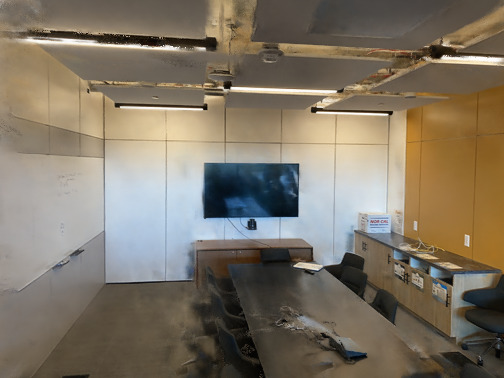}
    \includegraphics[width=0.19\textwidth]{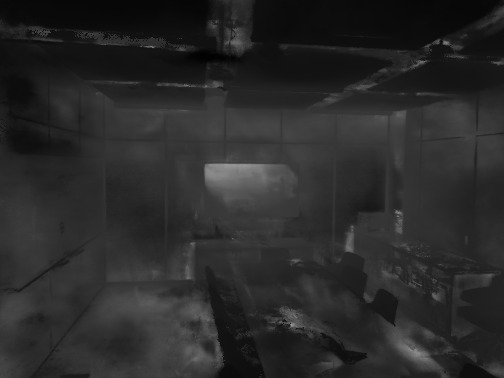}
    \includegraphics[width=0.19\textwidth]{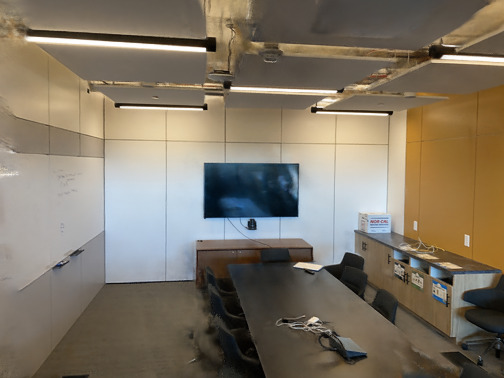}
    \includegraphics[width=0.19\textwidth]{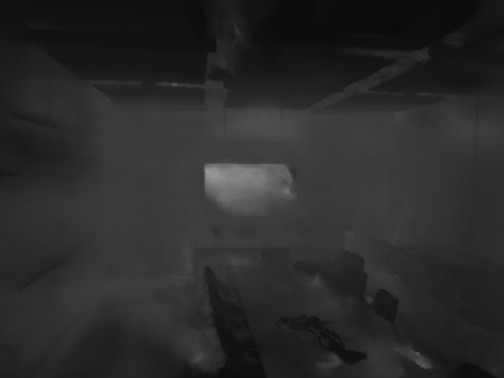}
    \\
    \includegraphics[width=0.19\textwidth]{images/GT/orchids/image000.jpg}
    \includegraphics[width=0.19\textwidth]{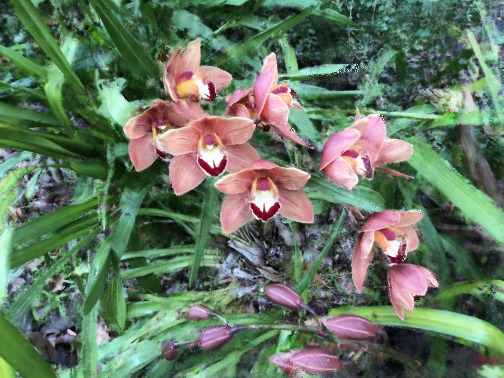}
    \includegraphics[width=0.19\textwidth]{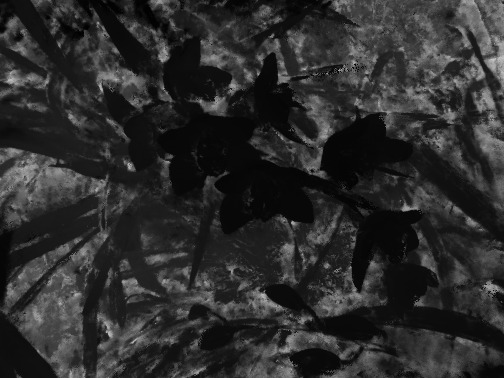}
    \includegraphics[width=0.19\textwidth]{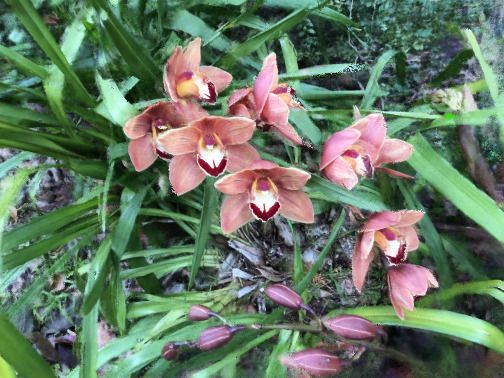}
    \includegraphics[width=0.19\textwidth]{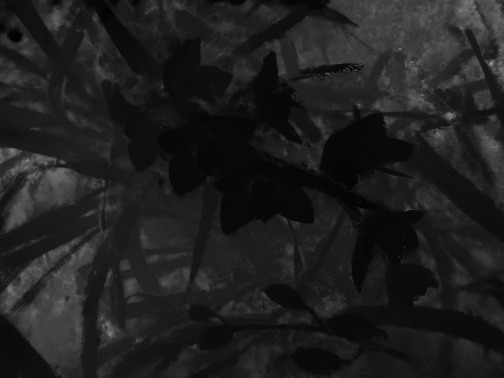}
    \\
    \includegraphics[width=0.19\textwidth]{images/GT/fortress/image000.jpg}
    \includegraphics[width=0.19\textwidth]{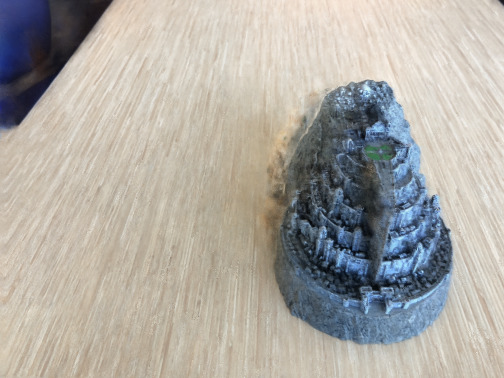}
    \includegraphics[width=0.19\textwidth]{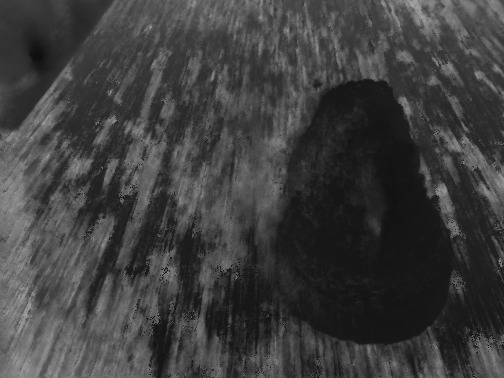}
    \includegraphics[width=0.19\textwidth]{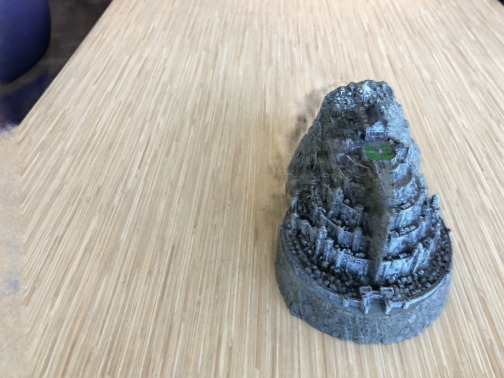}
    \includegraphics[width=0.19\textwidth]{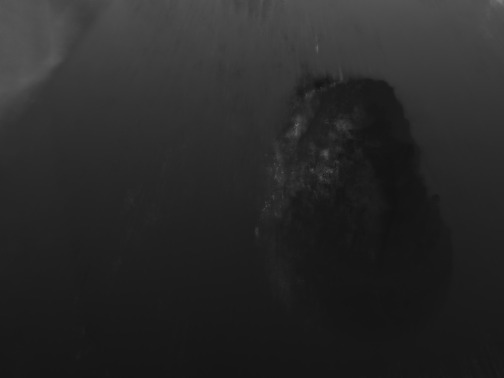}
    \caption{Additional qualitative results on LLFF dataset with 6-view setting.} 
    \label{fig:figure8}
\end{figure*}
\begin{figure*}[t] \centering
    \makebox[0.19\textwidth]{Ground Truth}
    \makebox[0.19\textwidth]{Vanilla NeRF}
    \makebox[0.19\textwidth]{}
    \makebox[0.19\textwidth]{CombiNeRF}
    \makebox[0.19\textwidth]{}
    \\
    \includegraphics[width=0.19\textwidth]{images/GT/fern/image000.jpg}
    \includegraphics[width=0.19\textwidth]{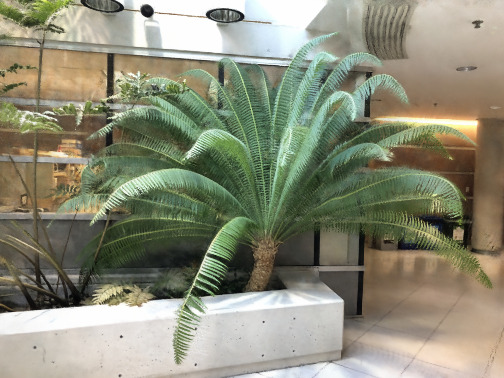}
    \includegraphics[width=0.19\textwidth]{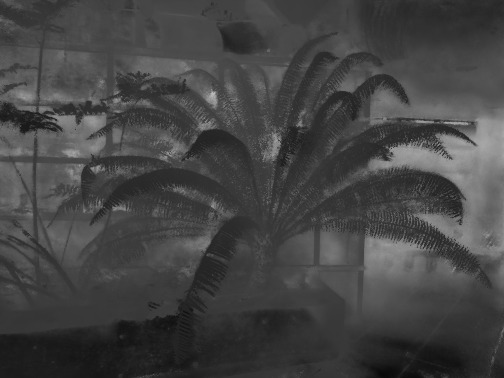}
    \includegraphics[width=0.19\textwidth]{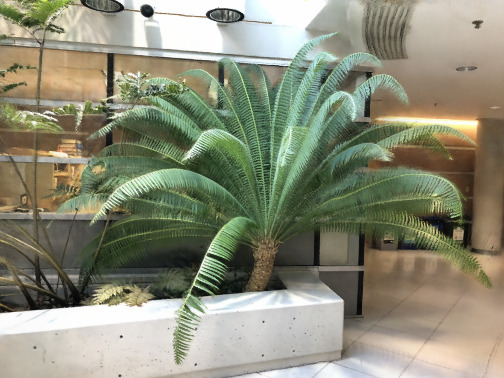}
    \includegraphics[width=0.19\textwidth]{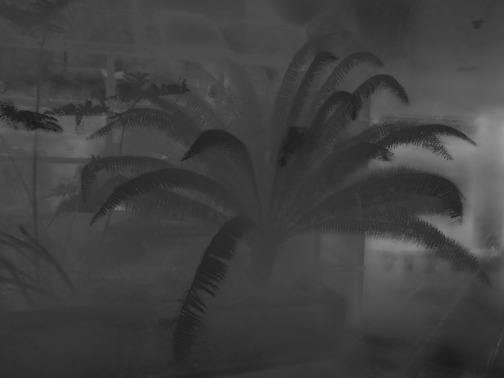}
    \\
    \includegraphics[width=0.19\textwidth]{images/GT/trex/DJI_20200223_163548_810.jpg}
    \includegraphics[width=0.19\textwidth]{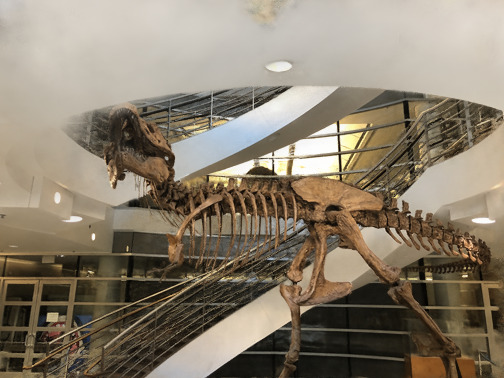}
    \includegraphics[width=0.19\textwidth]{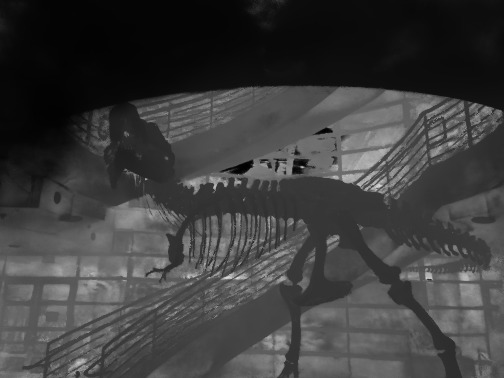}
    \includegraphics[width=0.19\textwidth]{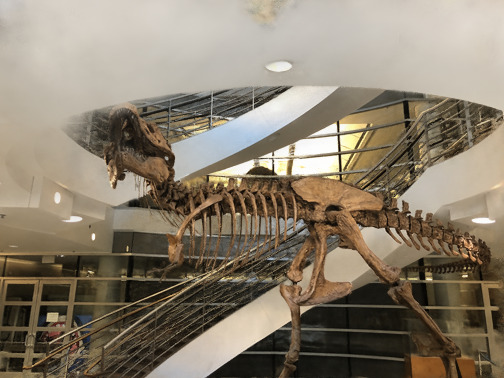}
    \includegraphics[width=0.19\textwidth]{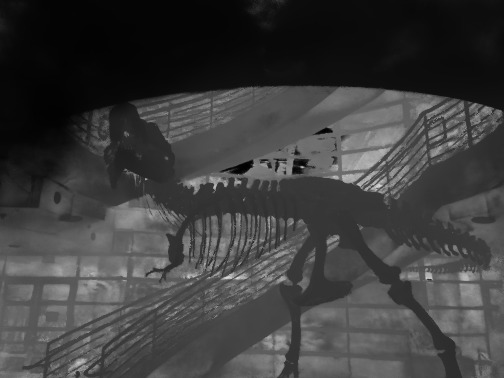}
    \\
    \includegraphics[width=0.19\textwidth]{images/GT/horns/DJI_20200223_163016_842.jpg}
    \includegraphics[width=0.19\textwidth]{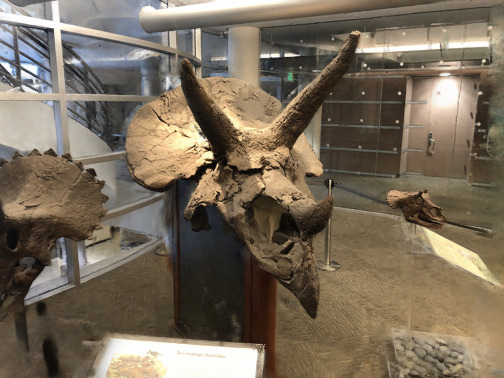}
    \includegraphics[width=0.19\textwidth]{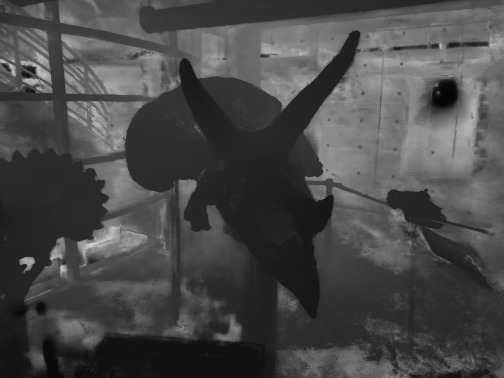}
    \includegraphics[width=0.19\textwidth]{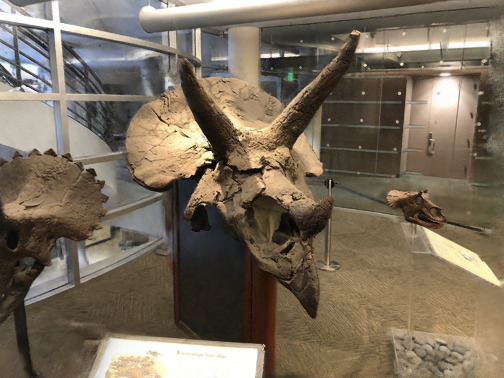}
    \includegraphics[width=0.19\textwidth]{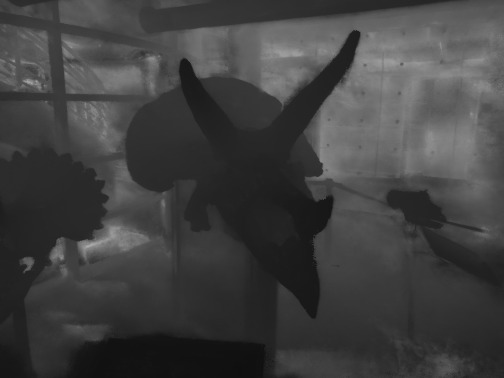}
    \\
    \includegraphics[width=0.19\textwidth]{images/GT/flower/image000.jpg}
    \includegraphics[width=0.19\textwidth]{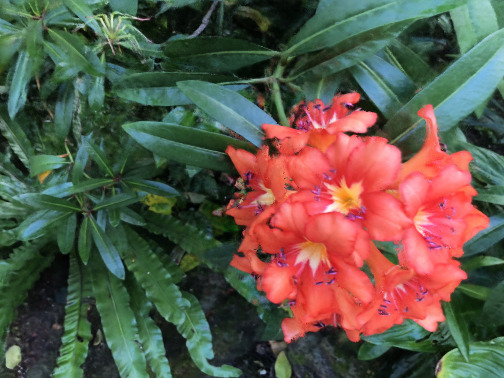}
    \includegraphics[width=0.19\textwidth]{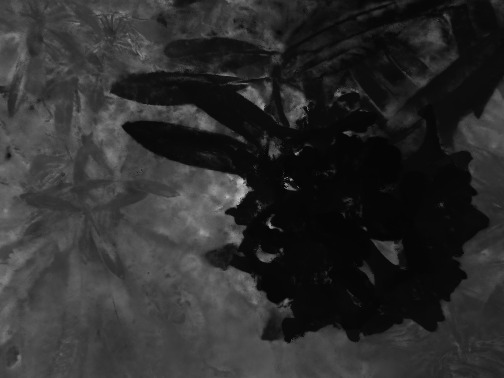}
    \includegraphics[width=0.19\textwidth]{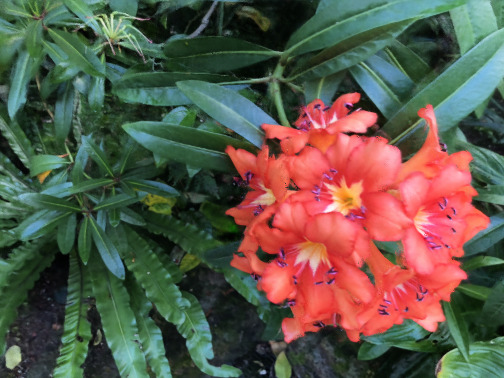}
    \includegraphics[width=0.19\textwidth]{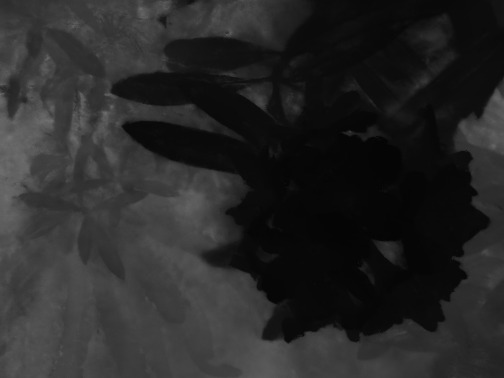}
    \\
    \includegraphics[width=0.19\textwidth]{images/GT/leaves/image000.jpg}
    \includegraphics[width=0.19\textwidth]{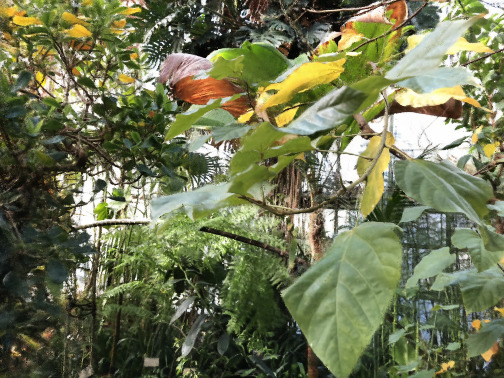}
    \includegraphics[width=0.19\textwidth]{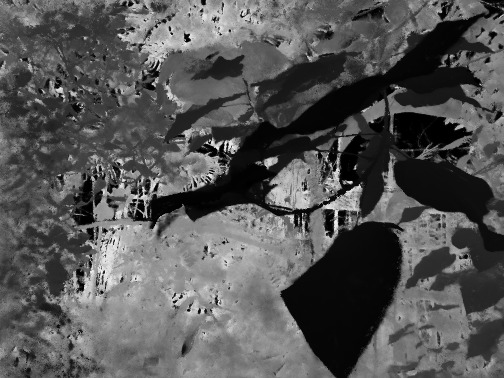}
    \includegraphics[width=0.19\textwidth]{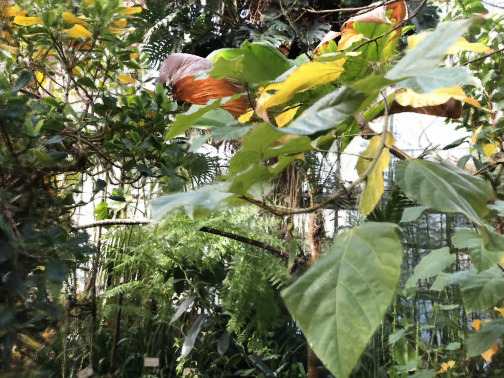}
    \includegraphics[width=0.19\textwidth]{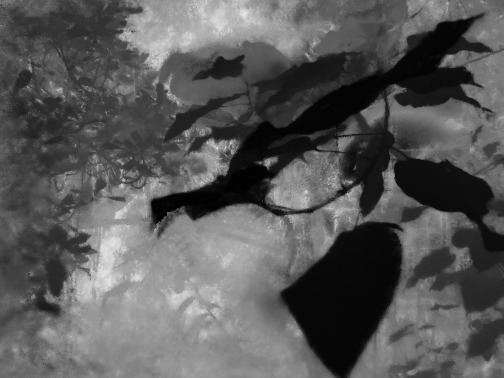}
    \\
    \includegraphics[width=0.19\textwidth]{images/GT/room/DJI_20200226_143850_006.jpg}
    \includegraphics[width=0.19\textwidth]{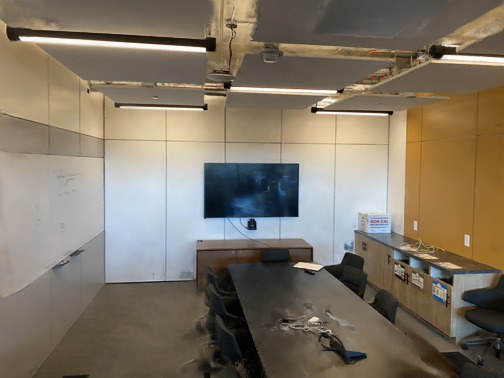}
    \includegraphics[width=0.19\textwidth]{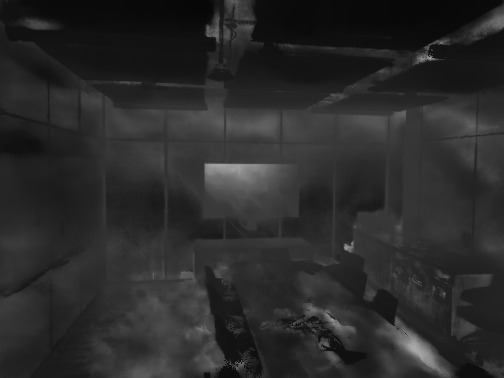}
    \includegraphics[width=0.19\textwidth]{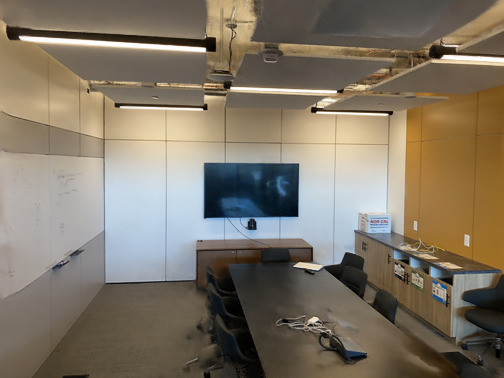}
    \includegraphics[width=0.19\textwidth]{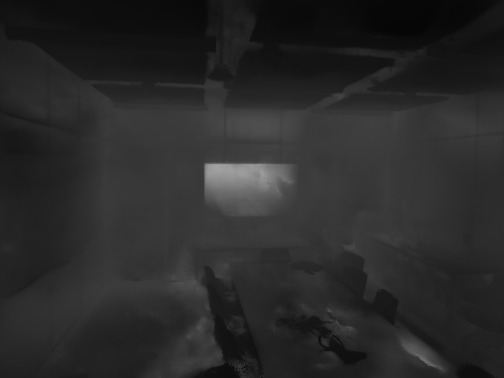}
    \\
    \includegraphics[width=0.19\textwidth]{images/GT/orchids/image000.jpg}
    \includegraphics[width=0.19\textwidth]{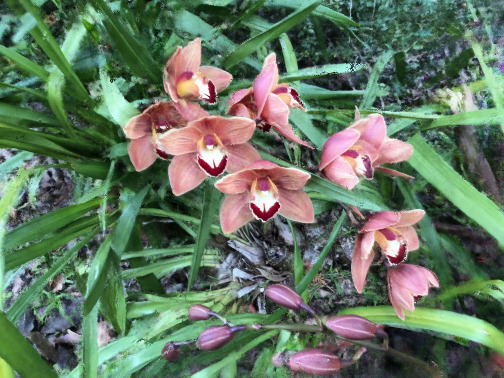}
    \includegraphics[width=0.19\textwidth]{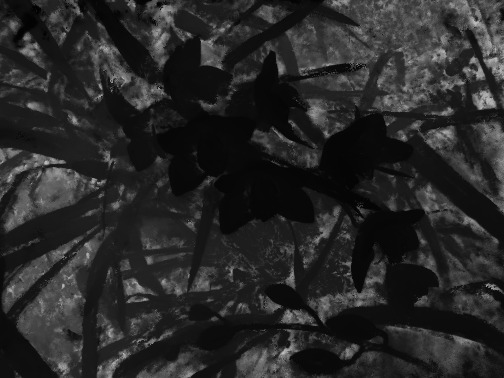}
    \includegraphics[width=0.19\textwidth]{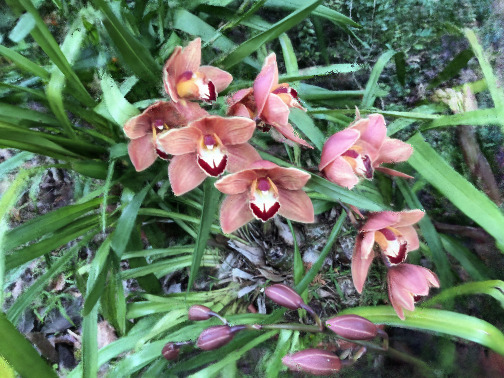}
    \includegraphics[width=0.19\textwidth]{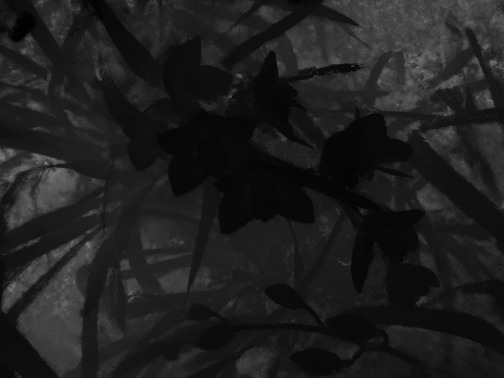}
    \\
    \includegraphics[width=0.19\textwidth]{images/GT/fortress/image000.jpg}
    \includegraphics[width=0.19\textwidth]{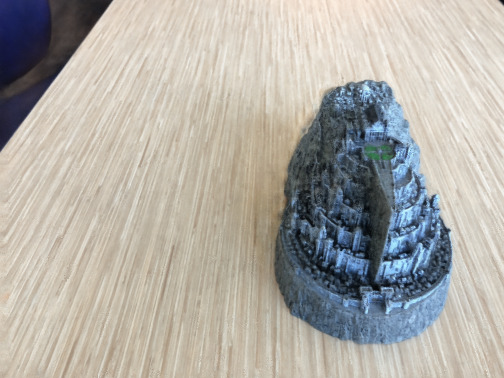}
    \includegraphics[width=0.19\textwidth]{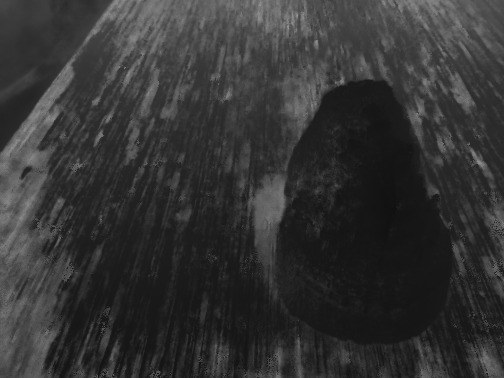}
    \includegraphics[width=0.19\textwidth]{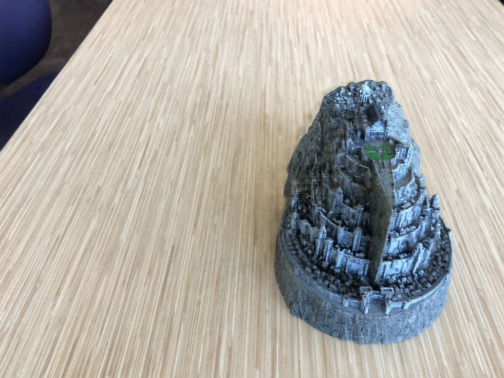}
    \includegraphics[width=0.19\textwidth]{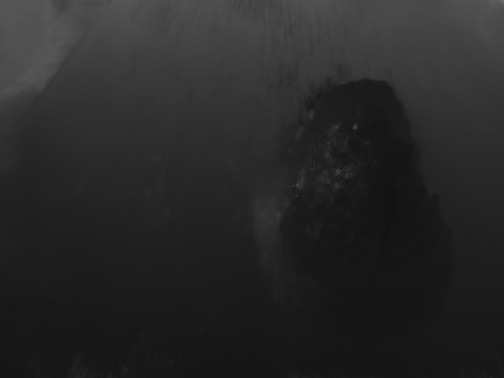}
    \caption{Additional qualitative results on LLFF dataset with 9-view setting.} 
    \label{fig:figure9}
\end{figure*}
\begin{figure*}[t] \centering
    \makebox[0.15\textwidth]{Ground Truth}
    \makebox[0.15\textwidth]{Vanilla NeRF}
    \makebox[0.15\textwidth]{}
    \makebox[0.15\textwidth]{CombiNeRF}
    \makebox[0.15\textwidth]{}
    \\
    \includegraphics[width=0.15\textwidth]{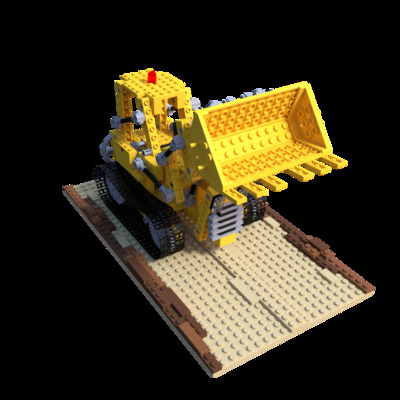}
    \includegraphics[width=0.15\textwidth]{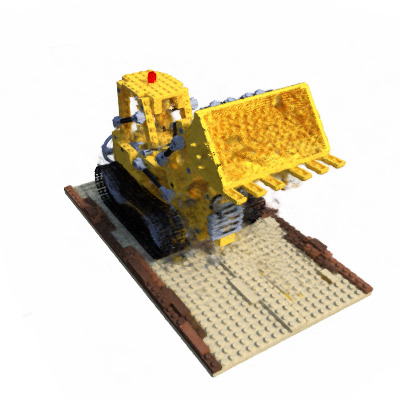}
    \includegraphics[width=0.15\textwidth]{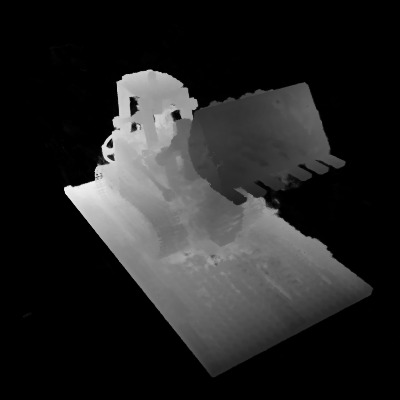}
    \includegraphics[width=0.15\textwidth]{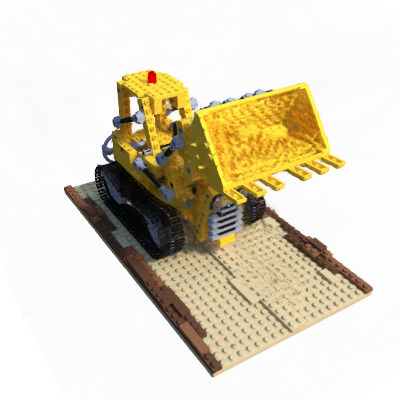}
    \includegraphics[width=0.15\textwidth]{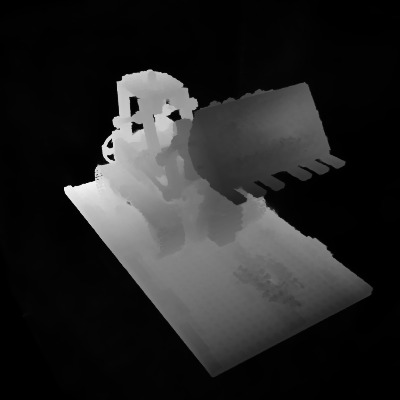}
    \\
    \includegraphics[width=0.15\textwidth]{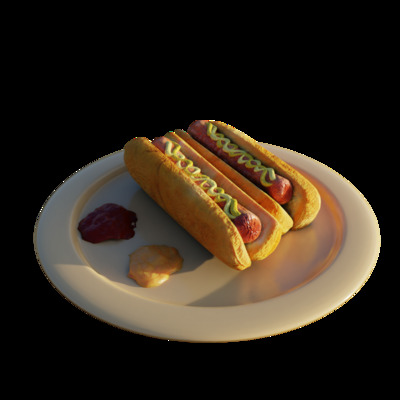}
    \includegraphics[width=0.15\textwidth]{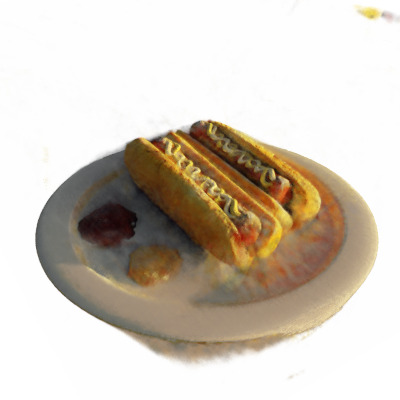}
    \includegraphics[width=0.15\textwidth]{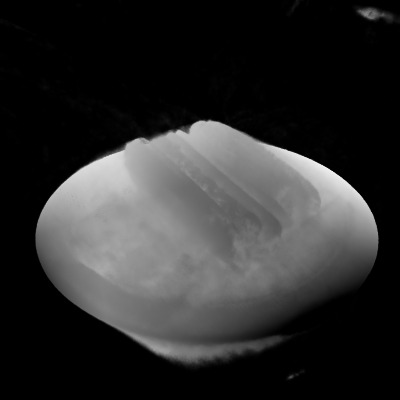}
    \includegraphics[width=0.15\textwidth]{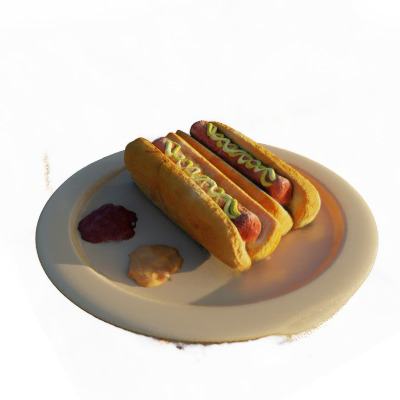}
    \includegraphics[width=0.15\textwidth]{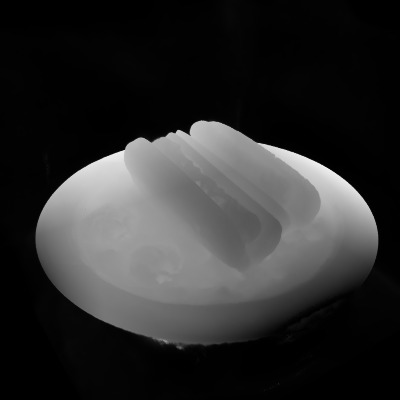}
    \\
    \includegraphics[width=0.15\textwidth]{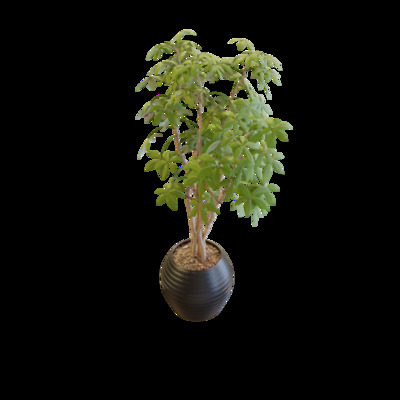}
    \includegraphics[width=0.15\textwidth]{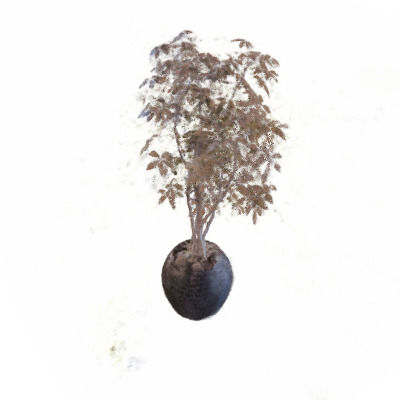}
    \includegraphics[width=0.15\textwidth]{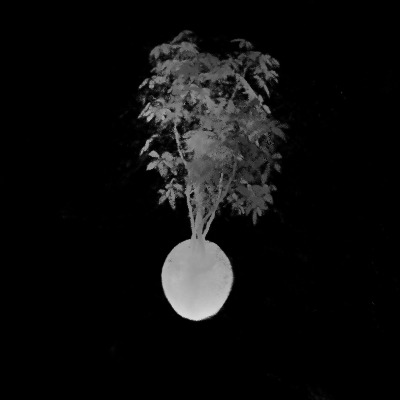}
    \includegraphics[width=0.15\textwidth]{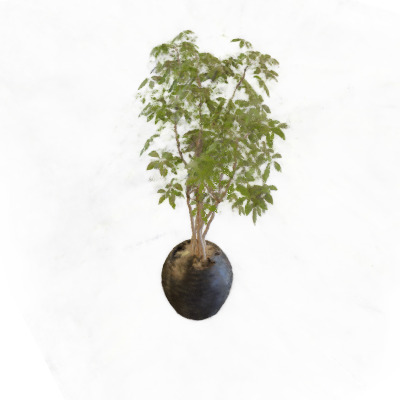}
    \includegraphics[width=0.15\textwidth]{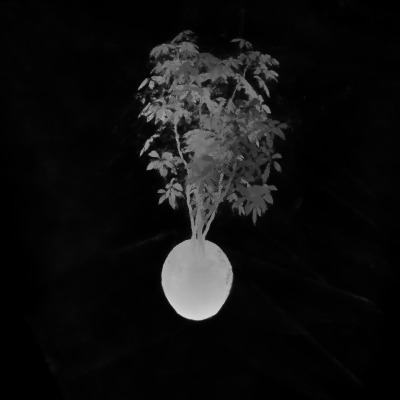}
    \\
    \includegraphics[width=0.15\textwidth]{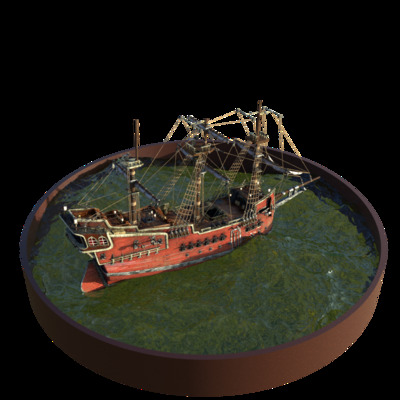}
    \includegraphics[width=0.15\textwidth]{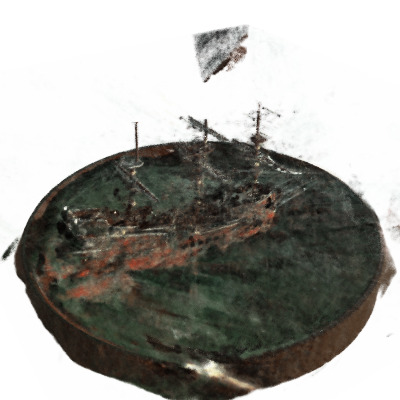}
    \includegraphics[width=0.15\textwidth]{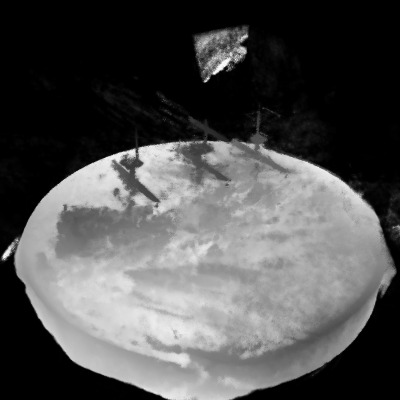}
    \includegraphics[width=0.15\textwidth]{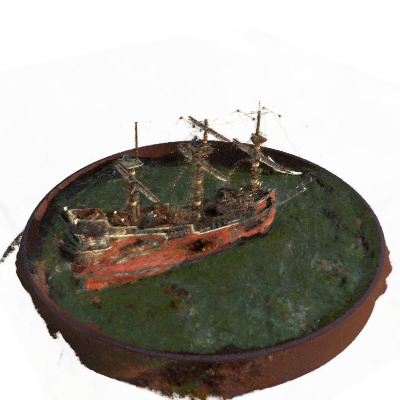}
    \includegraphics[width=0.15\textwidth]{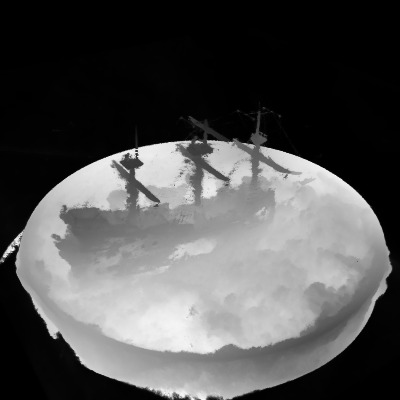}
    \\
    \includegraphics[width=0.15\textwidth]{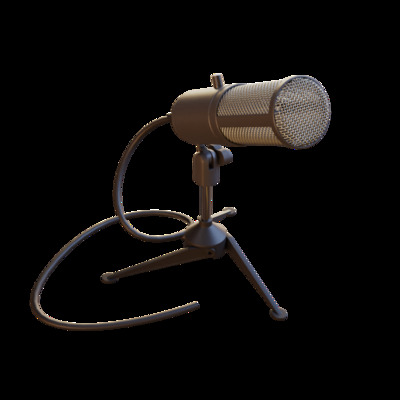}
    \includegraphics[width=0.15\textwidth]{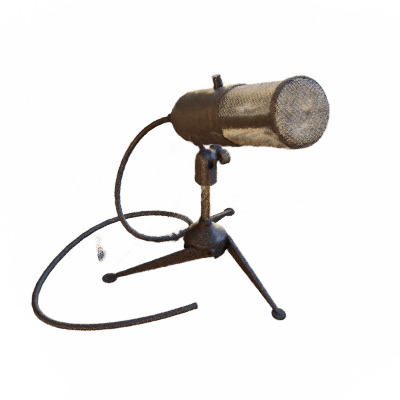}
    \includegraphics[width=0.15\textwidth]{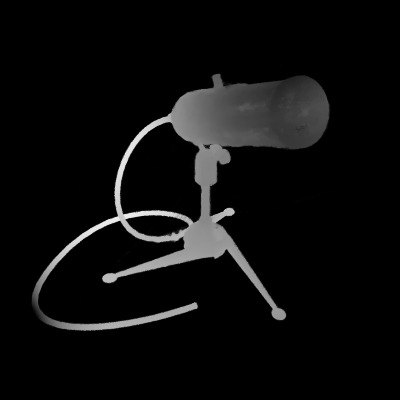}
    \includegraphics[width=0.15\textwidth]{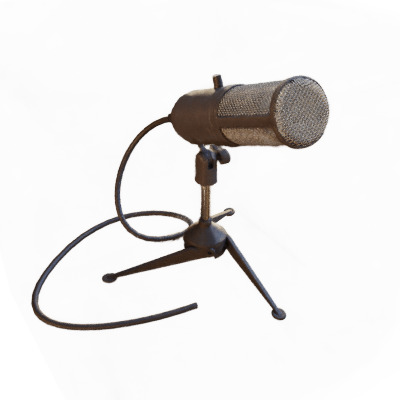}
    \includegraphics[width=0.15\textwidth]{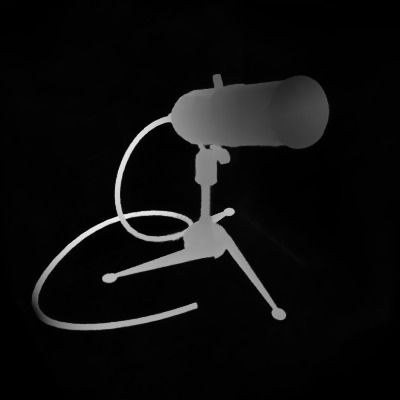}
    \\
    \includegraphics[width=0.15\textwidth]{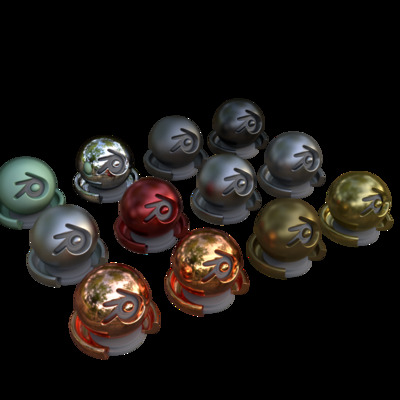}
    \includegraphics[width=0.15\textwidth]{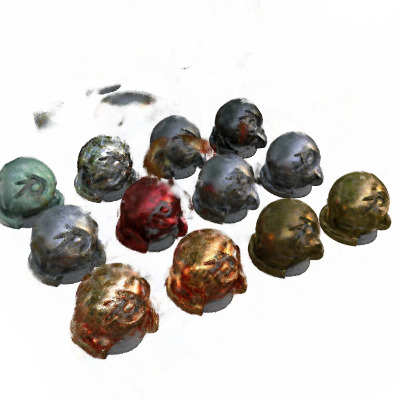}
    \includegraphics[width=0.15\textwidth]{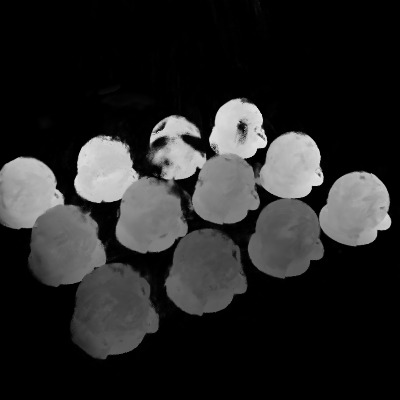}
    \includegraphics[width=0.15\textwidth]{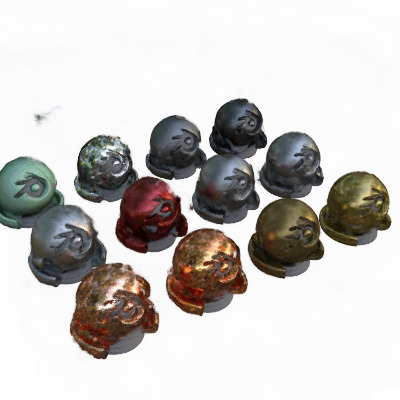}
    \includegraphics[width=0.15\textwidth]{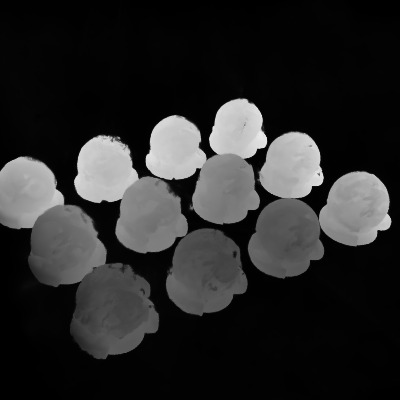}
    \\
    \includegraphics[width=0.15\textwidth]{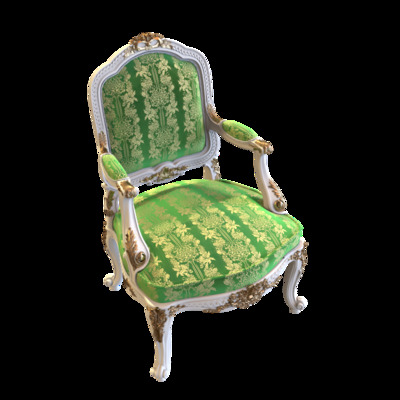}
    \includegraphics[width=0.15\textwidth]{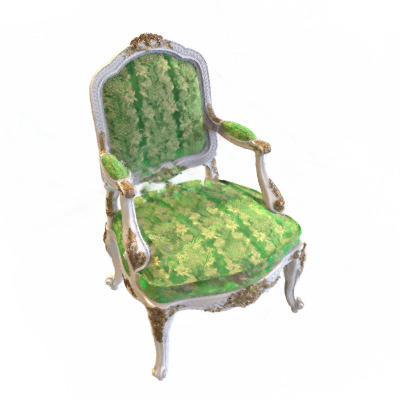}
    \includegraphics[width=0.15\textwidth]{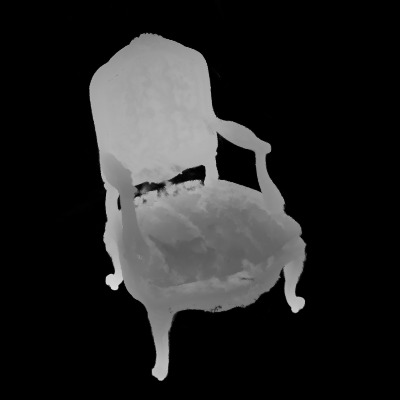}
    \includegraphics[width=0.15\textwidth]{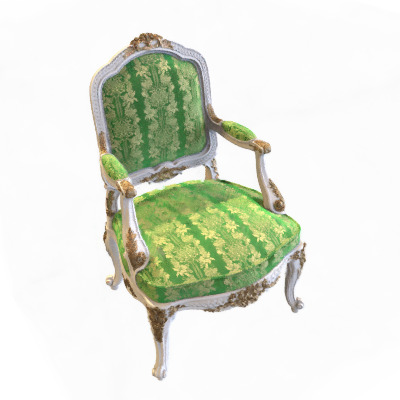}
    \includegraphics[width=0.15\textwidth]{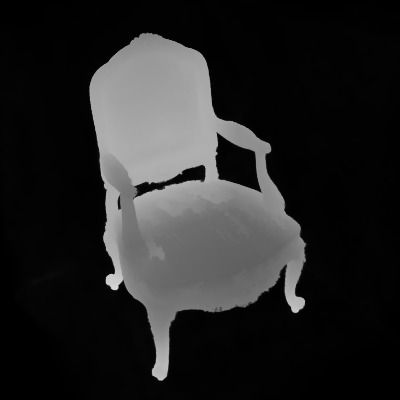}
    \\
    \includegraphics[width=0.15\textwidth]{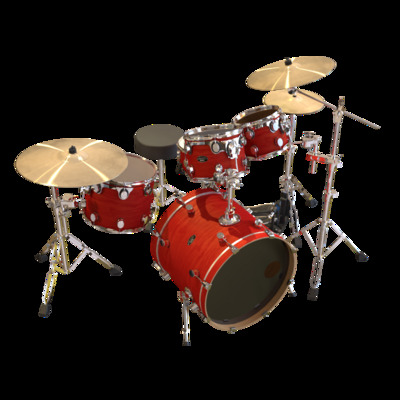}
    \includegraphics[width=0.15\textwidth]{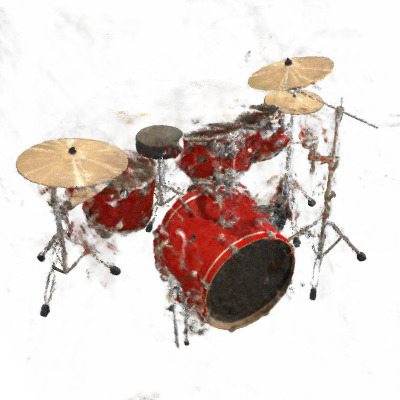}
    \includegraphics[width=0.15\textwidth]{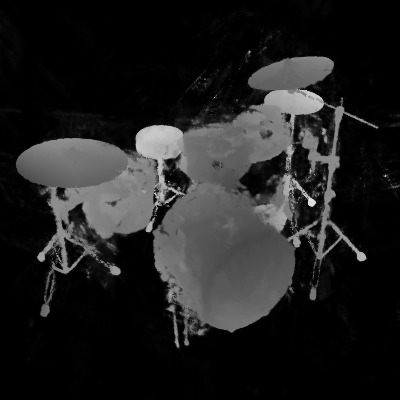}
    \includegraphics[width=0.15\textwidth]{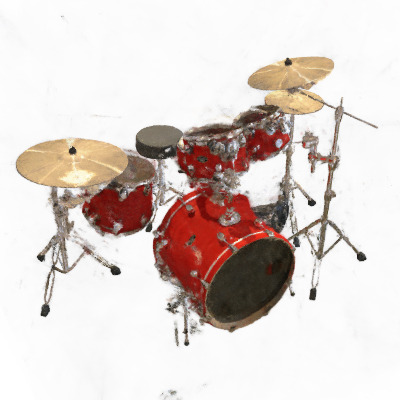}
    \includegraphics[width=0.15\textwidth]{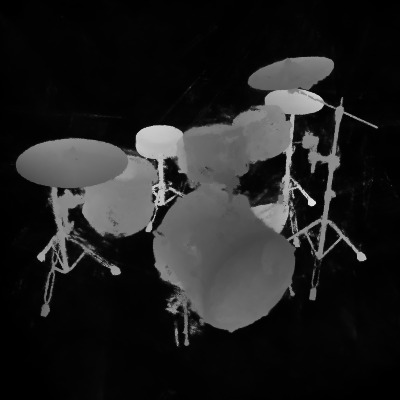}
    \caption{Additional qualitative results on NeRF-Synthetic dataset with 8-view setting.} 
    \label{fig:figure10}
\end{figure*}
In \cref{fig:figure7}, \cref{fig:figure8} and \cref{fig:figure9} we show additional qualitative results on all scenarios of the LLFF dataset under 3-view, 6-view, 9-view settings, respectively. Under the 3-view setting, we can see the effectiveness of CombiNeRF against Vanilla NeRF in both RGB images and depth images. When the number of input images increases, the performance gap between Vanilla NeRF and CombiNeRF decreases, but the depth quality still remains definitely better in the latter.

\cref{fig:figure10} shows additional qualitative results on all scenarios of the NeRF-Synthetic dataset under the 8-view setting. CombiNeRF is able to better reconstruct the scenarios, and this is particularly visible in the geometry of "Ship" and in the color of "Ficus". Finer details are less visible in Vanilla NeRF, as we can see in "Mic" and "Chair", while a lot of floaters and noise are removed in "Drums" and "Materials" with CombiNeRF. 
\subsection{Quantitative Results}
In \cref{tab:table9}, \cref{tab:table10} and \cref{tab:table11} we show per-scene quantitative results on PNSR, SSIM, and LPIPS metrics, respectively.
We can observe that in "Lego", "Chair", "Mic", "Hotdog" and "Ficus" CombiNeRF gets the better results in all the metrics. Instead in "Drums", "Materials" and "Ship", which are the most challenging scenarios, CombiNeRF outperforms the other methods in the overall metric scores.

In \cref{tab:table12}, \cref{tab:table13} and \cref{tab:table14} we show quantitative results on PSNR, SSIM, and LPIPS metrics respectively, including 3/6/9-view settings. CombiNeRF clearly outperforms Vanilla NeRF in all the scenarios.

\begin{table*}[t]
\resizebox{\textwidth}{!}{
\rowcolors{2}{LightCyan}{White}
\begin{tabular}{c | c c c c c c c c}
\toprule
NeRF-Synthetic 8-views & Drums & Material & Ship & Ficus & Lego & Chair & Mic & Hotdog  \\
\midrule
Vanilla NeRF & 18.07	& 19.36	& 18.69	&22.93	&22.49	&25.25	&27.34	&24.55  \\
DietNeRF, $\mathcal{L}_{MSE}$ ft &  20.029& \textbf{21.621}& \textbf{22.536} & 20.940 & 24.311& 25.595 & 26.794 & 26.626 \\
\midrule
CombiNeRF & \textbf{20.165} &20.73	&21.392	&\textbf{23.493} &\textbf{24.958}	&\textbf{27.862}	&\textbf{28.172}	&\textbf{28.383}   \\
\bottomrule
\end{tabular}
}
\caption{Per-scene quantitative results on NeRF-Synthetic dataset w.r.t. \textbf{PSNR$\uparrow$} metric.} 
\label{tab:table9}
\end{table*}

\begin{table*}[t]
\resizebox{\textwidth}{!}{
\rowcolors{2}{LightCyan}{White}
\begin{tabular}{c | c c c c c c c c}
\toprule
NeRF-Synthetic 8-views & Drums & Material & Ship & Ficus & Lego & Chair & Mic & Hotdog  \\
\midrule
Vanilla NeRF & 0.767	&0.804	&0.69	&0.897	&0.851	&0.901	&0.949	&0.901 \\
DietNeRF, $\mathcal{L}_{MSE}$ ft &  \textbf{0.845} & \textbf{0.851} & \textbf{0.757} & 0.874 & 0.875 & 0.912 & 0.950 & 0.924  \\
\midrule
CombiNeRF & 0.838	&0.838	&0.754	& \textbf{0.91}	& \textbf{0.885}	& \textbf{0.933}	& \textbf{0.957}	& \textbf{0.945} \\
\bottomrule
\end{tabular}
}
\caption{Per-scene quantitative results on NeRF-Synthetic dataset w.r.t. \textbf{SSIM$\uparrow$} metric.} 
\label{tab:table10}
\end{table*}

\begin{table*}[t]
\resizebox{\textwidth}{!}{
\rowcolors{2}{LightCyan}{White}
\begin{tabular}{c | c c c c c c c c}
\toprule
NeRF-Synthetic 8-views & Drums & Material & Ship & Ficus & Lego & Chair & Mic & Hotdog  \\
\midrule
Vanilla NeRF & 0.222	&0.185	&0.269	&0.107	&0.105	&0.076	&0.054	&0.134 \\
DietNeRF, $\mathcal{L}_{MSE}$ ft &  \textbf{0.117} & \textbf{0.095} & 0.193 & 0.094 & 0.096 & 0.077 & 0.043 & 0.067  \\
\midrule
CombiNeRF & 0.14 &	0.114	&\textbf{0.151}	& \textbf{0.078}	& \textbf{0.072}	& \textbf{0.049}	&\textbf{0.035}	&\textbf{0.066} \\
\bottomrule
\end{tabular}
}
\caption{Per-scene quantitative results on NeRF-Synthetic dataset w.r.t. \textbf{LPIPS$\downarrow$} metric.} 
\label{tab:table11}
\end{table*}

\begin{table*}[t]

\makebox[\textwidth]{\small (a) 3 input views.}
\\[0.1em]
\resizebox{\textwidth}{!}{
\rowcolors{2}{LightCyan}{White}
\begin{tabular}{c | c c c c c c c c}
\toprule
LLFF 3-view & Orchids & Leaves & Horns & Flower & T-rex & Room & Fern & Fortress  \\
\midrule
Vanilla NeRF & 12.47	&17.72	&15.65	&18.54	&20.54	&19.52	&17.79	&19.47 \\
\midrule
CombiNeRF &\textbf{ 16.04	}&\textbf{18.46	}&\textbf{18.84	}&\textbf{21.01	}&\textbf{21.21	}&\textbf{21.47	}&\textbf{22.21	}&\textbf{23.75} \\
\bottomrule
\end{tabular}
}
\\[0.5em]
\makebox[\textwidth]{\small (b) 6 input views.}
\\[0.1em]
\resizebox{\textwidth}{!}{
\rowcolors{2}{LightCyan}{White}
\begin{tabular}{c | c c c c c c c c}
\toprule
LLFF 6-view & Orchids & Leaves & Horns & Flower & T-rex & Room & Fern & Fortress  \\
\midrule
Vanilla NeRF & 16.74	&20.1	&22.32	&22.25	&21.66	&25.58	&23.18	&24.42 \\
\midrule
CombiNeRF &\textbf{ 18.11	}&\textbf{20.46	}&\textbf{23.26	}&\textbf{24.52	}&\textbf{23.91	}&\textbf{28.72	}&\textbf{25.07	}&\textbf{27.9} \\
\bottomrule
\end{tabular}
}
\\[0.5em]
\makebox[\textwidth]{\small (c) 9 input views.}
\\[0.1em]
\resizebox{\textwidth}{!}{
\rowcolors{2}{LightCyan}{White}
\begin{tabular}{c | c c c c c c c c}
\toprule
LLFF 9-view & Orchids & Leaves & Horns & Flower & T-rex & Room & Fern & Fortress  \\
\midrule
Vanilla NeRF & 18.45	&21.07	&24.59	&24.88	&25.27	&27.11	&25.54	&26.76 \\
\midrule
CombiNeRF &\textbf{ 19.1	}&\textbf{21.22	}&\textbf{25.15	}&\textbf{26.13	}&\textbf{26.35	}&\textbf{28.57	}&\textbf{26.17	}&\textbf{28.51} \\
\bottomrule
\end{tabular}
}

\caption{Per-scene quantitative results on LLFF dataset w.r.t. \textbf{PSNR$\uparrow$} metric.} 
\label{tab:table12}
\end{table*}

\begin{table*}[t]

\makebox[\textwidth]{\small (a) 3 input views.}
\\[0.1em]
\resizebox{\textwidth}{!}{
\rowcolors{2}{LightCyan}{White}
\begin{tabular}{c | c c c c c c c c}
\toprule
LLFF 3-view & Orchids & Leaves & Horns & Flower & T-rex & Room & Fern & Fortress  \\
\midrule
Vanilla NeRF & 0.199	&0.61	&0.523	&0.548	&0.719	&0.755	&0.509	&0.486 \\
\midrule
CombiNeRF &\textbf{ 0.461	}&\textbf{0.666	}&\textbf{0.692	}&\textbf{0.668	}&\textbf{0.773	}&\textbf{0.821	}&\textbf{0.713	}&\textbf{0.692} \\
\bottomrule
\end{tabular}
}
\\[0.5em]
\makebox[\textwidth]{\small (b) 6 input views.}
\\[0.1em]
\resizebox{\textwidth}{!}{
\rowcolors{2}{LightCyan}{White}
\begin{tabular}{c | c c c c c c c c}
\toprule
LLFF 6-view & Orchids & Leaves & Horns & Flower & T-rex & Room & Fern & Fortress  \\
\midrule
Vanilla NeRF & 0.499	&0.738	&0.797	&0.725	&0.816	&0.894	&0.744	&0.688 \\
\midrule
CombiNeRF &\textbf{ 0.584	}&\textbf{0.753	}&\textbf{0.823	}&\textbf{0.815	}&\textbf{0.863	}&\textbf{0.921	}&\textbf{0.811	}&\textbf{0.873} \\
\bottomrule
\end{tabular}
}
\\[0.5em]
\makebox[\textwidth]{\small (c) 9 input views.}
\\[0.1em]
\resizebox{\textwidth}{!}{
\rowcolors{2}{LightCyan}{White}
\begin{tabular}{c | c c c c c c c c}
\toprule
LLFF 9-view & Orchids & Leaves & Horns & Flower & T-rex & Room & Fern & Fortress  \\
\midrule
Vanilla NeRF & 0.605	&0.777	&0.86	&0.82	&0.885	&0.912	&0.827	&0.805 \\
\midrule
CombiNeRF &\textbf{ 0.647	}&\textbf{0.782	}&\textbf{0.876	}&\textbf{0.857	}&\textbf{0.908	}&\textbf{0.93	}&\textbf{0.85	}&\textbf{0.877} \\
\bottomrule
\end{tabular}
}

\caption{Per-scene quantitative results on LLFF dataset w.r.t. \textbf{SSIM$\uparrow$} metric.} 
\label{tab:table13}
\end{table*}

\begin{table*}[t]

\makebox[\textwidth]{\small (a) 3 input views.}
\\[0.1em]
\resizebox{\textwidth}{!}{
\rowcolors{2}{LightCyan}{White}
\begin{tabular}{c | c c c c c c c c}
\toprule
LLFF 3-view & Orchids & Leaves & Horns & Flower & T-rex & Room & Fern & Fortress  \\
\midrule
Vanilla NeRF &  0.435	&0.181	&0.35	&0.3	&0.195	&0.269	&0.402	&0.292 \\
\midrule
CombiNeRF &\textbf{ 0.25	}&\textbf{0.155	}&\textbf{0.212	}&\textbf{0.226	}&\textbf{0.15	}&\textbf{0.184	}&\textbf{0.194	}&\textbf{0.157} \\
\bottomrule
\end{tabular}
}
\\[0.5em]
\makebox[\textwidth]{\small (b) 6 input views.}
\\[0.1em]
\resizebox{\textwidth}{!}{
\rowcolors{2}{LightCyan}{White}
\begin{tabular}{c | c c c c c c c c}
\toprule
LLFF 6-view & Orchids & Leaves & Horns & Flower & T-rex & Room & Fern & Fortress  \\
\midrule
Vanilla NeRF & 0.23	&0.124	&0.137	&0.156	&0.133	&0.112	&0.166	& 0.137 \\
\midrule
CombiNeRF &\textbf{ 0.178	}&\textbf{0.12	}&\textbf{0.122	}&\textbf{0.095	}&\textbf{0.094	}&\textbf{0.08	}&\textbf{0.105	}&\textbf{0.054} \\
\bottomrule
\end{tabular}
}
\\[0.5em]
\makebox[\textwidth]{\small (c) 9 input views.}
\\[0.1em]
\resizebox{\textwidth}{!}{
\rowcolors{2}{LightCyan}{White}
\begin{tabular}{c | c c c c c c c c}
\toprule
LLFF 9-view & Orchids & Leaves & Horns & Flower & T-rex & Room & Fern & Fortress  \\
\midrule
Vanilla NeRF & 0.171	& \textbf{0.106}	&0.091	&0.093	&0.078	&0.083	&0.101	&0.084 \\
\midrule
CombiNeRF &\textbf{ 0.151	}& 0.112 &\textbf{0.078	}&\textbf{0.071	}&\textbf{0.06	}&\textbf{0.066	}&\textbf{0.082	}&\textbf{0.05} \\
\bottomrule
\end{tabular}
}

\caption{Per-scene quantitative results on LLFF dataset w.r.t. \textbf{LPIPS$\downarrow$} metric.} 
\label{tab:table14}
\end{table*}

\end{document}